\documentclass[]{article}

\usepackage{amssymb}

\usepackage[a4paper, total={6in, 9.5in}]{geometry}

\usepackage{lineno}
\usepackage{natbib}

\usepackage[utf8]{inputenc} % allow utf-8 input
\usepackage[T1]{fontenc}    % use 8-bit T1 fonts
\usepackage{xurl}            % simple URL typesetting
\usepackage{booktabs}       % professional-quality tables

\usepackage{nicefrac}       % compact symbols for 1/2, etc.
\usepackage{listings}
\usepackage{amsmath}
\usepackage{amssymb}
\usepackage{amsthm}
\usepackage{amsfonts}       % blackboard math symbols
\usepackage{inputenc, caption}
\usepackage{balance}
\usepackage[normalem]{ulem}
\usepackage{multicol}
\usepackage{multirow}
\usepackage{placeins}
\usepackage{subcaption}
\usepackage{wrapfig}
\usepackage{enumitem}
\usepackage{hyperref}
\usepackage{custom_styles}
\usepackage{wrapfig}
\usepackage{tikz}
\usepackage{authblk}

\begin{document}



%\title{Joint Encoding of Optical and Radar Satellite Time Series for Crop Mapping}
\title{Multi-Modal Temporal Attention Models \\ for Crop Mapping from  Satellite Time Series}

\author[1]{Vivien Sainte Fare Garnot}
\author[1]{Loic Landrieu}
\author[2]{Nesrine Chehata}
\affil[1]{LASTIG, ENSG, IGN, Univ Gustave Eiffel, Saint-Mande,France}
\affil[2]{EA G\&E Bordeaux INP, Univ Bordeaux Montaigne, Bordeaux}

%\author{}

%\affiliation{organization={},%Department and Organization
\maketitle

\begin{abstract}
Optical and radar satellite time series are synergetic: optical images contain rich spectral information, while C-band radar captures useful geometrical information and is immune to cloud cover. Motivated by the recent success of temporal attention-based methods across multiple crop mapping tasks, we propose to investigate how these models can be adapted to operate on several modalities. We implement and evaluate multiple fusion schemes, including a novel approach and simple adjustments to the training procedure, significantly improving performance and efficiency  with little added complexity. We show that most fusion schemes have advantages and drawbacks, making them relevant for specific settings.
We then evaluate the benefit of multimodality across several tasks: parcel classification, pixel-based segmentation, and panoptic parcel segmentation. We show that by leveraging both optical and radar time series, multimodal temporal attention-based models can outmatch single-modality models in terms of performance and resilience to cloud cover.
To conduct these experiments, we augment the PASTIS dataset \citep{pastis} with spatially aligned radar image time series. The resulting dataset, PASTIS-R, constitutes the first large-scale, multimodal, and open-access satellite time series dataset {with semantic and instance annotations.}
%allowing for both parcel-, pixel-, and instance-level evaluations. 
\end{abstract}



\section{Introduction}
%Global-scale and open access satellite data are a valuable resource for the automation of many key earth observation tasks with machine learning methods. In particular, multi-temporal satellite datasets such those provided by the Sentinel mission, allow to model the temporal evolution of complex phenomena such as the growth cycle of plants. Moreover, the increasing variety of sensors available allows one to combine different multi-temporal modalities to leverage their respective advantages.

%The growing quantity and quality of open-access satellite images is also accompanied by an increase in the number of available modality for embarked sensors. In particular, multi-spectral optical and radar images capture complimentary radiometric and geometric properties which are especially relavant for crop type mapping.
The multiplication of Earth Observation satellites with various sensors represents an opportunity to improve the automated analysis of remote sensing data for tasks such as crop mapping.
Indeed, different modalities capture information of different natures and distinct spatial and temporal resolutions and have varying resilience to atmospheric conditions.
Machine learning models can leverage these complementary characteristics to learn richer and more robust representations.
In particular, {C-band} radar and optical images possess well-known synergies for automated crop mapping  \citep{van2018synergistic, steinhausen2018combining, campos2019copernicus}, the driving application of this paper.
More specifically, multispectral time series contain highly relevant information for monitoring the evolution of plant phenology \citep{vrieling2018vegetation,segarra2020remote}. For example, the  {study of red and infrared reflectances helps monitoring} photosynthetic activity \citep{TUCKERNDVI1979127}.
However, passive optical sensors are highly susceptible to cloud cover and atmospheric distortion \citep{sudmanns2020assessing}.
Conversely, %radar images \VIVIEN{due to dependency on a variety of factors tend to} contain information which are less conductive to crop mapping,
due to the influence of extrinsic factors such as humidity and terrain, it is harder to extract discriminative information from radar images for crop mapping. On the other hand, the high revisit frequency and imperviousness to cloud cover make them uniquely well-suited for monitoring the rapid-changing biological processes of agricultural parcels \citep{mcnairn2014early}.

Automated crop mapping is necessary for various applications carrying crucial economic and ecological stakes, such as environmental monitoring, subsidy allocation, and food price prediction. For example, the Common Agricultural Policy is responsible for the allocation of over $57$ billion euros each year of agricultural subsidies in the European Union \citep{CAP}. This application is at the center of an effort towards  algorithmic solutions for machine learning-based crop monitoring \citep{koetz2019sen4cap}. This endeavor is aided by the accessibility of high-quality satellite data worldwide \citep{drusch2012sentinel} and individual parcel annotations for some European countries such as France \citep{RPG}, which is particularly conducive to training large-scale deep networks.

In the context of crop type mapping, the fusion of optical and radar time series has been extensively explored with traditional machine learning methods \citep{van2018synergistic, steinhausen2018combining,he2018multi,campos2019copernicus,orynbaikyzy2020crop,giordano2020improved}, and more recently recurrent neural networks \citep{ienco2019combining}.
However, despite the significant performance gain offered by methods based on temporal attention \citep{russwurm2020self, garnot2020satellite, kondmann2021denethor, garnot2020lightweight},
these approaches are mostly restricted to the analysis of optical Satellite Image Time Series (SITS). {Recently, \citet{pelletier2021fusion} proposed a first exploration of the benefit of fusion strategies for parcel-based crop type classification from Sentinel-1 and Sentinel-2 time series with attention-based methods. The present paper extends their analysis to a broader set of crop mapping tasks, including semantic and panoptic segmentation\citep{kirillov2019panoptic, garnot2021utae} of agricultural parcels. We also study the performance benefit of standard enhancements such as auxiliary supervision and temporal dropout. %To encourage further work on the topic, we realease PASTIS-R, a large-scale dataset of multimodal time series with semantic and panoptic annotations.
}
%In this paper, we propose to explore different strategies for combining SITS from multiple modalities and temporal attention models, with a focus on crop mapping and the v satellites. We implement several feature fusion schemes commonly encountered in the literature and also propose a novel strategy. We present simple enhancements such as auxiliary supervision and temporal dropout to improve performance. We evaluate the benefit of multimodality on three tasks: parcel classification, semantic segmentation, and panoptic segmentation \citep{kirillov2019panoptic, garnot2021utae}.
\begin{figure}[t!]
    \centering
    \includegraphics[width=\textwidth, trim=0cm 17cm 0cm 0cm, clip]{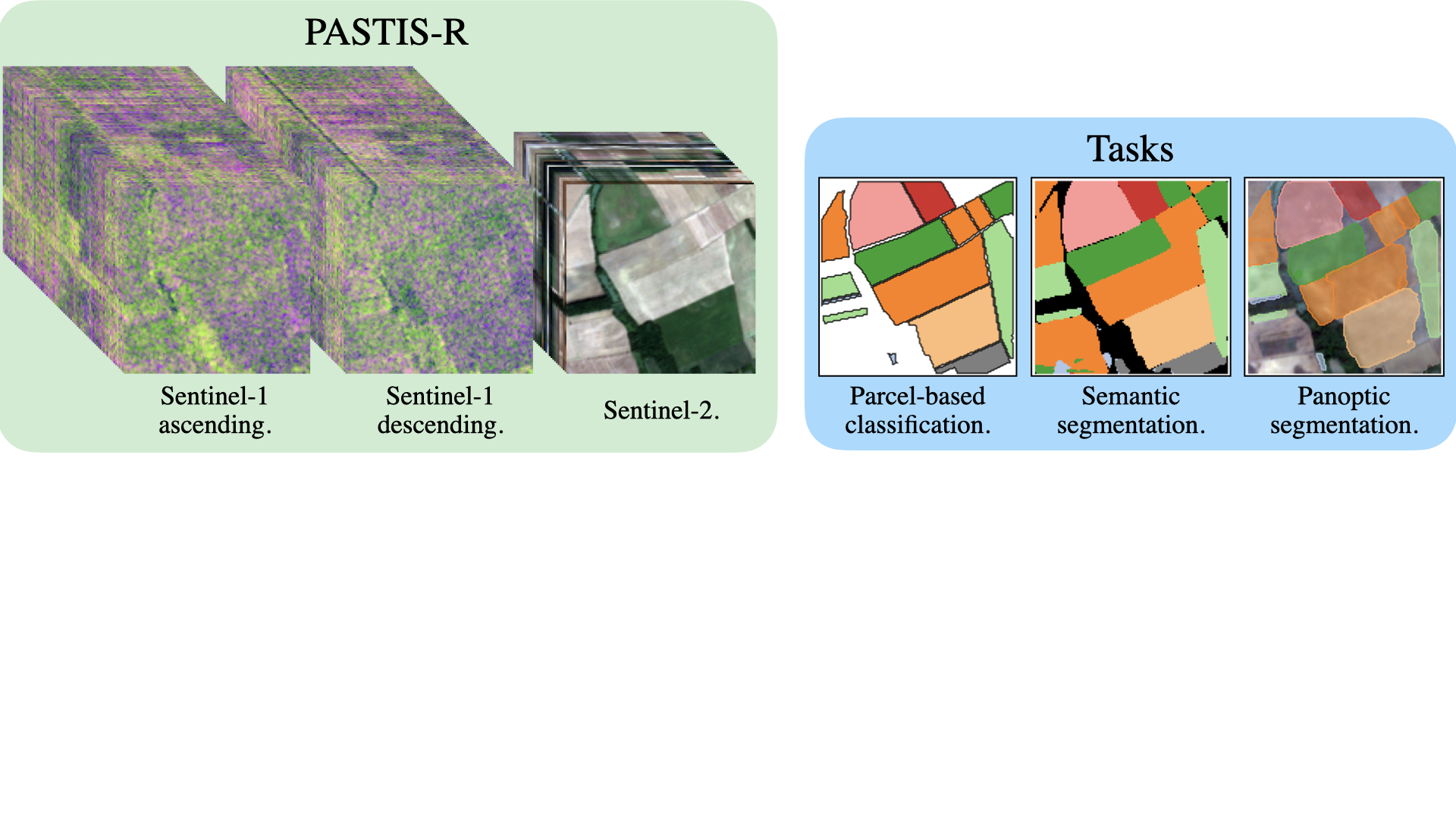}
    \caption{ We introduce the PASTIS-R dataset containing $2433$ multimodal image time series of Sentinel-2 and Sentinel-1 data. On PASTIS-R, we evaluate different fusion strategies and enhancements on parcel-based classification, semantic segmentation, and panoptic segmentation.}
    \label{fig:teaser}
\end{figure}
%The first step of our method processes each image simultaneously with a modality-dedicated spatial encoder. Then, an attention-based temporal encoder \cite{garnot2020lightweight} processes the joint sequence of features into a single multimodal learned representation. This general approach can be integrated into various Earth observation tasks, such as parcel classification, semantic segmentation, and even panoptic segmentation \cite{garnot2021utae}.

To train and evaluate our models, we augment the open-access PASTIS optical time series dataset \citep{pastis} with corresponding Sentinel-1 radar acquisitions for each $2\,433$ time series for a total of $339\,174$ radar images. We demonstrate that the right choice of fusion scheme can lead to improvement across the board for all tasks and increased robustness to varying cloud cover.
%Our modality fusion scheme makes our model  more resilient to varying cloud coverage than other classic data fusion schemes. Moreover, when combined with a well-suited augmentation strategy, our joint temporal encoding allows for the extraction of cross-modality features, ultimately leading to a higher classification accuracy. 
%
%We argue that a key aspect of multi-temporal modality-fusion methods is the ability to perform joint temporal encoding. Such joint encoding facilitates the extraction of cross-modality temporal patterns. Methods that concatenate the two modalities before encoding allow such joint encoding but resort to costly preprocessing, as the two time series need to be resampled to a common temporal grid. Conversely, our approach allows for joint temporal encoding without  pre-processing. Lastly, we introduce an augmentation scheme designed to encourage the extraction of cross-modality patterns.
%We showcase the interest of our methods on a novel public benchmark dataset for parcel-based classification, semantic segmentation and panoptic segmentation of agricultural parcels.
The main contributions of this paper are as follows:
\begin{itemize}
    \item We present a complete reformulation of  fusion strategies in the context of temporal attention-based SITS encoders, as well as standard model enhancements.
    \item We present PASTIS-R, the first large-scale, multimodal, open-access SITS dataset with panoptic annotations.
    \item We evaluate our fusion schemes and enhancements on parcel classification,  semantic and panoptic segmentation, defining a new state-of-the-art for all tasks.
    \item We show that combining optical and radar imagery grants significant improvement in terms of robustness to varying cloud cover.
\end{itemize}
\section{Related Work}
\label{sec:fusion_related}

In the following paragraphs, we review the recent literature on fusion approaches for the multi-temporal fusion of SITS. In particular, we detail commonly implemented fusion strategies.
%We note that some ambiguity can arise in the terms used to describe the different approaches, mostly because of the shift from machine learning to deep learning approaches. We thus give an explicit description of each approach in the concluding paragraph.

\paragraph{\bf Traditional Approaches for Multimodal SITS}
Multiple traditional machine learning approaches such as random forest or support vector machines have been adapted to handle information from  optical and radar images.
As highlighted by the review of \citet{joshi2016review}, the joint processing of both modalities can mitigate the sensitivity of optical images to cloud cover.
%Joshi \etal\cite{joshi2016review}  advocate for the joint processing of optical and radar images to overcome the sensitivity of optical images to cloud cover.
%Multiple traditional machine learning approaches such as random forest or support vector machines have been adapted to handle information from both sensors.
Most methods use an early fusion scheme in which the radar and optical features are stacked before being processed by the model \citep{van2018synergistic, mercier2019evaluation}. This approach can be further improved by selecting the most relevant acquisitions \citep{steinhausen2018combining} or features \citep{campos2019copernicus, giordano2020improved}.
\citet{orynbaikyzy2020crop} compare this feature concatenation approach with a decision fusion approach in which two separate random forest classifiers predict posterior probabilities over classes, and the most confident prediction is retained as the final classification. Their results show that decision fusion performs slightly worse than early feature concatenation.

\paragraph{\bf Deep learning for MultiModal SITS}
The first multimodal deep learning models advocated for an \emph{early fusion} scheme: the channels of all acquisitions from optical and radar time series are concatenated to form a single image with both multimodal and multitemporal pixel features. The resulting images are then processed pixelwise \citep{tarpanelli2018daily} or with convolutional networks \citep{kussul2017deep}.
{In contrast,} \citet{ienco2019combining} propose to encode each radar and an optical time series separately using a combination of dedicated convolutional and recurrent-convolutional networks. In a \emph{late-fusion} fashion, all resulting embeddings are concatenated channelwise and classified pixelwise by a Multi-Layer Perceptron (MLP). They observe that, as long as each branch is also supervised separately with auxiliary loss terms, this fusion scheme outperforms early fusion. 
{More recently, \citet{pelletier2021fusion} studied four fusion strategies for parcel-based classification with a PSE-TAE architecture \citep{garnot2020satellite}. Early fusion yields the best improvement on their dataset of Sentinel-2 time series and Sentinel-1 observations in descending orbit. 
We extend their analysis by evaluating the impact of multimodality for different tasks, evaluate the effects of typical enhancements such as auxiliary classifiers, and use both Sentinel-1 orbits in our analysis.
%which are found to be crucial in \citet{ienco2019combining} but not evaluated in \citet{pelletier2021fusion}. %Furthermore, we comparatively assess the fusion strategies on three crop type mapping tasks: parcel-based classification, semantic segmentation, and panoptic segmentation. %Lastly, we assemble a large multi-modal dataset with Sentinel-1 observations in ascending and descending orbit, and make it publicly available.
}

\paragraph{\bf Other Fusion settings}
In a different setting, \citet{benedetti2018m} use a late fusion approach to combine mono-temporal high spatial resolution images with low spatial resolution time series, and  \citet{tom2021fusionVIIRSS1} exploit three different mono-temporal modalities for lake ice monitoring by training three encoders to map the different acquisitions to a common feature space. \citet{liu2016deep} explore multimodal change detection on mono-temporal pairs. They propose to train two encoders in an unsupervised fashion to map  simultaneously-acquired images  of different modalities to a common feature space. 
More broadly, the synergy between radar and optical SITS has motivated other exciting applications such as the regression of optical signals from radar images \citep{garioud2020joint, meraner2020cloudcgan,he2018multi}.

%
%\paragraph{Attention-Based multimodal Analysis.}Self-attention has proven a well-suited mechanism for merging complementary data modalities. Such approaches have been successfully employed for joint text and image processing \cite{caglayan2016multimodal, huang2016attention, gu2018multimodal}, visual question answering \cite{fukui2016multimodal, ben2019block}, audio and video analysis \cite{hori2017attention, long2018multimodal}, and joint 2D-3D processing \cite{li2020attention, dai20183dmv}.
%Assumes temporal alignment:
%\cite{caglayan2016multimodal, gu2018multimodal}
%One modality is mono-temporal:
%\cite{huang2016attention, fukui2016multimodal, %ben2019block}
%Use instead of concatenation:
%\cite{hori2017attention}

\paragraph{\bf Radar processing}

Data analysis from Synthetic-Aperture Radar (SAR) relies on either extracting backscattering coefficients, interferometric,  or polarimetric features from a measured radar signal \citep{richards2009remote}.
Backscattering coefficients are most commonly used for crop type mapping applications \citep{orynbaikyzy2019crop}. These approaches derive information on the observed surface's geometric properties and dielectric constant from the amplitude of the complex SAR signal, and discard the phase information. In contrast, interferometric SAR measure phase shift to detect potentially small deformations between two acquisitions. Interferometric features are traditionally used in geodesy \citep{simons2007interferometric} and surface \citep{monserrat2014review,tarchi2003landslide} or structural \citep{tomas2012subsidence,tarchi1997monitoring, tisontupin2007fusion} monitoring, but also proved discriminative for crop type mapping. Indeed, coeherence estimation in interferometry can help detecting mowing, harvesting, and seeding events \citep{tamm2016relating, mestre2020time, shang2020detection}, as well as providing information on crop height and density \citep{srivastava2006application}. Lastly, polarimetric SAR data analysis relies on target decomposition of polarimetric information \citep{cloude1996review, yamaguchi2005four} to provide additional terrain information, and can be used for canopy structure estimation \citep{srikanth2016comparison}, topography \citep{schuler1996measurement}, or land cover estimation \citep{tupin1998detection,kourgli2010land}. However, such approaches require full polaristion radar images, \ie, acquired with a sensor emitting radar waves along both polarisation directions.
In this paper, we focus on crop type mapping from data of the open acces Sentinel-1 sensor which does not allow such full polarimetric analyses. Furthermore, to limit the complexity of our experiments and avoid downloading very large Single Look Complex datasets, we focus on SAR backscattering coefficients and leave the extension to interferometric features to further work.

\section{Methods}
\begin{figure}
    \centering
    \begin{subfigure}{0.32\linewidth}
    \includegraphics[width=\linewidth, trim=0cm 3.5cm 0cm 2cm, clip]{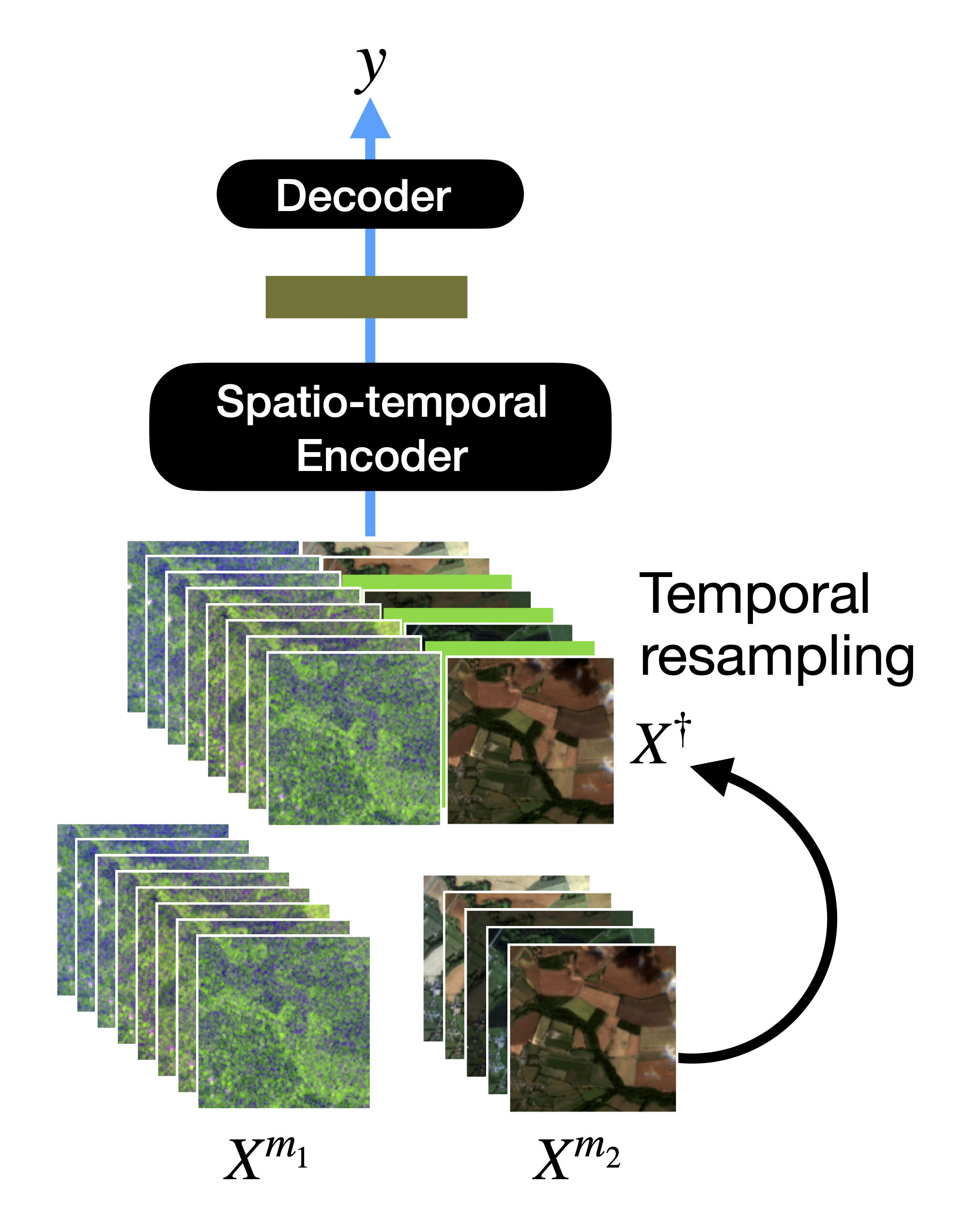}
    \caption{Early fusion}
    \label{fig:fusion_schemes:early}
    \end{subfigure}
    \hfill
    \begin{subfigure}{0.32\linewidth}
    \includegraphics[width=\linewidth]{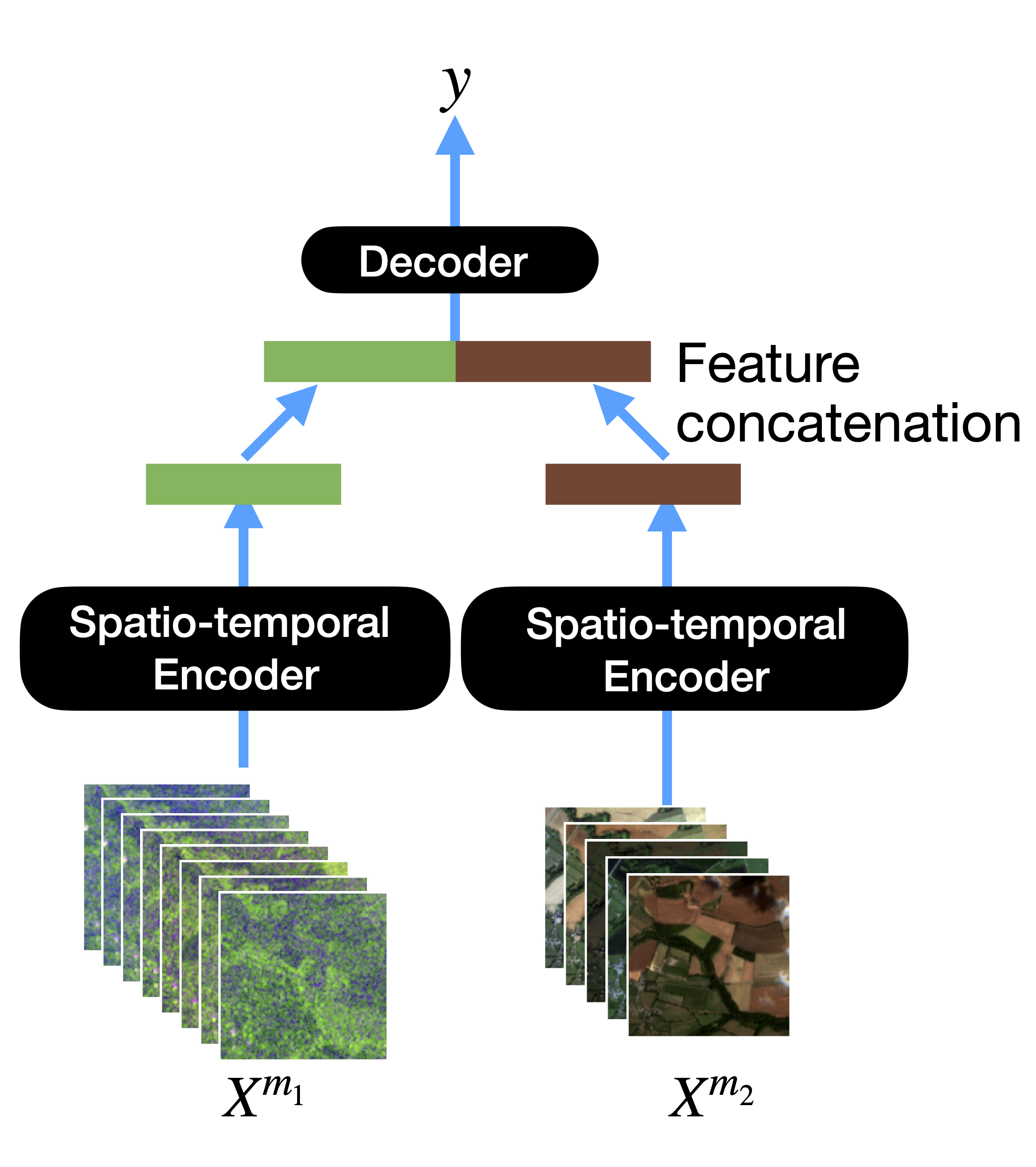}
    \caption{Late fusion}
    \label{fig:fusion_schemes:late}
    \end{subfigure}
    \hfill
    \begin{subfigure}{0.32\linewidth}
    \includegraphics[width=\linewidth]{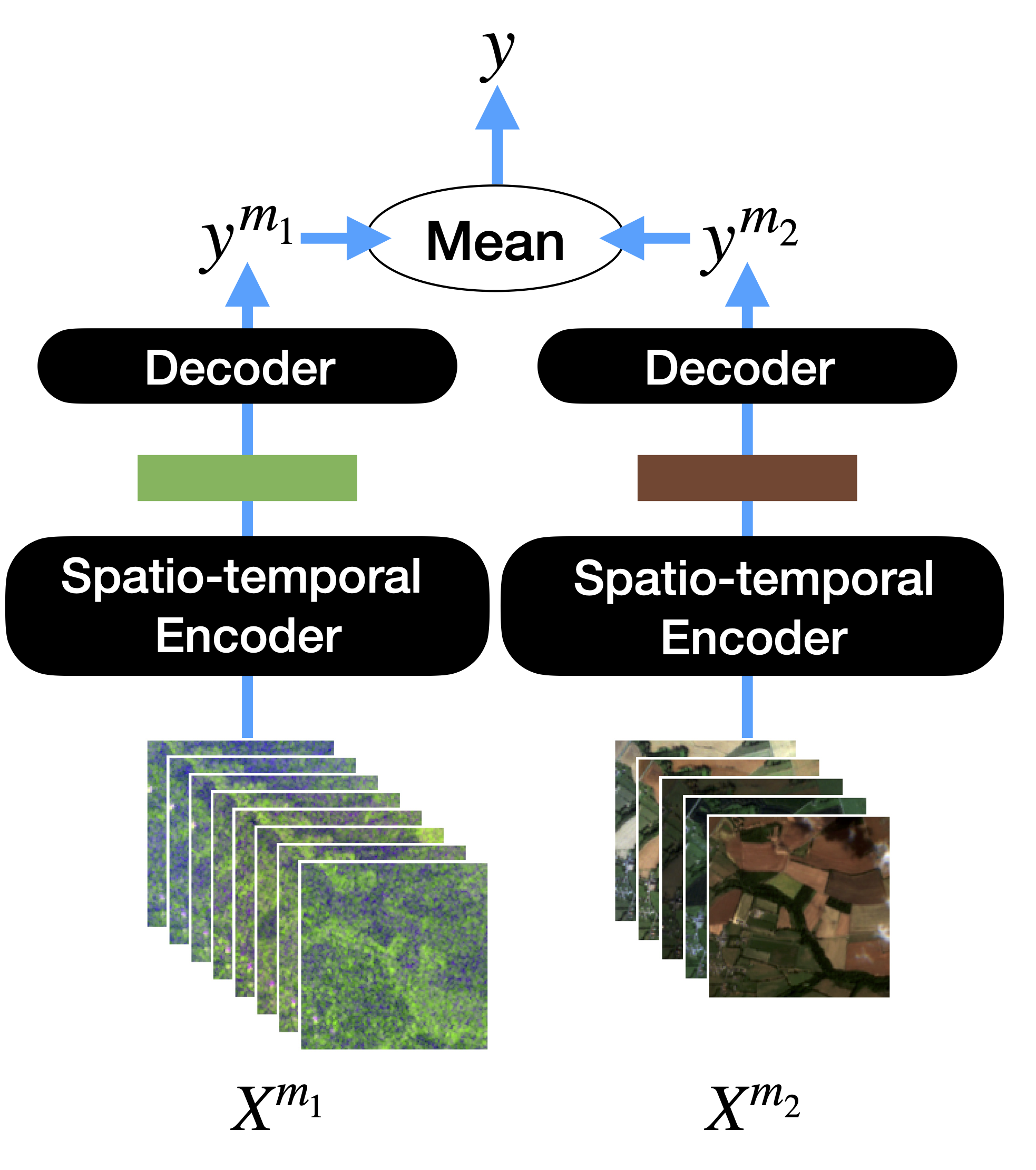}
    \caption{Decision fusion}
    \label{fig:fusion_schemes:decision}
    \end{subfigure}
    
    \caption{{\bf Fusion Schemes.} We represent the three fusion strategies commonly found in the recent literature. \Subfigref{fig:fusion_schemes:early} the raw features are interpolated and concatenated into a single sequence. \Subfigref{fig:fusion_schemes:late} the learned spatio-temporal features of each modality are concatenated prior to classification. \Subfigref{fig:fusion_schemes:decision} each modality is processed independently and the resulting decision averaged.}
    \label{fig:fusion_schemes}
\end{figure}
%

%\begin{figure}
%    \centering
%    \begin{tabular}{ccc}
%     \includegraphics[width=.25\linewidth]{Fusion/gfx/early.pdf}    & 
%     \includegraphics[width=.25\linewidth]{Fusion/gfx/early.pdf}&
%     \includegraphics[width=.25\linewidth]{Fusion/gfx/early.pdf}\\
%      (a)  & (b) &(c) 
%    \end{tabular}
%    
%    \caption{Caption}
%    \label{fig:my_label}
%\end{figure}

We consider a set of $M$ image time series $\{X\m\}_{m=1}^M$ corresponding to $M$ distinct modalities for a single geo-referenced patch containing one or several agricultural parcels.
We assume that all modalities are resampled to the same resolution for simplicity's sake.
Each time sequence $X\m$ can be expressed as a tensor of size $T\m \times C\m \times H \times W$ with $T\m$ the number of available temporal acquisitions for modality $m$,  $C\m$ the feature size for each pixel for the modality $m$, and $H \times W$ the spatial extent of the patch.

%..................................................................
\subsection{Fusion Strategies}
%..................................................................
The methods reviewed previously can be categorised into three main strategies: \emph{early}, \emph{late}, and \emph{decision} fusion, all represented in \figref{fig:fusion_schemes}. We also present  \emph{mid}-fusion, a novel fusion scheme specifically adapted for multimodal time sequences.
Certain terms---such as ``features''---have seen their accepted meaning evolve with the gradual adoption of the deep learning paradigm, leading to ambiguity in terms such as ``early'' or ``late'' feature fusion.
We propose redefining the terminology of fusion schemes unambiguously for analyzing temporal sequences of images in the following.
%This terminology presents some ambiguity due to the shift from machine learning approaches to deep learning models. Indeed, in approaches using traditional machine learning classifiers, the input data usually consists of engineered features. In that sense, early machine learning feature fusion rather corresponds to late feature fusion in a deep learning context. To avoid any confusion, we first formalise in an abstract form the way in which each strategy operates in a deep learning context.

\paragraph{\bf Early Fusion}
This approach combines the different modalities  at the raw feature level. In our context, this amounts to concatenating the modalities \emph{channel-wise} at each observation date. If the different acquisitions are simultaneous, and since the resolutions are identical, this is a straightforward step.
However, when the modalities are captured at different times, {a} preprocessing step is required to interpolate all modalities to a standard temporal sampling.
We denote by $T^\dagger$ the number of time steps in the chosen temporal sampling and by $\tX$ the resulting aggregated tensor of size $T^\dagger \times C^\dagger \times H \times W$ with $C^\dagger=\sum_m C\m$ as defined in \equaref{eq:interpolate}.

%This approaches combines the different modalities at the earliest stage. In the deep learning paradigm, this amounts to concatenating observations of different sensors into one fused observation. This scheme thus starts by interpolating all modalities to a common temporal sampling and concatenating all pixel features in the channel dimension.
%We denote by $T^\dagger$ the number of time steps in the chosen temporal sampling and by $\tX$ the resulting aggregated tensor of size $T^\dagger \times C^\dagger \times H \times W$ with $C^\dagger=\sum_m C\m$.

This interpolation step can be costly in terms of computation and memory. Furthermore, the relevance of temporal interpolation for fast-changing processes such as plant growth and harvesting is questionable, and this is only made worse by clouds obstructing the optical modalities. 
However, an advantage of this approach is the simplicity of encoding $X^\dagger$: a single spatio-temporal encoder $\spatiotemporal$ 
can be used to learn a truly cross-modal representation, and a unique decoder $\cD$ produces the final prediction:
\begin{align} \label{eq:interpolate}
  \tX = & 
  \merge{C}{
  \left\{
    \text{interpolate}(X\m)
    \;\text{to}\; T^\dagger
  \right\}_{m=1}^M}\\
  y^{\text{early}} = &\cD\comp\spatiotemporal(\tX)~.
\end{align}

\paragraph{\bf Late Feature Fusion}
This fusion scheme starts by encoding each modality $m$ separately with dedicated spatio-temporal encoders $\spatiotemporal\m$ into embeddings of size $F\m$. These vectors are then concatenated for all modalities along the channel dimension into a vector of size $\sum_m F\m$, which is ultimately mapped to a prediction $y^{\text{late}}$ by a unique decoder $\cD$:
\begin{align}
  y^{\text{late}} & = 
  \cD\comp
  \merge{C}{
    \left\{
      \spatiotemporal\m
      \left(
        X\m
       \right)
    \right\}_{m=1}^M}~,
\end{align}
with $\text{merge}^{(C)}$ the channelwise concatenation operator. While each latent features are derived from a single modality, this method allows the decoder to make decisions taking all modalities into account simultaneously.
%.................................................
\paragraph{\bf Decision Fusion}
This approach ignores the interplay between modalities and makes a prediction for each modality independently. A set of $M$ spatio-temporal encoder $\spatiotemporal\m$ maps each sequence of size $T\m \times C\m \times H \times W$ to a latent space of size $F\m$. {Then, a} set of $M$ decoders $\cD\m$ maps each spatio-temporal feature into a prediction. {Finally, an} aggregation rule {is} applied to combine {all} $M$ predictions into {a} final prediction $y^{\text{decision}}$. Typically, predictions are averaged across all available modalities :
\begin{align}
 y^{\text{decision}} & =
 \frac1M \sum_{m=1}^M
 \cD\m
 \comp
 \spatiotemporal\m
     \left(
       X\m
      \right)
 ~.
\end{align}
%.................................................

\paragraph{\bf Mid-Fusion}

\begin{figure*}[h]
  \begin{center}
    \includegraphics[width=0.32\textwidth]{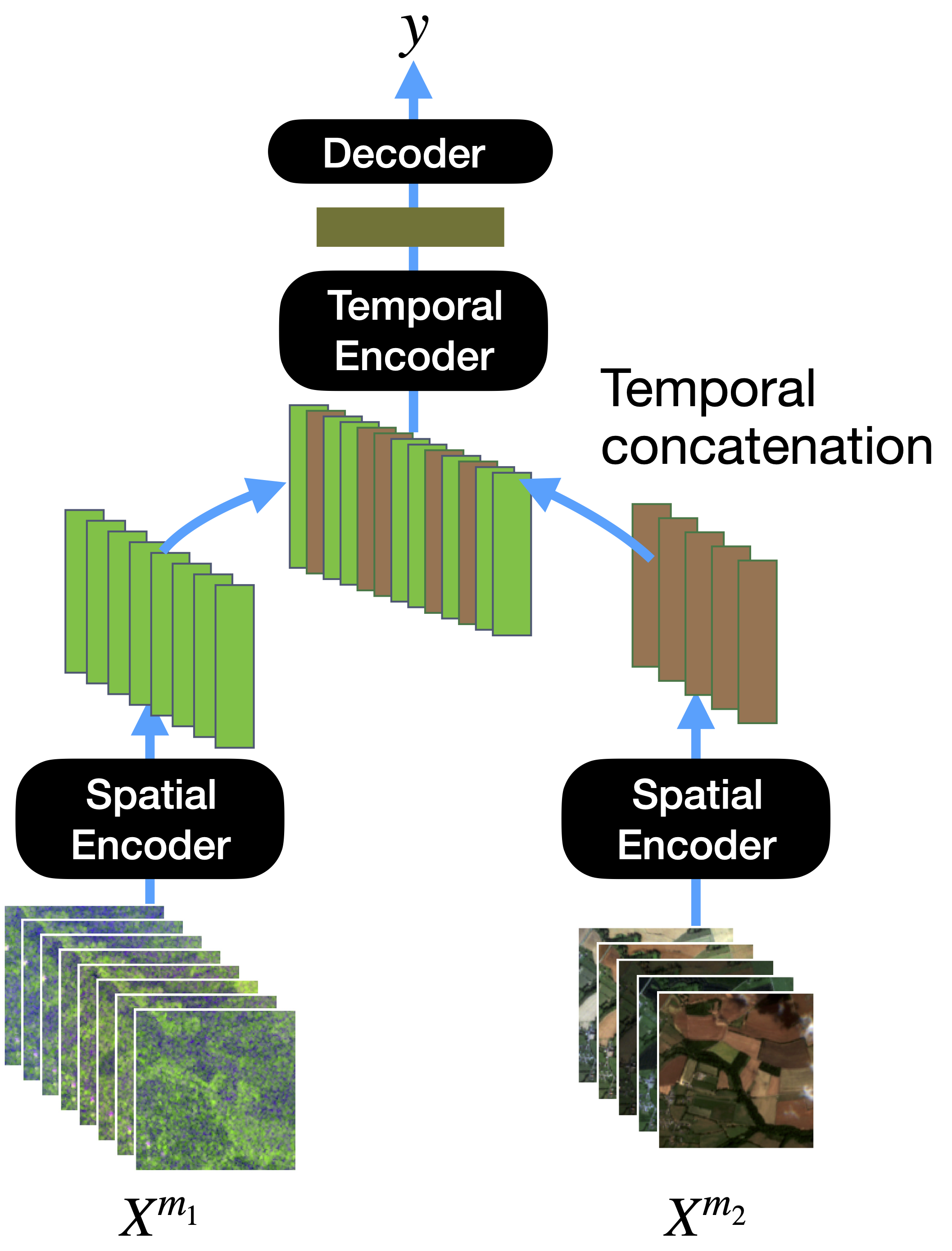}
  \end{center}
  \caption{{\bf Mid-Fusion.} A dedicated spatial encoder processes each modality, and the resulting features are stacked into a single sequence of features.}
  \label{fig:stack}
\end{figure*}

Specific network architectures used to process temporal sequences such as SITS can be broken down into a spatial and a temporal encoder. In such cases, the spatial features can be interwoven, \ie temporally stacked, into a single multimodal sequence, see \figref{fig:stack}.
This approach can be seen as a compromise between early and late fusion and combines three of their advantages: 
(i) the temporal encoder can leverage all modalities simultaneously,
(ii) only one temporal encoder is needed, 
(iii) no heavy preprocessing is necessary to merge the feature sequences as they are stacked.

Each modality $m$ has a dedicated spatial encoder $\spatial\m$ mapping images to a feature vector of size $F\m$. These vectors are then concatenated chronologically along the temporal dimension into a unique sequence of length $\sum_m T\m$. A unique temporal encoder $\temporal$ maps this sequence of features into a unique vector, which is in turn classified by a unique decoder $\cD$:
\begin{align}
  y^{\text{mid}} &=
  \cD
 \comp
   \temporal
     \comp
       \merge{T}
       {
         \left\{
           \spatial\m
             \left(
               X\m
             \right)
         \right\}_{m=0}^{M}
         }
  ~,
\end{align}
with $\text{merge}^{(T)}$ the operator concatenating a set of tensors along {the} temporal dimension.
%--------------------------------------------
\subsection{Auxiliary Supervision}
%--------------------------------------------
\label{sec:aux}
We denote by $\text{criterion}(\cdot\,,\,\cdot)$ the function used to compare the prediction $y$ with the target signal $\hat{y}$. This is typically the cross-entropy for parcel or pixel classification and can be more complex for panoptic or instance segmentation \citep{garnot2021panoptic}.
The resulting function $\mathcal{L}_\text{obj}$ is called the {objective} loss and supervizes the {prediction $y$ of the }network to realize the {sought task}:
\begin{align}
\mathcal{L}_\text{obj}=\text{criterion}(y,\hat{y})
\end{align}
%In all the schemes described earlier, the final prediction $y$ of the model is supervised against the target signal $\hat{y}$ with a loss function that depends on the problem at hand, \eg cross entropy loss for parcel-based classification or semantic segmentation :
%\begin{align}
%    \mathcal{L} = \text{loss} (y,  \hat{y})  ~.
%\end{align}

A common problem in deep feature fusion is encountered when most (but not all) discriminative information is concentrated among a reduced number of modalities.
In this case, the other modalities yield predictions and less relevant features for the considered task. Consequently, the final decision taken by the multimodal network focuses on the \emph{better} modalities, and the parts of the network operating on the \emph{lesser} modalities receive a weaker supervisory signal. This results in a network that may not fully leverage the inter-modal patterns that would otherwise allow the multimodal prediction to outperform the \emph{best} modality. This is typically the case for Sentinel SITS, as multispectral optical acquisitions are {often} more conductive to capture phenological patterns than SAR information. Sentinel-1 signal is indeed affected by local terrain angle \citep{kaplan2021S1localinc}, humidity \citep{garkusha2017research}, and is subject to speckle \citep{abramov2017speckle}. 

To mitigate this issue, we can add auxiliary losses to supervise each modality independently on top of the {objective} loss $\mathcal{L}_\text{obj}$. \citet{ienco2019combining} has shown this strategy to help to combine optical and radar imagery.
To this end, we associate a prediction $y^m$ to each modality, which is supervised by the auxiliary loss {$\mathcal{L_\text{aux}}$}:
\begin{align}\label{eq:aux}
    \mathcal{L_\text{aux}} =  \sum_{m=1}^M  \lambda^m \: \text{criterion} (y^m,  \hat{y})  ~,
\end{align}
with $\lambda_m$ the strength associated with each modality. Note that, depending on the chosen fusion scheme, computing the single-modality prediction $y^m$ may imply adding new modules {to the backbone network}. 
This requires $M$ decoders $\cD\m$, in the case of late fusion. For {mid-}fusion, we must add $M$ {temporal} encoders $\temporal\m$ as well. No additional modules are necessary for decision fusion as single-modality predictions $y^m$ are already necessary to produce the final prediction $y$.
In contrast, auxiliary supervision in the case of early fusion would amount to duplicating the entire network, making it both fruitless and costly.

Since the added modules do not participate in the multimodal decision, only the gradients they propagate are beneficial to the training and not their predictions. 
Hence, auxiliary supervision only benefits the modules that  receive gradients from both the main and auxiliary losses. Consequently, auxiliary loss with early supervision would not affect the performance. By the same reasoning, we can expect auxiliary supervision to be beneficial for late and decision fusion than {mid-}fusion for which only the spatial decoders are affected by the auxiliary losses.

%In the case of the decision and late fusion schemes $M$ different spatio-temporal encoders are supervised with a single loss. This can be problematic if the predictions rely more on one modality, and consequently if the gradients of the loss with respect to the parameters of the other modalities' encoders remain relatively low. In the case of optical and radar fusion for crop type mapping, the radar modality is notoriously less discriminative than the optical input (REF). As seen in Ienco \etal \cite{ienco2019combining}, a way to circumvent this issue is to also supervise each spatio-temporal encoder individually using auxiliary losses. For decision fusion, the prediction made from each modality can be used. For late fusion, this requires the addition of $M$ decoders $\cD\m$ to decode each individual feature vector of size $F\m$ into a prediction $y^m$. Each modality-based prediction is then supervised with an additional auxiliary loss term that uses the same loss function:
%\begin{align}
%    \mathcal{L_\text{aux}} =  \sum_{m=1}^M  \lambda^m \: \text{loss} (y^m,  \hat{y})  ~.
%\end{align}
%...............................................
\subsection{Temporal Dropout} 
%...............................................
We propose a simple data augmentation strategy called temporal dropout to promote a multimodal model that leverages all available modalities.
Inspired by the classic dropout strategy \citep{srivastava2014dropout}, we {randomly} drop observations from the input sequences. The idea is to prevent the network from over-relying on a single modality since its presence is never assured.
Formally, we associate a dropout probability  $p^m \in [0,1]$ for each modality $m \in [1, M]$. During training, each observation of the sequence is dropped with probability $p^m$. At inference time, the network can use all available observations. Note that this technique can also be used on models operating on a single modality by randomly dropping some acquisitions.

%We also consider a simple data augmentation strategy to promote the dependency of the fusion model on all available modalities. In this aim, we propose to randomly drop observations in the input time series. 
%Formally, this augmentation is parametrized by the dropout probabilities for each modality $p^m \in [0,1], \: m = 1, \cdots, M$. Then every time a sample time series of modality $m$ is processed by the model during training, each observation of the sequence is dropped with probability $p^m$.
%At inference time, all available observations are used. Note that this technique can also be used on models operating on a single modality as a simple data augmentation. When applied to fusion models, we argue that setting a higher drop probability on the information rich modality can increase the dependency of the model on the weaker modalities.

%a set  of image time series combining $M$ distinct modalities. All images are spatially aligned, such that %We consider a multimodal dataset $\mathcal{D}$ of satellite image time series indexed by $\mathcal{N}$:

%\begin{align}
%    \mathcal{D} &= \left\{ \left(\left(X_i^\text{rad}, %X_i^\text{opt}\right), y_i\right) , i \in \mathcal{N} %\right\}\\
%    &\text{with} \quad X_i^\text{rad} \in T_i \times %C^\text{rad} \times H \times W, \\
%    &\text{and} \quad \- X_i^\text{opt} \in T_i \times %C^\text{opt} \times H \times W, \quad \forall i \in  %\mathcal{N}\;.
%\end{align}
%................................................
\subsection{{Implementation.}}
%................................................
\label{sec:implem}
{As we set out to evaluate the benefit of multimodality for several tasks, we detail how our fusion schemes can be integrated into temporal attention-based, state-of-the-art networks.}

\paragraph{\bf Parcel-based classification} 
We first implement the different fusion strategies for parcel-based crop type classification. In this setting, the contour of parcels is {known}, and the task is to classify the cultivated crop {from} a corresponding yearly SITS. 
The spatial and temporal encoding modules we selected are the Pixel-Set Encoder (PSE) and Lightweight Temporal Attention Encoder (L-TAE), whose accuracy and computational efficiency have been solidified in recent studies  \citep{schneider2020re, kondmann2021denethor, garnot2020lightweight, garnot2020satellite}, and whose implementations are available. \footnote{\url{VSainteuf/lightweight-temporal-attention-pytorch}}
All spatio-temporal encoders $\spatiotemporal$ are a combination of a PSE encoding all images of the time series simultaneously and an L-TAE processing the resulting sequence of embeddings, in the manner of \citet{garnot2020satellite}. All decoders $\cD$ are simple Multi-Layer Perceptrons (MLP). All models are trained with {the} cross-entropy loss.
Since we are using these networks in a straightforward manner and without {parameter} alteration, we refer the reader to the \cite[Sec.~3]{garnot2020satellite} for more details on their configuration.
{As explained in \secref{sec:aux}, auxiliary supervision does not affect the early supervision beyond an increased memory load, and we do not evaluate it.}% We also give the corresponding detailed equations in the appendix.

\paragraph{\bf Semantic segmentation} In this setting, the contours of the parcels are unknown, and the model predicts a crop type for each pixel of a {given} patch {from the corresponding yearly SITS}. For this task, we use the state-of-the-art U-TAE architecture \citep{garnot2021panoptic} as {spatio-temporal encoder \footnote{ \url{github.com/VSainteuf/utae-paps}}. The fact that this network's spatial and temporal encoders are intertwined prevents us from applying the mid-fusion scheme.} We use a 2-layer convolutional neural net as decoder $\cD$. The models are trained with cross-entropy loss. 

\paragraph{\bf Panoptic segmentation}
{Panoptic segmentation amounts to retrieving both the boundary and crop type of each agricultural parcel. In practice, we predict a set of non-overlapping instance masks and their semantic labels
 \citep{kirillov2019panoptic}.
In our setting, pixels that do not belong to a predicted parcel are classified as background. For this task, we also use U-TAE for spatio-temporal encoding, combined with the instance segmentation module Parcel-as-Points (PaPs) and its associated loss function for supervision \citep{garnot2021panoptic}.$^{2}$ {Averaging instance masks predicted by different modality-specific modules is not straightforward {and costly in terms of memory}, hence } 
we do not evaluate the decision-fusion scheme for this task.
 }

\section{Experiments}
We present in this section our numerical experiments to assess the benefit of multimodality for crop mapping with temporal attention-based networks. We evaluate several modality-fusion schemes and several mapping tasks. We also introduce a new large-scale open-access  and multimodal  dataset with annotations fit for all tasks.
%.................................
\subsection{Pastis-R}
%.................................
To evaluate the benefit of multimodality, we extend the open-access PASTIS dataset \citep{pastis} with corresponding Sentinel-1 observations.
PASTIS is composed of $2433$ time series of  multi-spectral patches sampled in four different regions of France. Each patch has a spatial extent of $1.28$km$\times1.28$km and contains all available Sentinel-2 observations for the $2019$ season for a total of $115$k images.
Note that PASTIS does not filter out  observations with high cloud cover, hence, certain patches can be partially or entirely obstructed by clouds.
%\VIVIEN{Note that PASTIS does not constraint the cloud cover of the patches, hence certain observations can be partially or totally obstructed by clouds. }

\begin{figure}
    \centering
    \includegraphics[width=\textwidth, trim=0cm 12.5cm 0cm 0cm,clip]{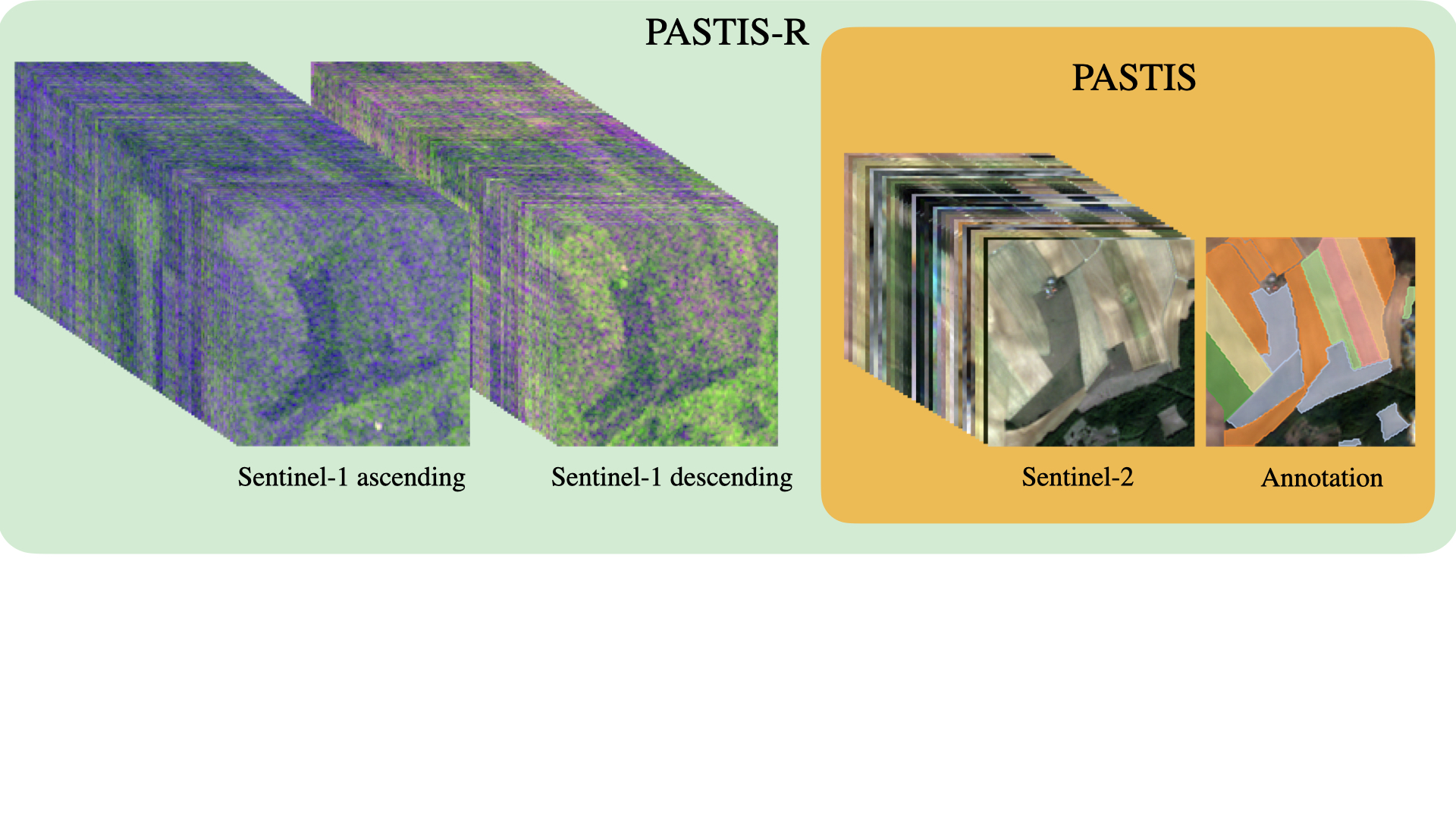}
    \caption{\textbf{{Pastis-R.}} We extend the PASTIS dataset with radar time series corresponding to ascending and descending orbits of Sentinel-1. For each square patch of $1.28$km$\times1.28$km, PASTIS-R thus provides the image time series of $3$ different modalities, along with semantic and instance annotation for each pixel.}
    \label{fig:pastis}
\end{figure}

We use Sentinel-1 in Ground Range Detected format processed into { $\sigma_0$ backscatter coefficient } in decibels, {orthorectified at a } $10$m spatial resolution with Orfeo Toolbox \citep{christophe2008orfeo}. We do not apply any spatial or temporal speckle filtering, {nor radiometric terrain correction}: {following the deep learning paradigm, we limit data preprocessing to the minimum.} We assemble each Sentinel-1 observation into a $3$-channel image: vertical polarization (VV), horizontal polarisation (VH), and the ratio of vertical over horizontal polarization (VV/VH). We separate observations made in ascending and descending orbit into two distinct time series. Indeed, the incidence angle of space-borne radar can significantly influence the return signal \citep{singhroy1999effects}. As represented in \figref{fig:pastis}, each time series comprises around $70$ radar acquisitions { for each of the $2433$ patches. This amounts to } a total of $339$k added radar images.
We use the annotations of PASTIS: semantic class and instance identifier for each pixel, allowing us to evaluate models for parcel-based classification, semantic segmentation, and panoptic segmentation. We make the PASTIS-R dataset \citep{pastis-r} publicly available at: \url{github.com/VSainteuf/pastis-benchmark} . 

%......................................
\subsection{Implementation details}
%......................................
As detailed in \secref{sec:implem}, we use the official PyTorch implementations of PSE+LTAE and U-TAE with default hyperparameters.
We  use the official $5$ cross-validation folds of PASTIS \citep{garnot2021utae} to evaluate the performance of the different models. We train our models using the Adam optimizer \citep{kingma2014adam} with default parameters $\text{lr}=0.001$, $\beta=(0.9, 0.999)$ unless specified otherwise, and train all networks on a single TESLA V100 GPU with $32$Gb of VRAM. 

\paragraph{\bf Multimodality Configuration}  We consider the two orbits of Sentinel-1 as separate modalities to account for their difference in incident angle, which corresponds to $M=3$. When using auxiliary loss terms, we set $\lambda^m=0.5$ for all modalities. When using temporal dropout, we set $p_0=0.4$ for the optical modality and $p_1=p_2=0.2$ for the radar time series. For early fusion, we interpolate the Sentinel-1 observations to the dates of the Sentinel-2 time series. Indeed, the opposite interpolation strategy would imply tripling the temporal length of the Sentinel-2 time series, which would significantly increase memory usage. Interpolation is computed on the fly when loading dataset samples. 

\paragraph{\bf Parcel Classification} For this problem, we train the models for $100$ epochs in batches of $128$ parcels. We use the $K=18$ class nomenclature of PASTIS and report the classification Intersection over Union macro-averaged over the class set (mIoU) to evaluate the parcel-level predictions. 

\paragraph{\bf Semantic Segmentation} We train the semantic segmentation models for $100$ epochs in batches of $4$ multi-temporal patches. In this setting, the models also predict \emph{background} pixels, resulting in a $K=19$ class nomenclature. We report the mIoU of the pixel-level predictions: 
\begin{align}
    \text{mIoU} = \frac{1}{K} \sum_{k=1}^{K} \frac{TP_k}{TP_k + FP_k + FN_k}~,
\end{align}
with $TP_k$,$FP_k$, and $FN_k$ the count of true positives, false positives, and false negatives for the binary class predictions defined by a class $k$.
\paragraph{\bf Panoptic Segmentation} We follow the training procedure recommended by \citet{garnot2021utae} to train the PaPs network: the learning rate starts at $0.01$ for the first $50$ epochs, and decreases to $0.001$ for the last $50$ epochs. We report the class-averaged panoptic metrics introduced in \citet{kirillov2019panoptic}: Segmentation Quality (SQ), Recognition Quality (RQ), and Panoptic Quality (PQ). 
The RQ corresponds to the $F_1$ score for the problem of combined detection and classification: to be counted as a true positive, a parcel must be both detected (the intersect over union of the predicted and true instance masks is above $0.5$) and have its crop type correctly classified.
The SQ corresponds to the intersect over union between the true and predicted masks for correctly detected and classified parcels. Finally, the PQ is the product of both values, thus simultaneously combining information on the quality of detection, classification, and delineation. We report the unweighted classwise average of the three quality measurements. We refer the reader to \citet{kirillov2019panoptic} for more details on these metrics.

%In panoptic segmentation, an agricultural parcel is considered correctly predicted if \LOIC{both} the spatial intersect over union of the predicted and true instance masks is above $0.5$ and if the predicted crop type is correct. 
%\LOIC{SQ corresponds to the classwise average of the intersect over union between the true and predicted masks for parcels correctly detected and classified  }. The PQ metric is the product of SQ and RQ, thus combining information on whether parcels are correctly detected and on the precision of the delineation of the parcel's shape. 

%.....................................................
\subsection{Parcel Classification Experiment}
%.....................................................
We first implement and evaluate the different fusion schemes and their training enhancements for parcel classification. In this setting, the contour of parcels is known in advance, and the model predicts the type of crop cultivated during the period covered by the SITS.

\begin{table}[ht!]
\caption{\textbf{Parcel Classification.} We evaluate the performance of models operating on a single modality (top) and for different fusion strategies for parcel-based classification (bottom). We evaluate each model's baseline performance and the impact of the temporal dropout and auxiliary classifiers enhancements, when applicable. We report the 5-fold cross-validated classification scores in terms of mean classwise Intersect over Union, the base model's parameter count, and, when relevant, of the model with auxiliary classifiers.}
\label{tab:xp:parcelbased}
\begin{tabular}{lccccccc}
\toprule
         & \multicolumn{2}{c}{\multirow{2}{*}{Base}} &\,& Temp. & Auxiliary   & Auxiliary \&     &      Parameter      \\
         &       &&                  & dropout  & supervision & Temp. dropout &      Count      \\  \cline{2-3} \cline{5-7}
         &OA&mIoU& \multicolumn{4}{c}{mIoU} \\
         \midrule
S2       & 91.7 &73.9                  && 74.5     & -           & -                &       114k    \\
S1D      & 87.0&64.5                  && 64.7     & -           & -                &       114k     \\
S1A      & 86.4&63.3                  && 62.9     & -           & -                &       114k    \\\midrule
Early Fusion  &91.8  &74.9                  && 76.5       & -           & -                &       117k     \\
Mid Fusion   & 92.0 &75.1                  && 75.9     &      75.0       & 76.5             &       152k/185k    \\
Late Fusion  & 91.1&73.0                  &&   73.6       & 76.1        & \textbf{77.2}             &       254k/287k    \\
Decision Fusion & 91.0 &72.5                  &&   72.8       & 75.2        & 75.8            &        259k    \\ \bottomrule
\end{tabular}
\end{table}

\begin{figure}
    \centering
    \includegraphics[width=.9\linewidth]{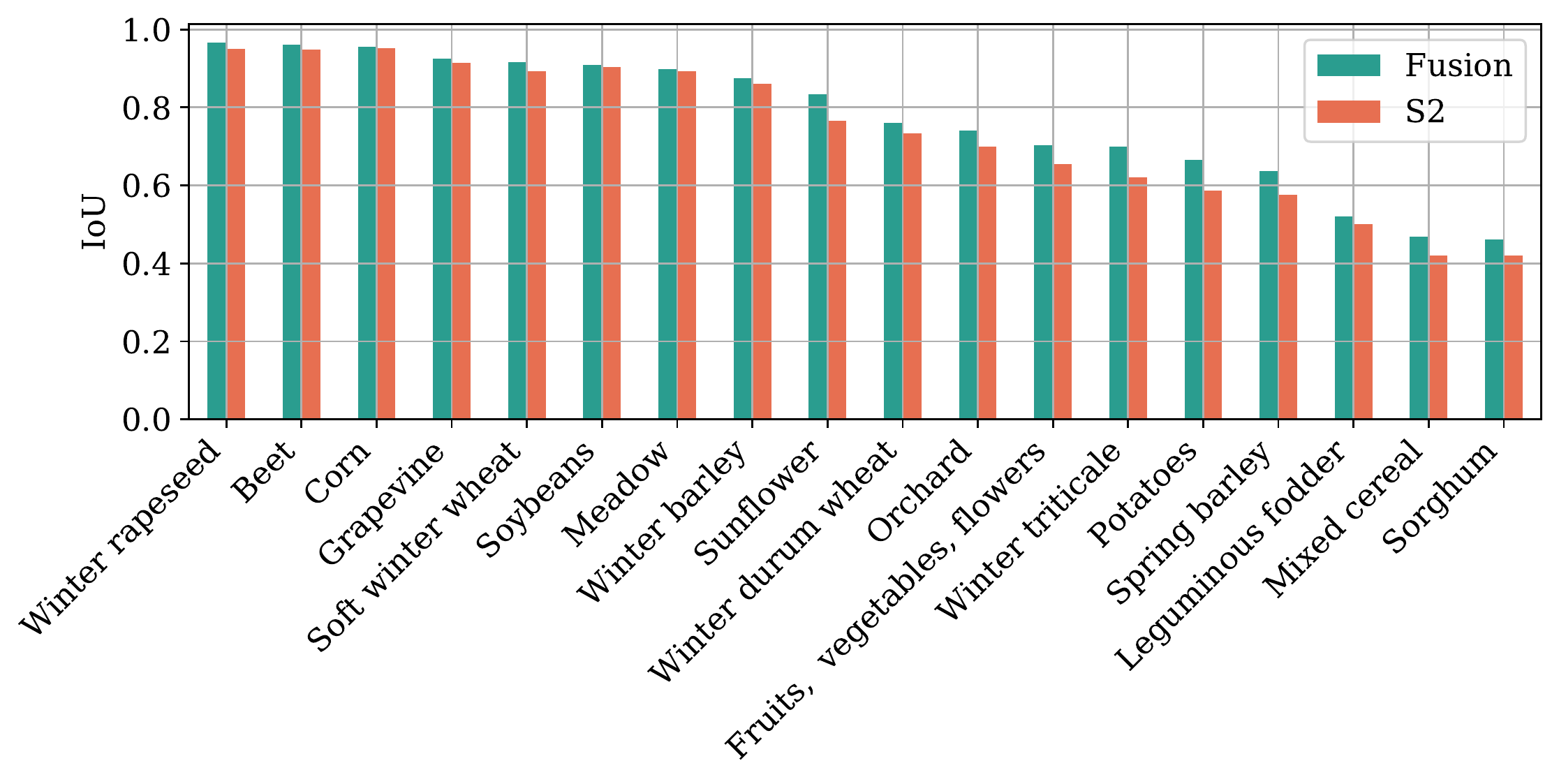}
    \caption{{\bf Classwise Performance for Parcel Classification.} We report the IoU of the late fusion model with auxiliary supervision and temporal dropout and {of} the model traine{d} purely on the optical modality. {Multimodality brings a consistent benefit across all classes, which is more notable for some of the most challenging classes such as \emph{Potatoes}, or \emph{Winter triticale}.}}
    \label{fig:perclass_parcel}
\end{figure}

\paragraph{\bf Analysis}
In \tabref{tab:xp:parcelbased}, we report the performance of all fusion schemes with and without  enhancements.
We first observe that the optical satellite S2 significantly outperforms the two radar time series by a margin of almost $10$ points of mIoU, confirming the relevance of Sentinel-2 for crop type mapping.
%We conclude that the spectral resolution of optical images contains crucial information for recognizing crop type.
We remark that, without enhancement (first column), multimodal models trained with early or mid-fusion schemes improve the performance {compared to} single optical modality network{s}, {while} decision and late fusion perform slightly worse consistently with the results of \citet{pelletier2021fusion}. This highlights the benefit of learning {to mix} modality features early on. 
In contrast, auxiliary supervision and temporal dropout improve the later models. This shows that these enhancements can encourage attention-based models to combine features and decisions efficiently from different modalities, as observed in \citet{ienco2019combining} for recurrent networks. 
All things considered, late fusion with both enhancements performs best with $+3.3$ mIoU compared to  {a network operating purely on the optical modality}, see \figref{fig:perclass_parcel} for {a classwise comparison}. {Mid-}fusion without enhancement provides good performances with a lower parameter count and none of the preprocessing necessary for early fusion. In practice, the {mid-fusion scheme} is $20$\% faster at inference time than late fusion, making it a valid choice when operating with limited computational resources.

%.....................................................
\paragraph{\bf Auxiliary Supervision and Gradient Flow}
%.....................................................
Motivated by the impact of auxiliary supervision on the performance of the late fusion approach, we propose to study its effect on the learning process further.  
Specifically, we wish to evaluate the different spatio-temporal encoders' contribution to the reduction of the objective loss $\Lmain$, with and without auxiliary supervision, and for the parcel classification task.
Note that, as auxiliary decisions are not computed at inference time, we only consider the decrease of $\Lmain$: a decrease in the auxiliary losses does not directly affect the model's performance.

Following the insights of  \citet{wang2020picking}, we consider the following first-order approximation of the decrease of $\Lmain$ incurred by taking a gradient step:
\begin{align}\label{eq:flow}
    \Delta \Lmain = \eta \langle \nabla \mathcal{L} , \nabla  \Lmain\rangle~,
\end{align}
with $\eta$ the current learning rate.The term $\nabla \mathcal{L}$ of the scalar product in \eqref{eq:flow} corresponds to the step size in the gradient descent and the term $\nabla  \Lmain$ to the slope of the objective loss. Their scalar product approximates the decrease in objective loss when taking a single gradient step. Note that this approximation, called gradient flow, is only valid when using stochastic gradient descent (SGD) and does not hold for momentum or adaptive optimization schemes {such as ADAM \citep{kingma2014adam}}. We thus retrain the late fusion model with SGD for parcel classification. By considering each term in the scalar product in \equaref{eq:flow}, we can estimate the contribution of each parameter of the network to the decrease of the objective loss $\Lmain$.
\begin{figure}[h]
    \centering
    \begin{subfigure}{\linewidth}
    \centering
    \includegraphics[width=\linewidth, trim=0cm 1cm 0cm 0cm, clip]{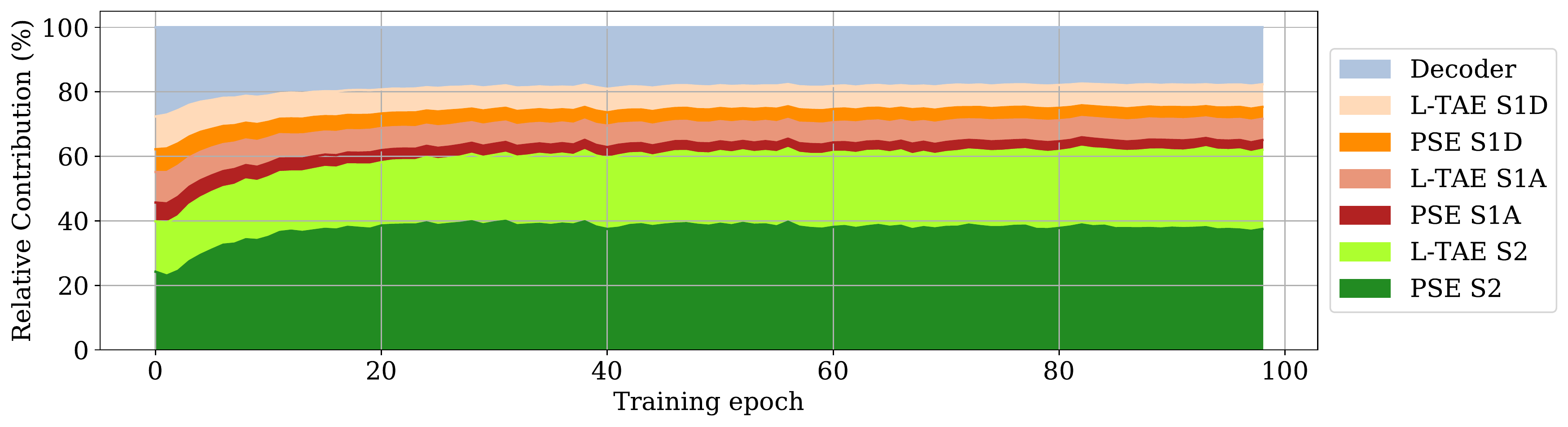}
    \caption{Training without auxiliary supervision: $\mathcal{L} = \Lmain$~.}
    \label{fig:gflow:base}
    \end{subfigure}
    \vfill
    \begin{subfigure}{\linewidth}
    \centering
    \includegraphics[width=\linewidth]{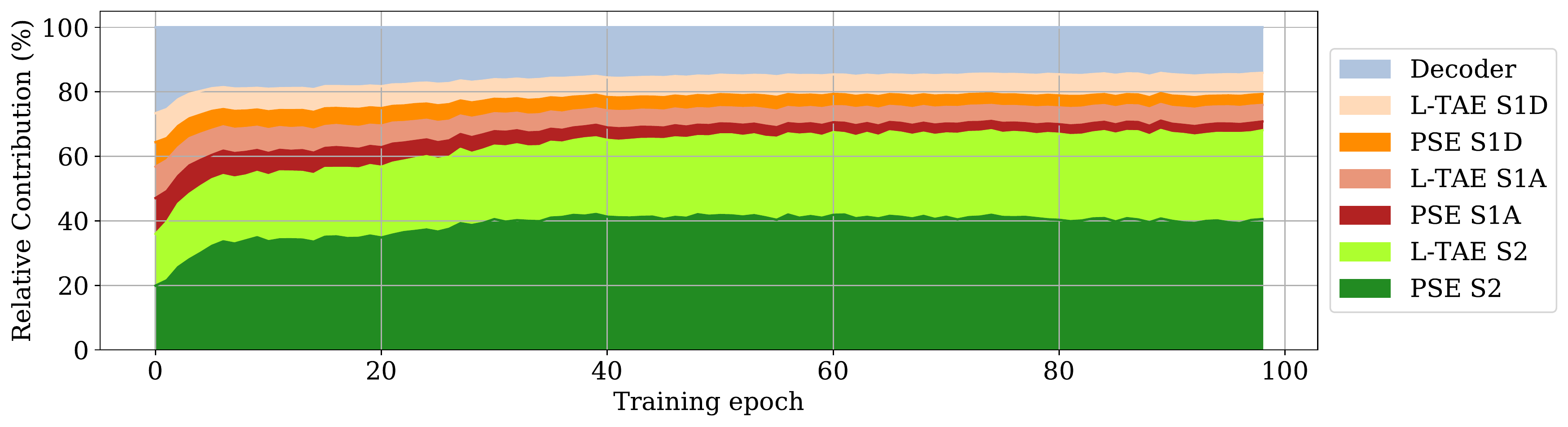}
    \caption{Training with auxiliary supervision: $\mathcal{L} = \Lmain + \Laux$~.}
    \label{fig:gflow:aux}
    \end{subfigure}
    \caption{{\bf Gradient Flow.} Evolution of the gradient flow for different modules of the late fusion model. The contribution of each modality is plotted as a fraction of the total flow, without auxiliary loss terms (top) and with the additional $\Laux$ term (bottom). We report the flow for the spatial encoders (PSE), temporal encoders (LTAE), and the MLP-based decoders.}
    \label{fig:gflow}
\end{figure}

In \figref{fig:gflow}, we represent the evolution of the gradient flow for different modules of our architecture by summing the contribution of their corresponding parameters.
We observe that, as expected, the gradient flow is concentrated in the modules dedicated to the optical modality. Interestingly, the spatial encoders contribute as much or even more than the temporal encoders despite having four times fewer parameters.

We remark that auxiliary losses lead the model to a different training regime. 
While auxiliary supervision results in an increase of the proportion of gradient flow in some radar modules such as PSE-S1A, the flow also increases in proportion in some optical modules as well. We conclude that auxiliary supervision affects all modalities, not only the weaker modalities.

%................................................
\subsection{Semantic Segmentation}
%................................................
In this section, we evaluate the performance of the late fusion scheme compared to single modality baselines for semantic segmentation. While the  {mid-fusion} scheme yields promising results on parcel-based experiments, its implementation into a semantic segmentation architecture is not trivial. Indeed,  the state-of-the-art network for this task \citep{garnot2021utae} relies on a U-Net architecture with temporal encoding. In this architecture, spatial and temporal encoding are performed conjointly. After several unsuccessful attempts, we limit our study to the other fusion schemes for this task.

\begin{table}[h]
\caption{\textbf{Semantic Segmentation Experiment}. We evaluate the  semantic segmentation performance of models operating on a single modality and multimodal models trained with early, late, and decision fusion strategies. We evaluate each model's baseline performance and the impact of temporal dropout and auxiliary classifiers, when applicable. We report the 5-fold cross-validated classification scores in mean classwise Intersect over Union (- not applicable). Note that temporal dropout is necessary for the late and decision fusion models to fit in memory.}
\label{tab:semantic}
\begin{center}
\begin{tabular}{lcccc}
\toprule
         & \multirow{2}{*}{Base} & Temporal   & Auxiliary \&     &      Parameter      \\
         &                       & dropout  & Temporal dropout &      Count      \\ \midrule
S2       & 63.1                  & 63.6            & -                &       1\;087k   \\
S1D      &     54.9             &   54.7           & -                &      1\;083k     \\
S1A      & 53.8                  & 53.3          & -                &       1\;083k    \\
Early Fusion        & 64.9        & 65.8         & -     & 1\;602k\\
Late Fusion  & -                 &    65.8          & \textbf{66.3}             &       1\;709k     \\
Decision Fusion  &-                &   64.7          & 64.3          &       1\;742k     \\
\bottomrule
\end{tabular}
\end{center}
\end{table}

\begin{figure}[th!]
    \centering
\begin{tabular}{c}
    \begin{tabular}{rlrlrlrl}
 \definecolor{tempcolor}{rgb}{0,0,0}
           \tikz \fill[fill=tempcolor, scale=0.3, draw=black] (0,0) rectangle (1,1);
           & \footnotesize{Background} 
           &
           \definecolor{tempcolor}{rgb}{0.6823529411764706, 0.7803921568627451, 0.9098039215686274}
           \tikz \fill[fill=tempcolor, scale=0.3, draw=black] (0,0) rectangle (1,1); 
           & \footnotesize{Meadow}
           &
           \definecolor{tempcolor}{rgb}{1.0, 0.4980392156862745, 0.054901960784313725}
           \tikz \fill[fill=tempcolor, scale=0.3, draw=black] (0,0) rectangle (1,1); 
           & \footnotesize{Soft W. wheat}
           &
           \definecolor{tempcolor}{rgb}{1.0, 0.7333333333333333, 0.47058823529411764}
           \tikz \fill[fill=tempcolor, scale=0.3, draw=black] (0,0) rectangle (1,1);
           & \footnotesize{Corn}
           \\
           \definecolor{tempcolor}{rgb}{0.17254901960784313, 0.6274509803921569, 0.17254901960784313}
           \tikz \fill[fill=tempcolor, scale=0.3, draw=black] (0,0) rectangle (1,1);
           & \footnotesize{W. barley} 
           &
           \definecolor{tempcolor}{rgb}{0.596078431372549, 0.8745098039215686, 0.5411764705882353}
           \tikz \fill[fill=tempcolor, scale=0.3, draw=black] (0,0) rectangle (1,1); 
           & \footnotesize{W. rapeseed}
           &
           \definecolor{tempcolor}{rgb}{0.8392156862745098, 0.15294117647058825, 0.1568627450980392}
           \tikz \fill[fill=tempcolor, scale=0.3, draw=black] (0,0) rectangle (1,1); 
           & \footnotesize{Spring barley}
           &
           \definecolor{tempcolor}{rgb}{1.0, 0.596078431372549, 0.5882352941176471}
           \tikz \fill[fill=tempcolor, scale=0.3, draw=black] (0,0) rectangle (1,1);
           & \footnotesize{Sunflower}
           \\
           \definecolor{tempcolor}{rgb}{0.5803921568627451, 0.403921568627451, 0.7411764705882353}
           \tikz \fill[fill=tempcolor, scale=0.3, draw=black] (0,0) rectangle (1,1);
           & \footnotesize{Grapevine} 
           &
           \definecolor{tempcolor}{rgb}{0.7725490196078432, 0.6901960784313725, 0.8352941176470589}
           \tikz \fill[fill=tempcolor, scale=0.3, draw=black] (0,0) rectangle (1,1); 
           & \footnotesize{Beet}
           &
           \definecolor{tempcolor}{rgb}{0.5490196078431373, 0.33725490196078434, 0.29411764705882354}
           \tikz \fill[fill=tempcolor, scale=0.3, draw=black] (0,0) rectangle (1,1); 
           & \footnotesize{W. triticale}
           &
           \definecolor{tempcolor}{rgb}{0.7686274509803922, 0.611764705882353, 0.5803921568627451}
           \tikz \fill[fill=tempcolor, scale=0.3, draw=black] (0,0) rectangle (1,1);
           & \footnotesize{W. durum wheat}
           \\
           \definecolor{tempcolor}{rgb}{0.8901960784313725, 0.4666666666666667, 0.7607843137254902}
           \tikz \fill[fill=tempcolor, scale=0.3, draw=black] (0,0) rectangle (1,1);
           & \footnotesize{Fruits, veg., flow.} 
           &
           \definecolor{tempcolor}{rgb}{0.9686274509803922, 0.7137254901960784, 0.8235294117647058}
           \tikz \fill[fill=tempcolor, scale=0.3, draw=black] (0,0) rectangle (1,1); 
           & \footnotesize{Potatoes}
           &
           \definecolor{tempcolor}{rgb}{0.4980392156862745, 0.4980392156862745, 0.4980392156862745}
           \tikz \fill[fill=tempcolor, scale=0.3, draw=black] (0,0) rectangle (1,1); 
           & \footnotesize{Leguminous fodder}
           &
           \definecolor{tempcolor}{rgb}{0.7803921568627451, 0.7803921568627451, 0.7803921568627451}
           \tikz \fill[fill=tempcolor, scale=0.3, draw=black] (0,0) rectangle (1,1);
           & \footnotesize{Soybeans}
           \\
           \definecolor{tempcolor}{rgb}{0.7372549019607844, 0.7411764705882353, 0.13333333333333333}
           \tikz \fill[fill=tempcolor, scale=0.3, draw=black] (0,0) rectangle (1,1);
           & \footnotesize{Orchard} 
           &
           \definecolor{tempcolor}{rgb}{0.8588235294117647, 0.8588235294117647, 0.5529411764705883}
           \tikz \fill[fill=tempcolor, scale=0.3, draw=black] (0,0) rectangle (1,1); 
           & \footnotesize{Mixed cereal}
           &
           \definecolor{tempcolor}{rgb}{0.09019607843137255, 0.7450980392156863, 0.81176470588235291}
           \tikz \fill[fill=tempcolor, scale=0.3, draw=black] (0,0) rectangle (1,1); 
           & \footnotesize{Sorghum}
           &
           \definecolor{tempcolor}{rgb}{1,1,1}
           \tikz \fill[fill=tempcolor, scale=0.3, draw=black] (0,0) rectangle (1,1);
           & \footnotesize{Void label}
    \end{tabular}
 \\
\begin{tabular}{ccccc}
    \includegraphics[width=.18\textwidth, trim=1cm 0.5cm 0cm 0cm, clip]{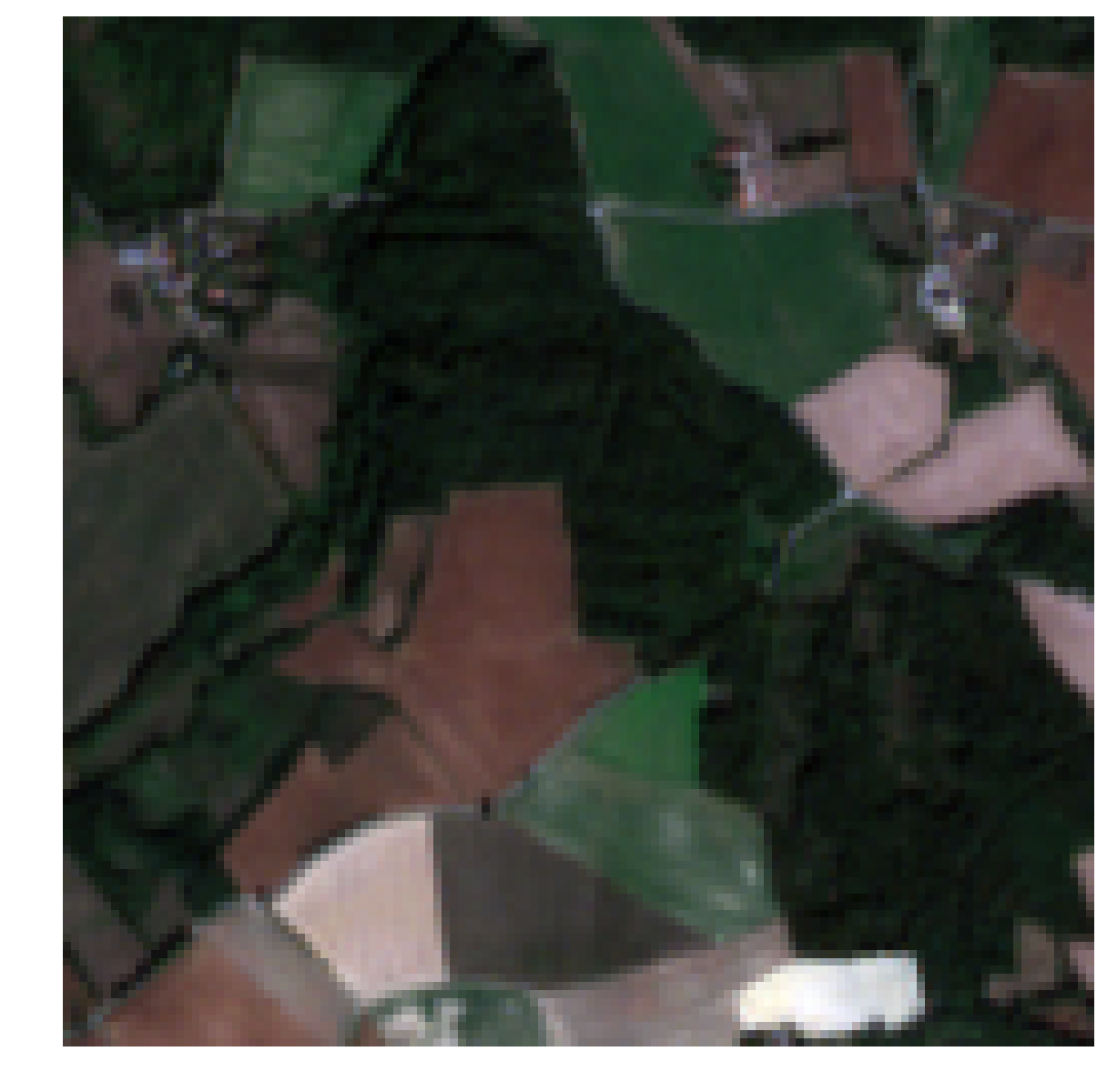}
     &
    \includegraphics[width=.18\textwidth, trim=1cm 0.5cm 0cm 0cm, clip]{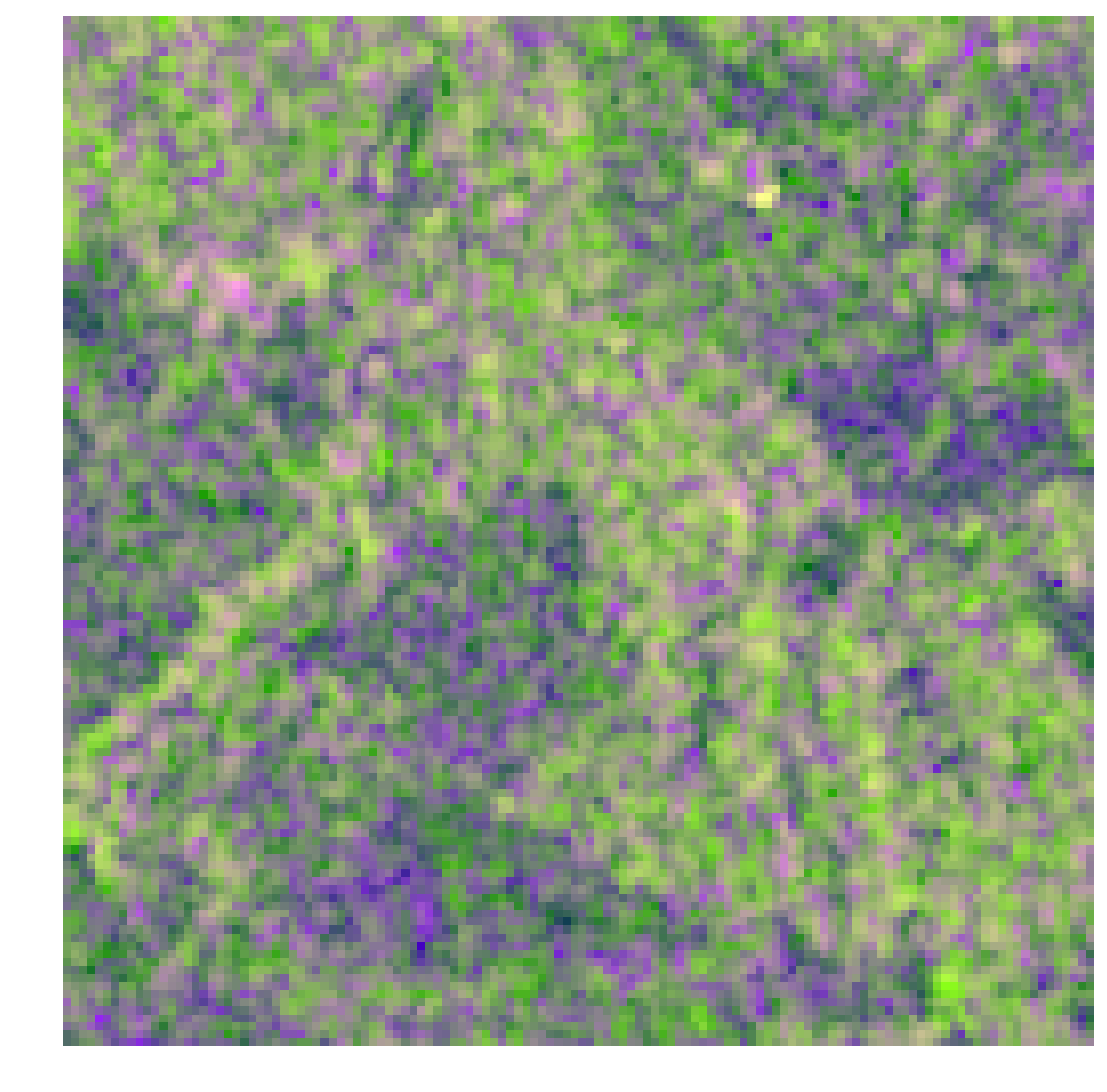}
     &
     \includegraphics[width=.18\textwidth, trim=1cm 0.5cm 0cm 0cm, clip]{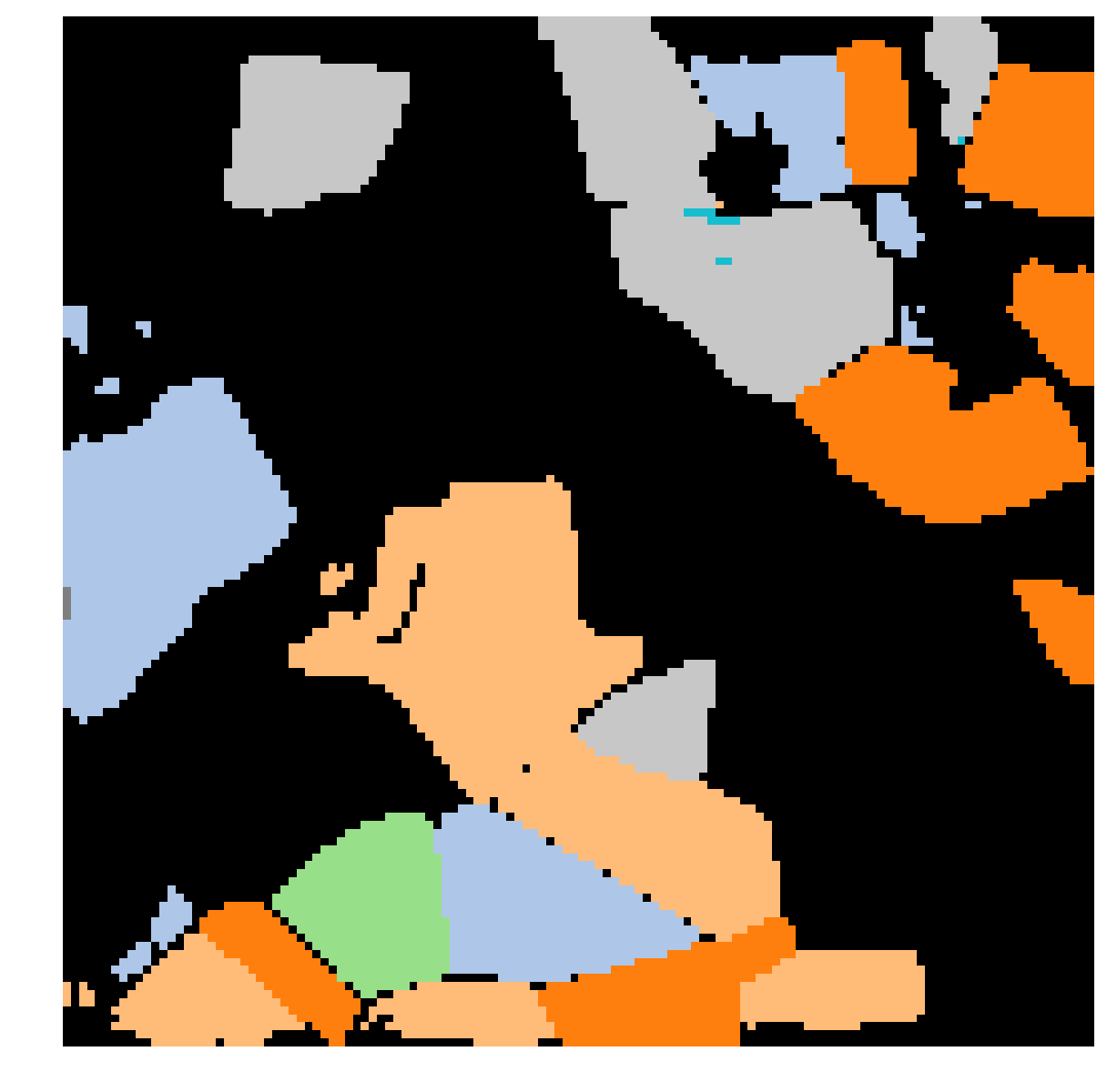}
     & 
     \begin{tikzpicture}
        \node[anchor=south west,inner sep=0] (image) at (0,0) {  \includegraphics[width=.18\textwidth, trim=1cm 0.5cm 0cm 0cm, clip]{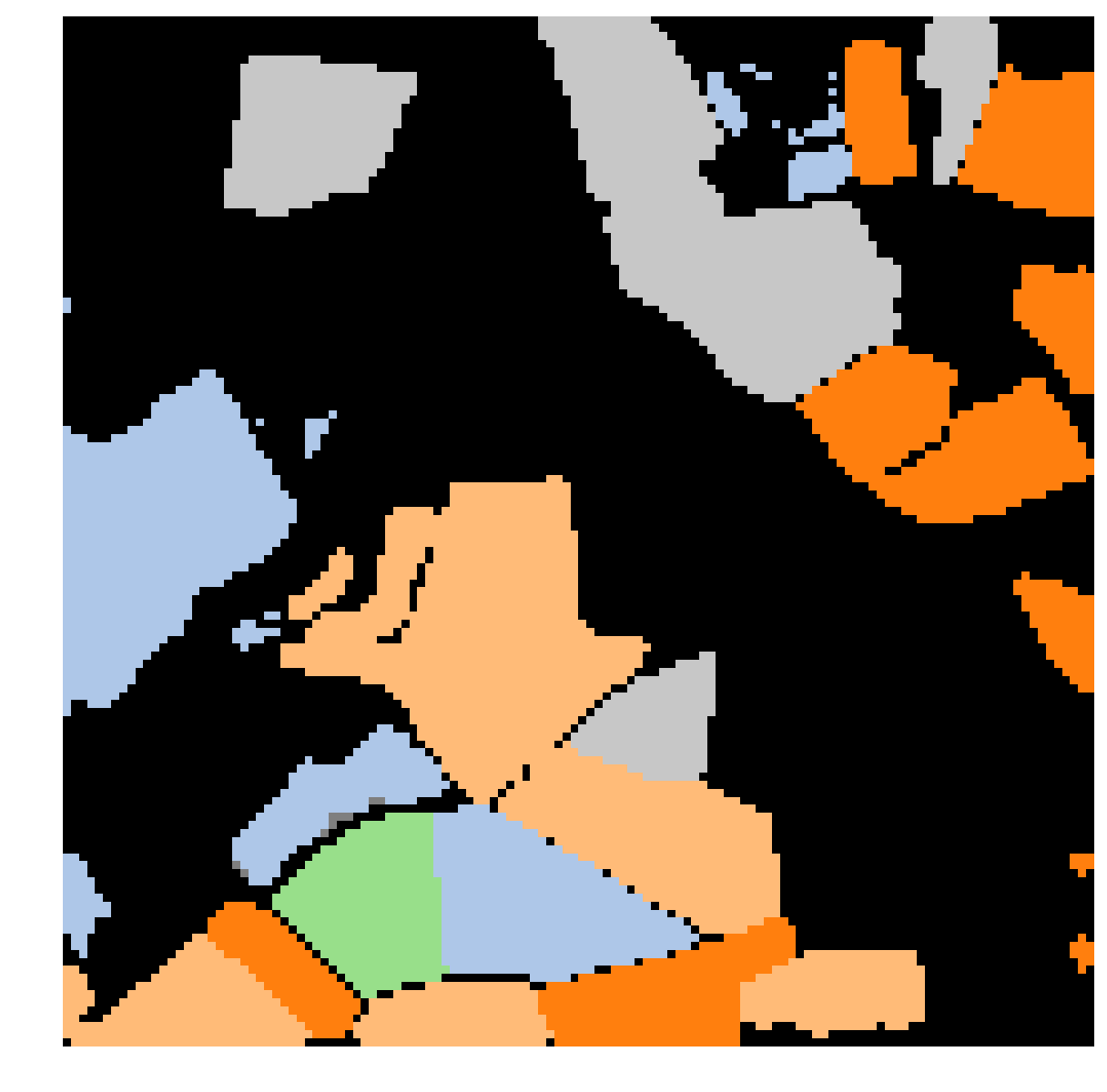}};
        \begin{scope}[x={(image.south east)},y={(image.north west)}]
            \draw[cyan,ultra thick] (0.52,0.34) circle (0.12);
        \end{scope}
     \end{tikzpicture}
  
     &
    \includegraphics[width=.18\textwidth, trim=1cm 0.5cm 0cm 0cm, clip]{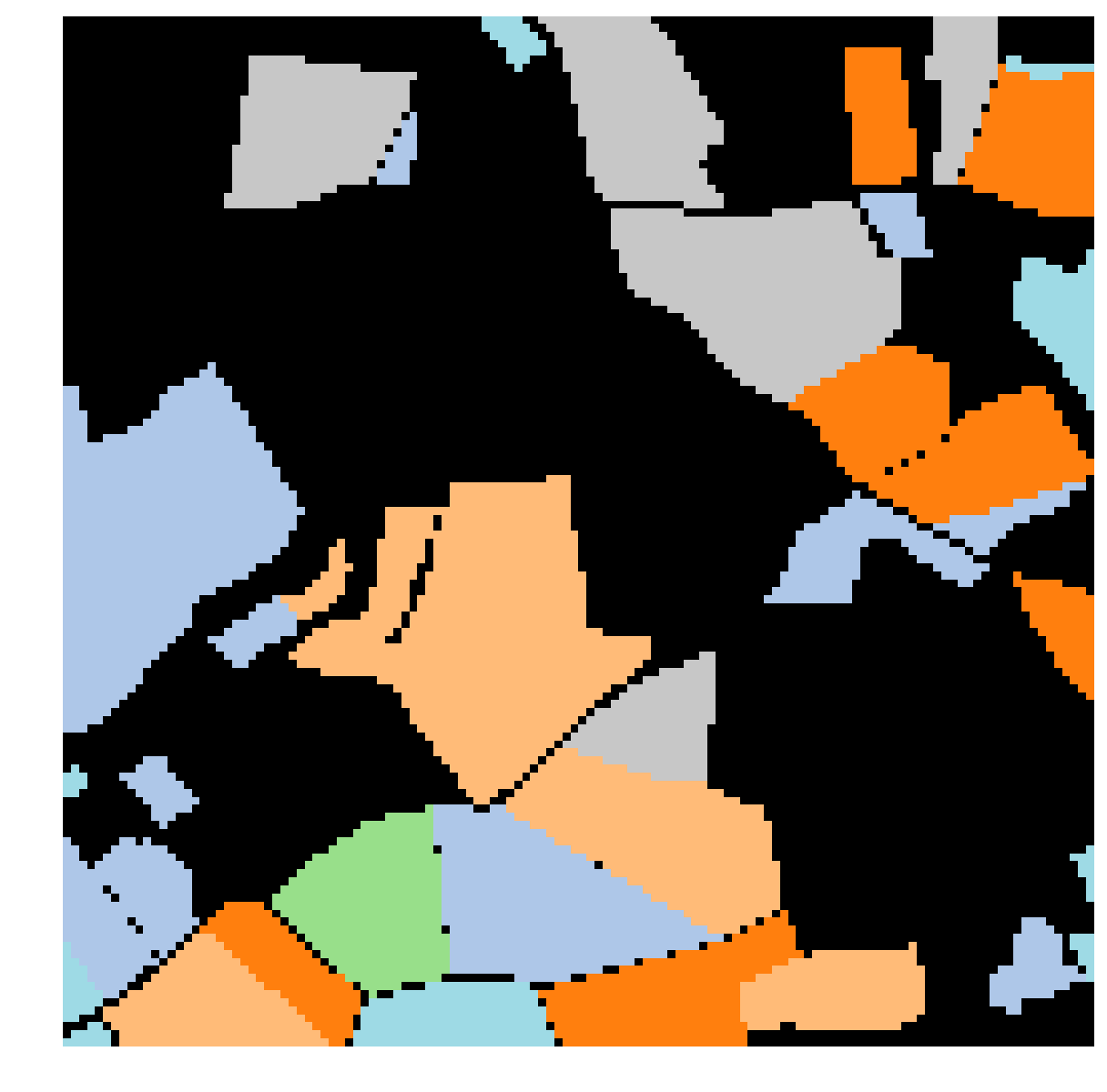}
     \\
     \includegraphics[width=.18\textwidth, trim=1cm 0.5cm 0cm 0cm, clip]{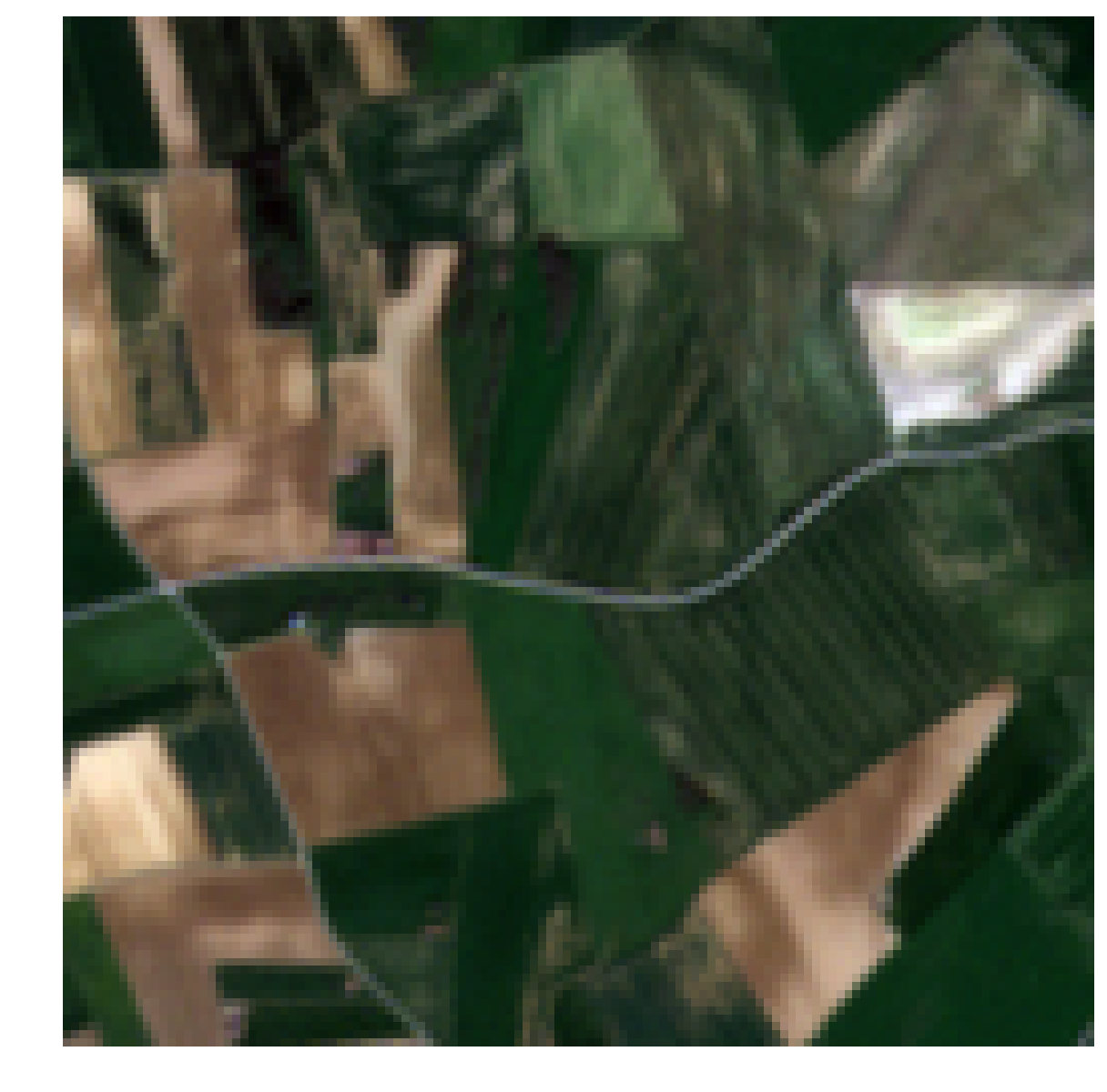}
     & 
    \includegraphics[width=.18\textwidth, trim=1cm 0.5cm 0cm 0cm, clip]{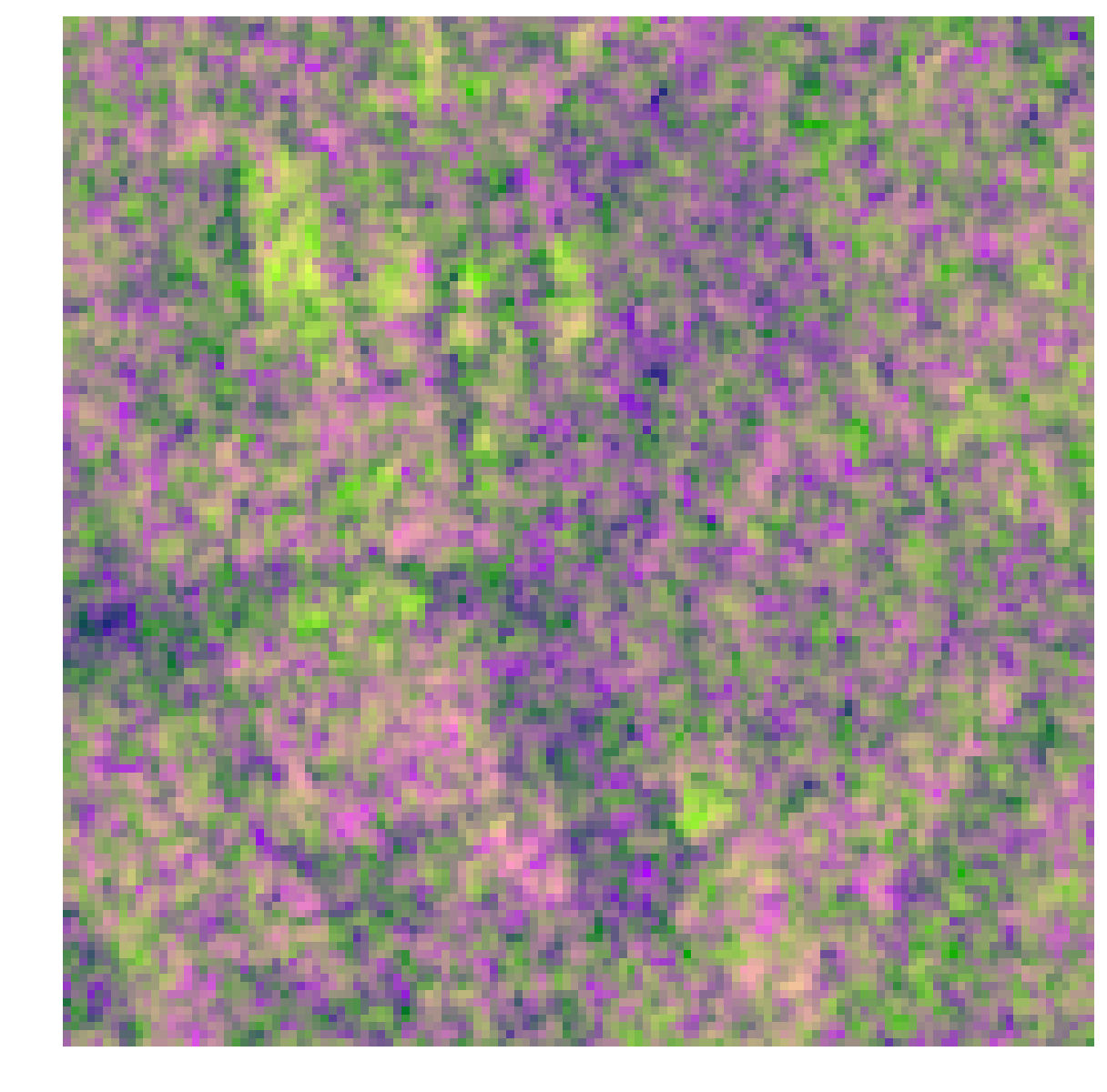}
     &
    \includegraphics[width=.18\textwidth, trim=1cm 0.5cm 0cm 0cm, clip]{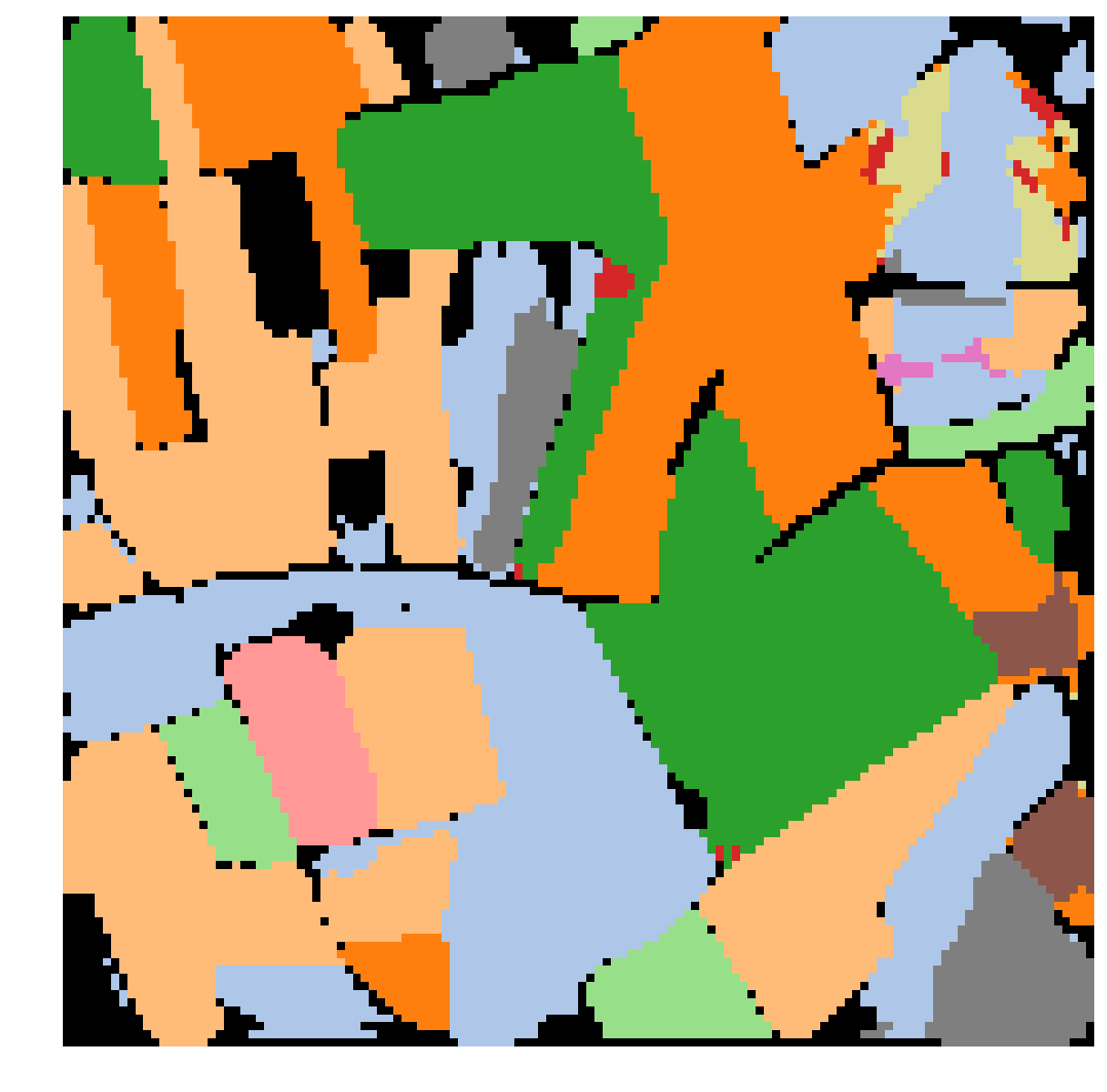}
     &
     \begin{tikzpicture}
        \node[anchor=south west,inner sep=0] (image) at (0,0) {   \includegraphics[width=.18\textwidth, trim=1cm 0.5cm 0cm 0cm, clip]{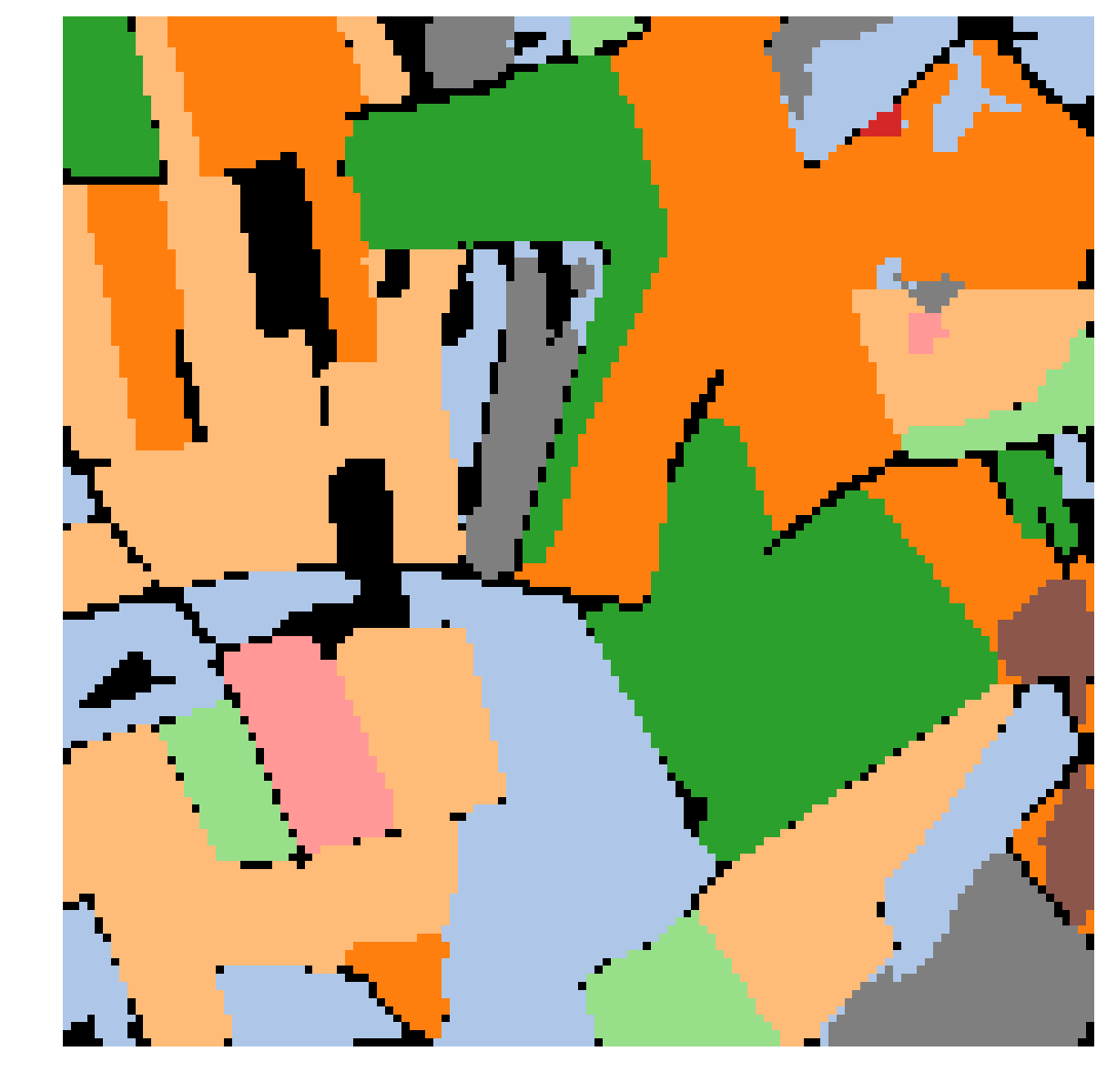}};
        \begin{scope}[x={(image.south east)},y={(image.north west)}]
            \draw[magenta,ultra thick] (0.85,0.85) circle (0.12);
            \draw[cyan,ultra thick] (0.25,0.3) circle (0.10);
        \end{scope}
     \end{tikzpicture}
     & 
    \includegraphics[width=.18\textwidth, trim=1cm 0.5cm 0cm 0cm, clip]{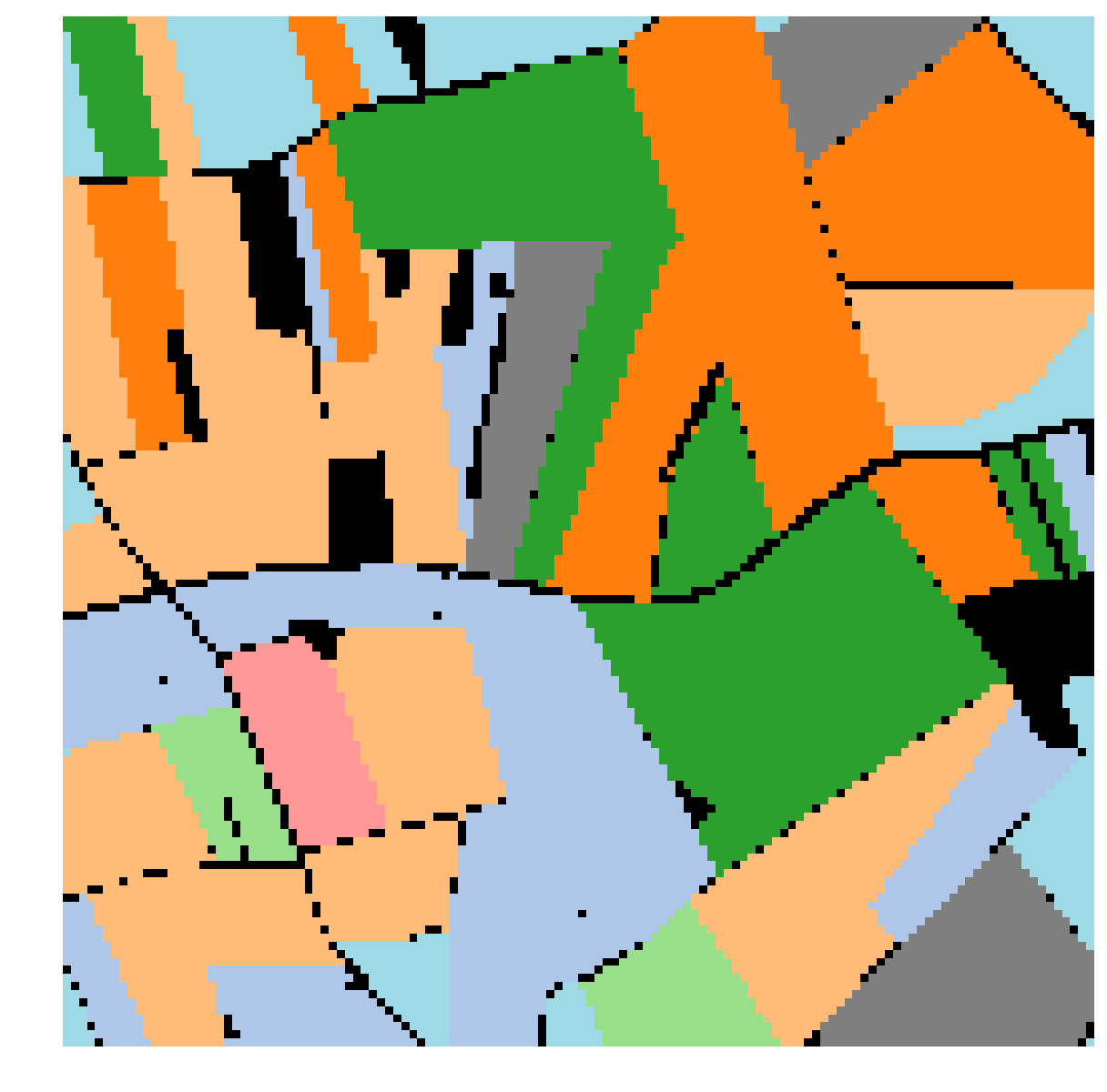}
     \\
    \includegraphics[width=.18\textwidth, trim=1cm 0.5cm 0cm 0cm, clip]{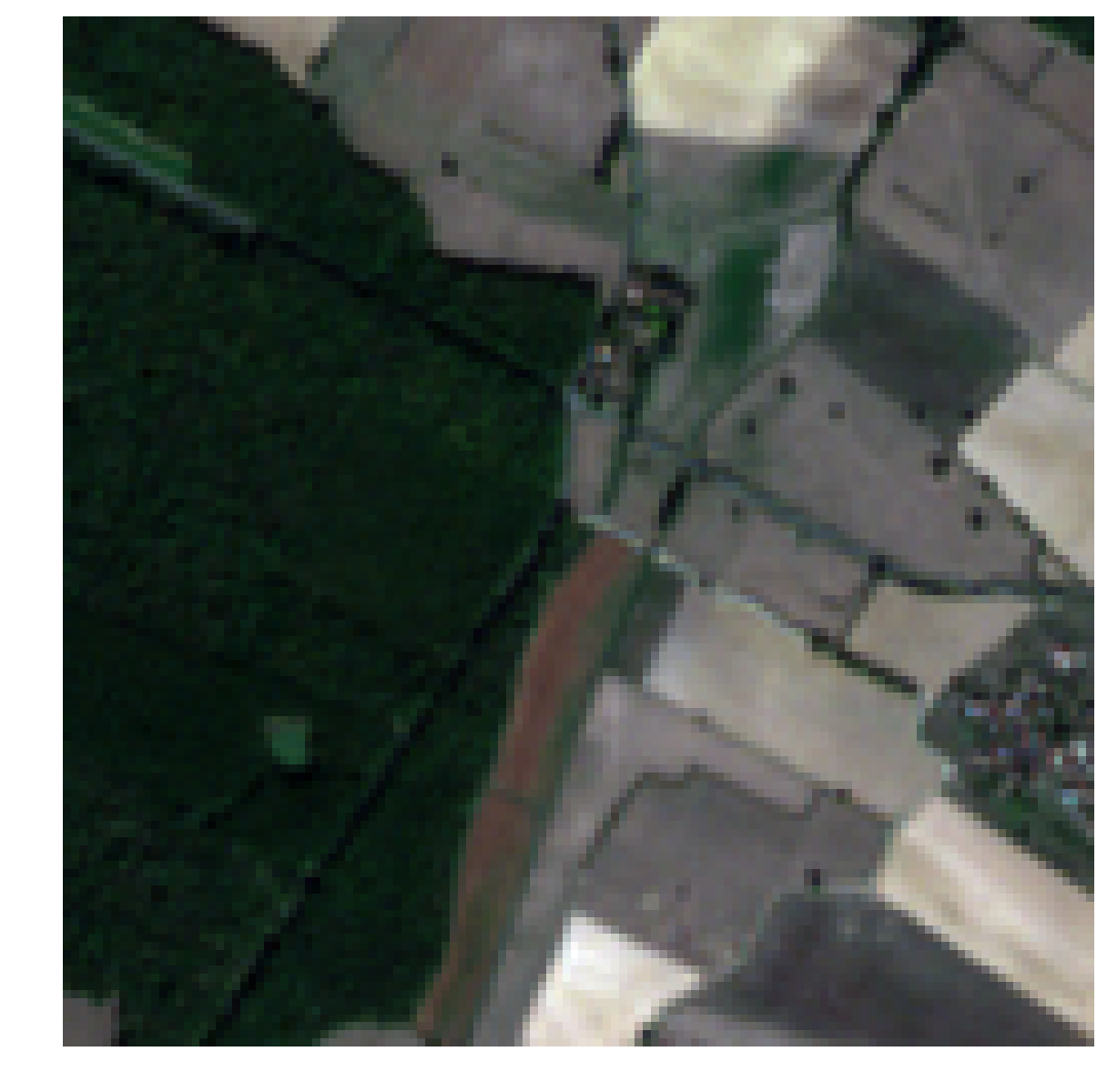}
     & 
    \includegraphics[width=.18\textwidth, trim=1cm 0.5cm 0cm 0cm, clip]{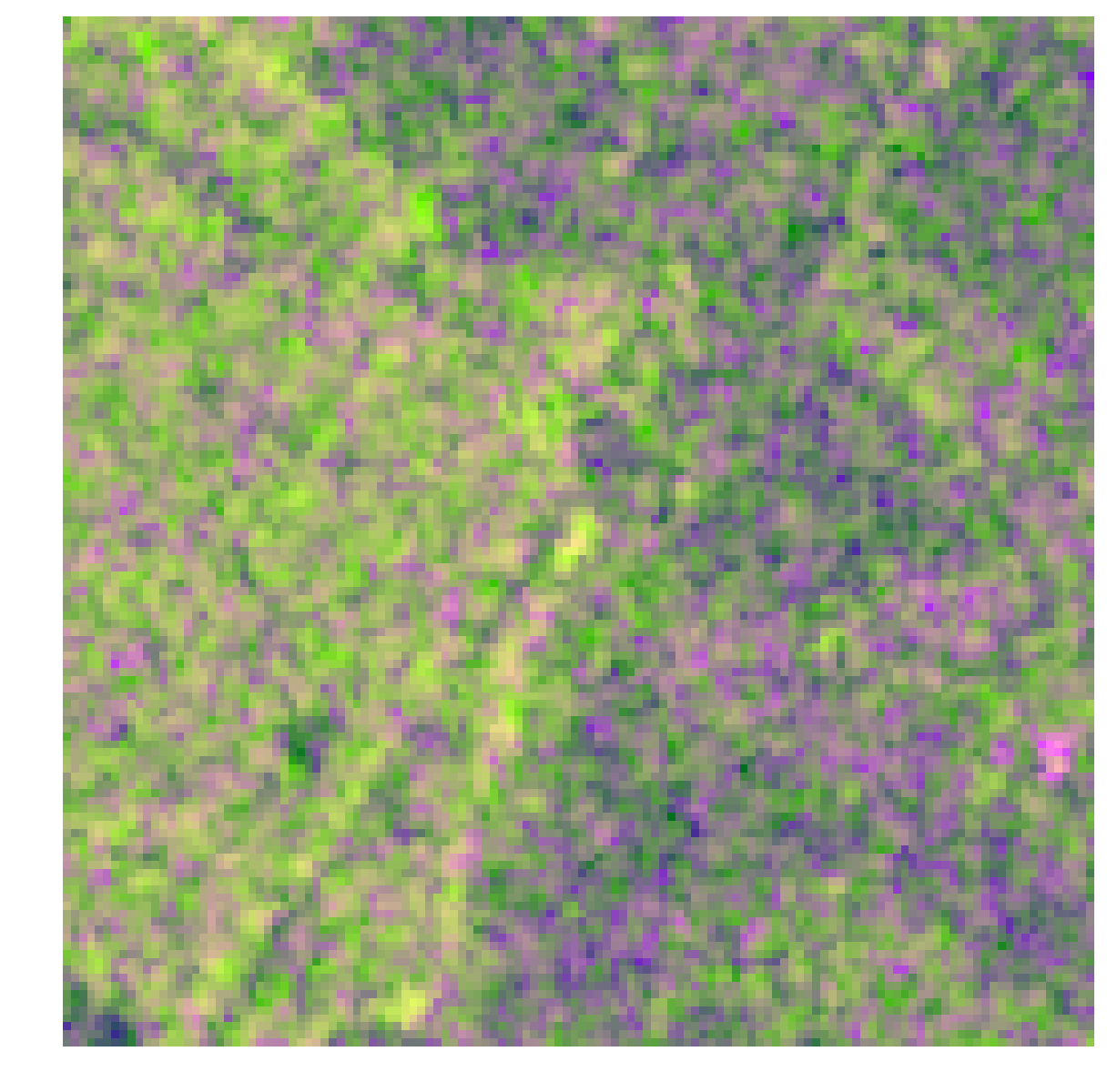}
     &
    \includegraphics[width=.18\textwidth, trim=1cm 0.5cm 0cm 0cm, clip]{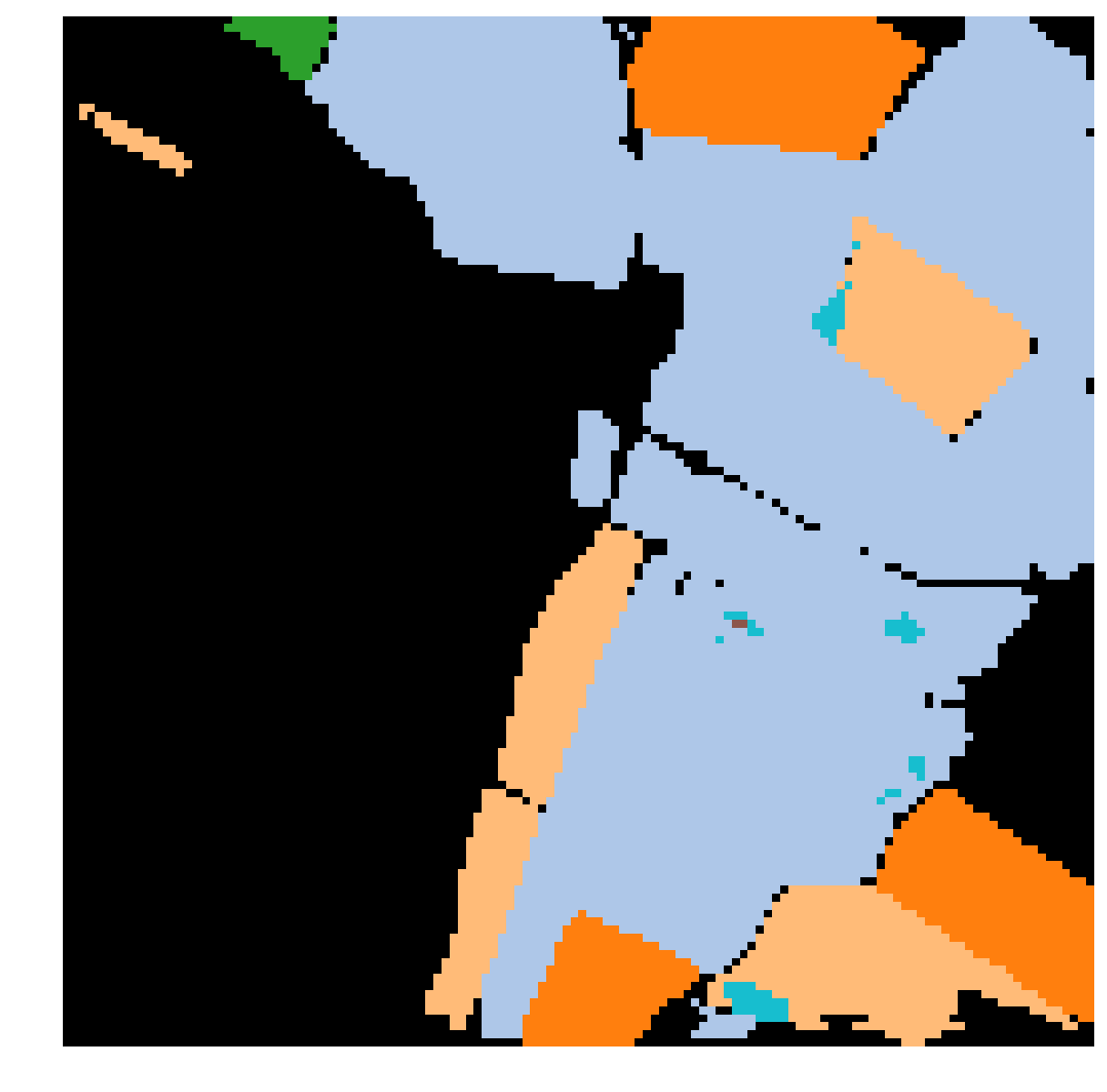}
     &
      \begin{tikzpicture}
        \node[anchor=south west,inner sep=0] (image) at (0,0) {       \includegraphics[width=.18\textwidth, trim=1cm 0.5cm 0cm 0cm, clip]{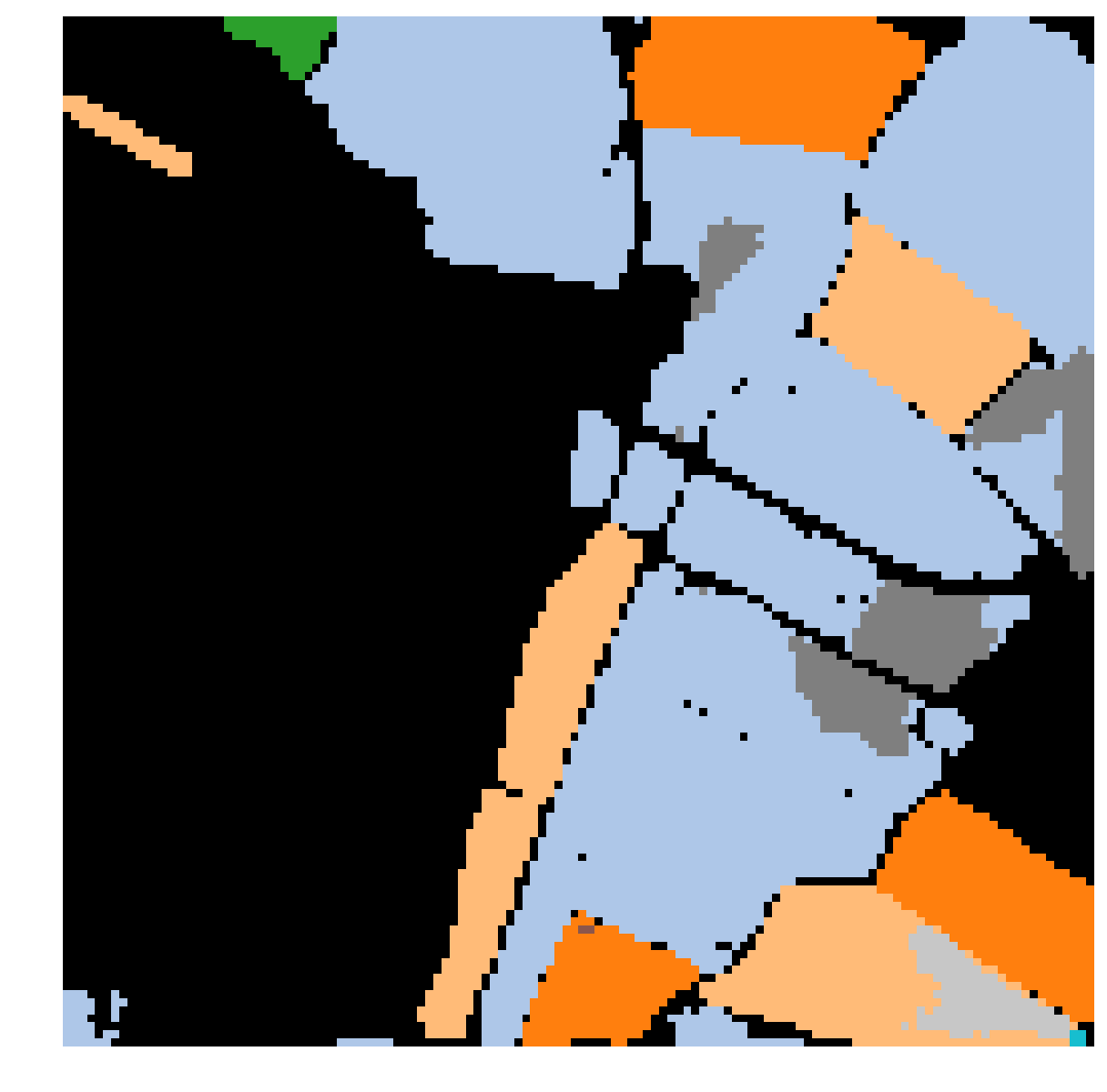}};
        \begin{scope}[x={(image.south east)},y={(image.north west)}]
            \draw[magenta,ultra thick] (0.82,0.43) circle (0.08);
            \draw[cyan,ultra thick] (0.66,0.45) circle (0.08);
        \end{scope}
     \end{tikzpicture}
     & 
    \includegraphics[width=.18\textwidth, trim=1cm 0.5cm 0cm 0cm, clip]{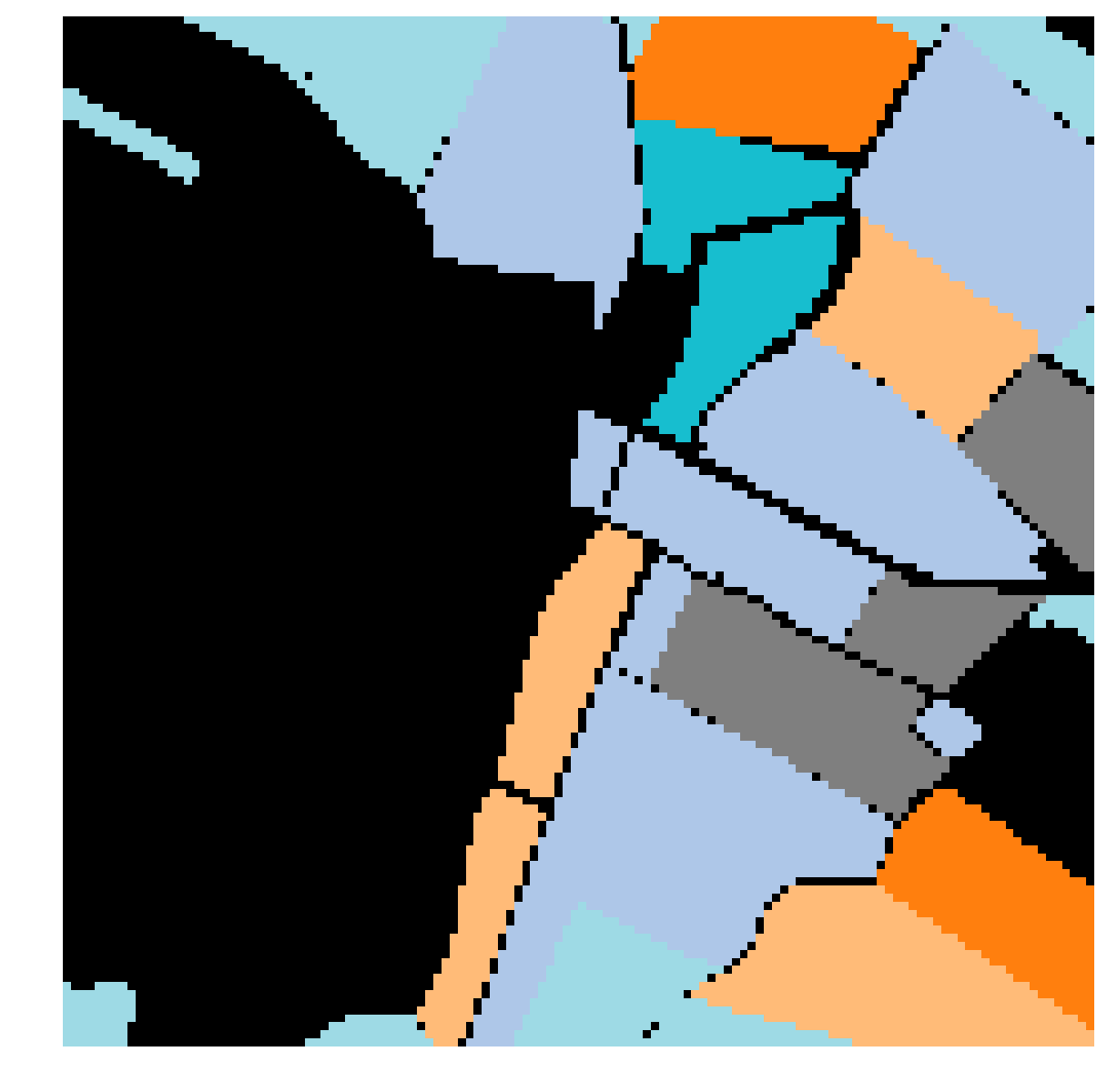}
     \\
     \begin{subfigure}{.19\textwidth}
    \caption{Optical}
    \label{fig:qualisem:s2}
    \end{subfigure}
    &
    \begin{subfigure}{.19\textwidth}
    \caption{Radar}
    \label{fig:qualisem:s1}
    \end{subfigure}

    &
    \begin{subfigure}{.19\textwidth}
    \caption{S2 Prediction}
    \label{fig:qualisem:mono}
    \end{subfigure}
    &
    \begin{subfigure}{.19\textwidth}
    \caption{Fusion Prediction}
    \label{fig:qualisem:late}
    \end{subfigure}
    &
     \begin{subfigure}{.19\textwidth}
    \caption{Ground Truth}
    \label{fig:qualisem:gt}
    \end{subfigure}
    \end{tabular}
  \end{tabular}
    \caption{{\bf Qualitative Results for Semantic Segmentation.} We show one observation from the optical time series in \Subfigref{fig:qualisem:s2} and {one} from the radar time series in \Subfigref{fig:qualisem:s1}. The prediction for the unimodal optical model is represented in \Subfigref{fig:qualisem:mono}, a{our late fusion multimodal model} in \Subfigref{fig:qualisem:late}, and finally the ground truth in \Subfigref{fig:qualisem:gt}. We observe that the multimodal model produces results with clearer and more distinct borders between close parcels (cyan circle \protect\tikz \protect\node[circle, thick, draw = cyan, fill = none, scale = 0.7] {};). The multimodal model also displays fewer errors for hard and ambiguous parcels, showing the benefit of learning intermodal features (magenta circle \protect\tikz \protect\node[circle, thick, draw = magenta, fill = none, scale = 0.7] {};). Crop types are represented according to the color code above (W. stands for Winter). The same legend is {used} in all subsequent figures representing crop labels. }
    \label{fig:qualisem}
\end{figure}

\paragraph{\bf Analysis} We report the performance of the different models in \tabref{tab:semantic}. In our experimental setup, the late fusion model with {over} $\sim200$ total {multimodal} observations did not fit in {the $32$Gb of memory of our GPU} with a batch size of $4$ image time series. {By reducing the size of the input sequences, temporal dropout allowed us to  train this memory-intensive model.}
The late fusion model improves the performance of the unimodal models by $2.7$ mIoU points. The performance is further improved by {another} $0.5$ point with the addition of auxiliary supervision. {The early fusion model performs slightly below late fusion, even with temporal dropout. }
As represented in \figref{fig:qualisem}, the radar {modality allows for prediction with crisper contours, in particular between adjacent or nearly adjacent parcels}. This suggests that {the image rugosity}  {of the radar} acquisitions is {can be} valuable to detect inter-parcel zones. {These areas, often of sub-pixel extent,} may display optical reflectances {similar to their} neighboring parcels {but often present surfaces such as fences or groves with a volumetric scatter and thus a distinct radar response . }

Note that the performance of our models on semantic segmentation is around $10$pts below that for parcel classification. This result was expected as {the semantic segmentation task prevents us from exploiting knowledge about the contour of parcels and has the supplementary class \emph{background}, corresponding to non-agricultural land.}

\begin{figure}[h!]
    \centering
    \begin{subfigure}{\linewidth}
    \centering
    \includegraphics[width=.8\linewidth]{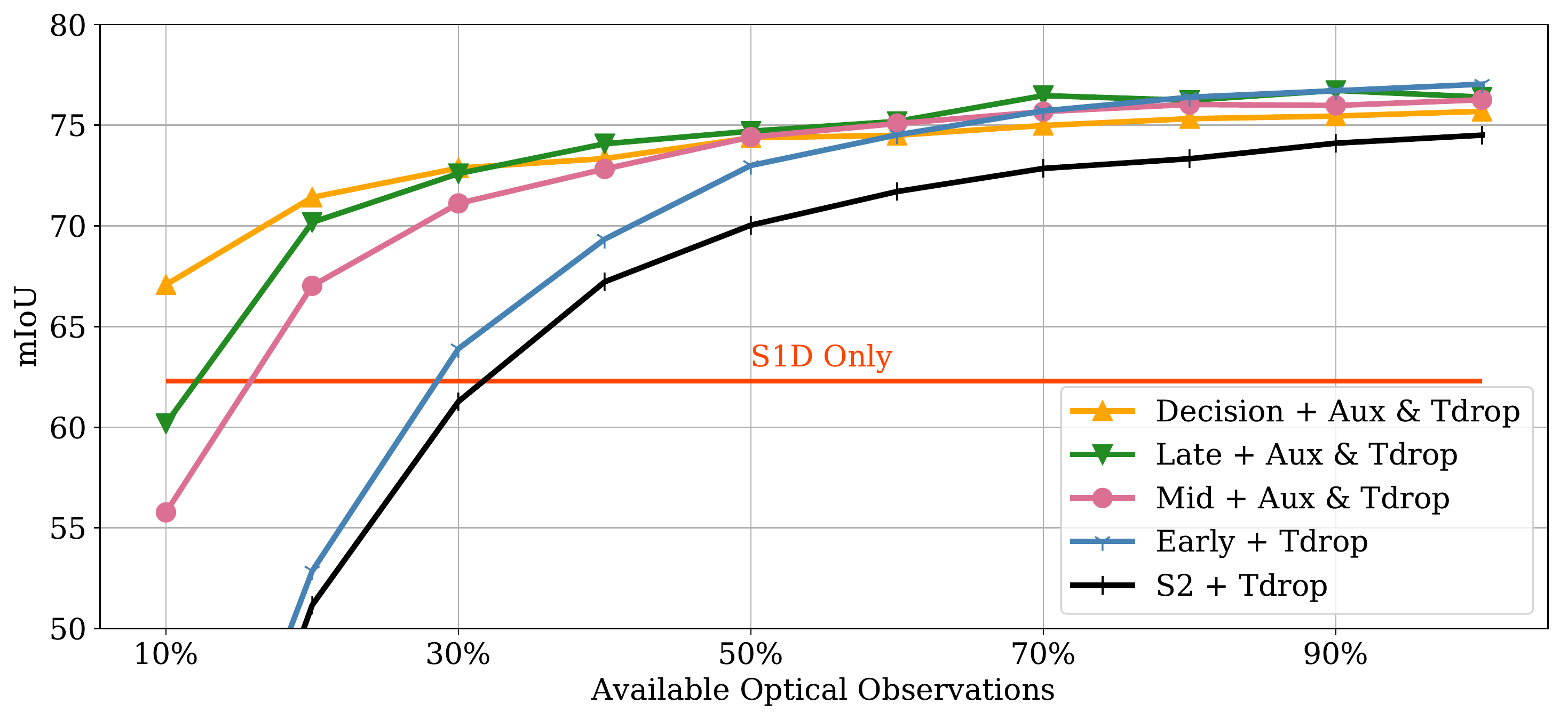}
    \caption{Parcel-based classification}
    \label{fig:london:parcel}
    \end{subfigure}
    \vfill
    \begin{subfigure}{\linewidth}
    \centering
    \includegraphics[width=.8\linewidth]{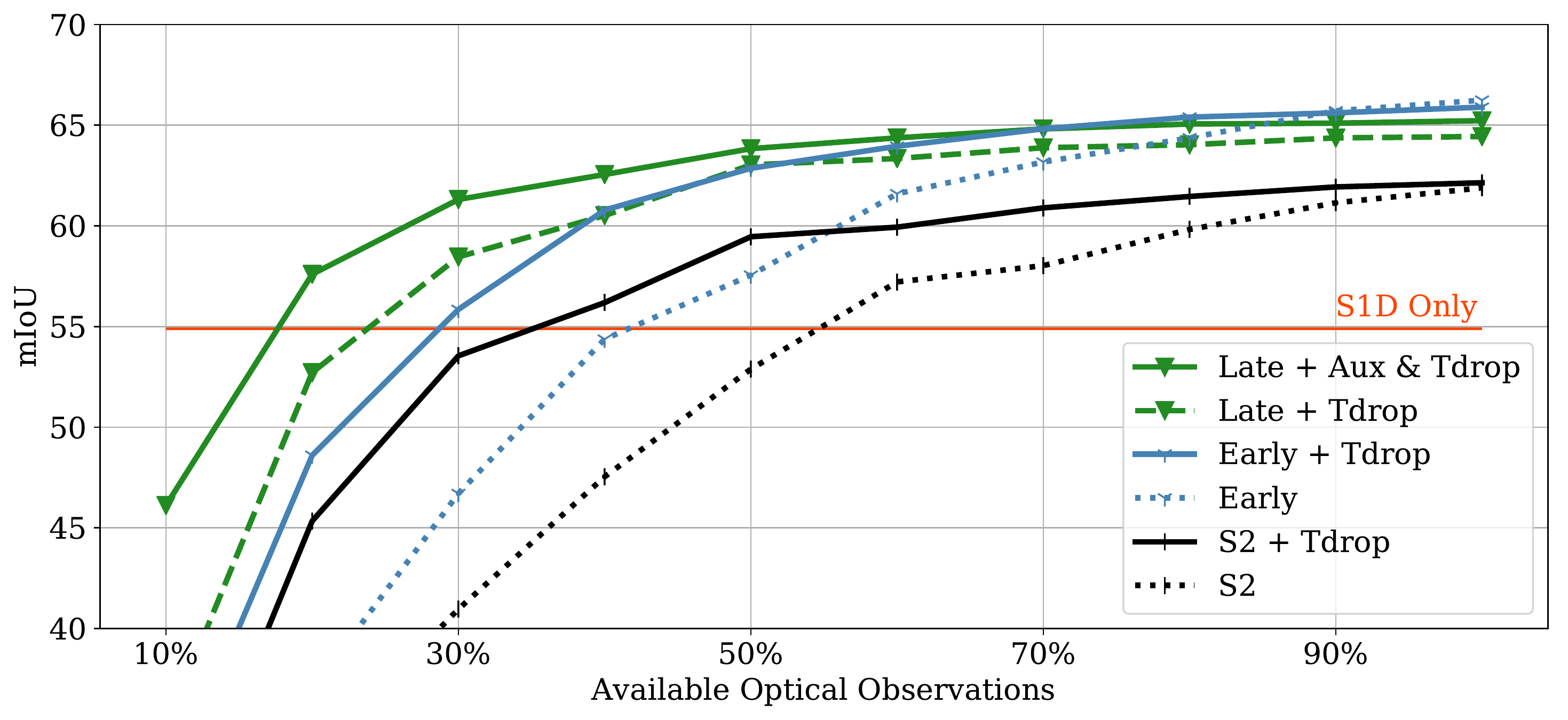}
    \caption{Semantic segmentation}
    \label{fig:london:semantic}
    \end{subfigure}
    \caption{\textbf{Varying Cloud Cover Experiment.} We {evaluate} the different models with {varying} ratios optical observations {remaining}. In both parcel-based classification \Subfigref{fig:london:parcel} and semantic segmentation \Subfigref{fig:london:semantic}, the fusion models prove more robust to a reduced number of optical observations.  }
    \label{fig:london}
\end{figure}

\paragraph{ \bf Varying Cloud Cover Experiment}
One of the motivations for using both optical and radar images in the context of crop type mapping is to exploit the imperviousness of radar signals to cloud cover. This potentially allows our model to rely on the radar signal when optical observations are obstructed by clouds, which is particularly crucial in countries with pervasive cloud cover, such as subtropical regions \citep{orynbaikyzy2020crop}. The parcel-based and semantic experiments allow for a first exploration of this capacity, but remain {bound} to the specific cloud conditions of the French metropolitan territory and the {year} of acquisition ({2019}). We propose to further investigate this benefit of multimodality by artificially simulating increased cloud obstruction {on the test set}. To do so, we evaluate the performance of models when removing random optical acquisitions while leaving the radar time series unchanged. We report the performance of the models in Figure \ref{fig:london}, for different ratios of remaining optical observations, corresponding to different levels of cloud obstruction.

As expected, the performance of the S2-only model drops drastically as the number of available optical observations decreases for both parcel classification and semantic segmentation, performing worse than unimodal radar models for {a ratio of} $70$\% of artificial occlusion. 
Multimodal fusion models can maintain an almost constant level of performance for up to $50$\% missing optical acquisitions. For more extreme ratios, the performances of the multimodal models eventually drop. The magnitude of the drop seems to be related to the amount of interplay between modalities in the network.
{Early fusion proves the least robust to missing optical observations.} Mid-fusion, and to a lesser extent, the late fusion are also affected by obstruction. These models rely on multimodal encoders and decoders, which are likely to be affected by a severe decrease in the quality of the optical sequence. In contrast, the decision fusion scheme comprises independent classifiers and proves to be the most resilient: even with $90$\% of optical images removed, it still outperforms the radar modality by several mIoU points.
We conclude that such models should be favored in regions with pervasive or inconsistent cloud cover.

We also observe that auxiliary supervision and temporal dropout make both unimodal and multimodal models  more resilient to missing optical acquisitions for semantic segmentation. The same phenomenon can be observed for parcel classification but was not represented for clarity.

%.............................................
\subsection{Panoptic segmentation experiment}
%.............................................
In this section, we evaluate the performance of the {early and} late fusion schemes compared to single modality baselines for panoptic segmentation. We do not assess auxiliary losses on the late fusion model  as the use of auxiliary decoders in this setting comes at a prohibitive computational cost. Indeed, the auxiliary decoders would be PaPs instance segmentation modules which already significantly impact training times on single modality architectures. Decision fusion is not evaluated here for the same reason.  Like in the semantic segmentation experiment, temporal dropout proved necessary to train the late fusion model.

\begin{table}[]
\centering
\caption{\textbf{Panoptic Segmentation Experiment}. We evaluate the panoptic segmentation performance of models operating on a single modality and multimodal models trained with the early and late fusion strategy with temporal dropout.}
\label{tab:xp:pano}
\begin{tabular}{lcccc}
\toprule
                    & SQ & RQ & PQ & Parameter count\\\midrule
S2                  &  81.3  & 49.2   & 40.4 & 1\;318k  \\
S1D                 &  77.0  &  39.3  &   30.9 & 1\;318k  \\
S1A                 &  77.4  &  38.8   &  30.6 & 1\;318k  \\
Early Fusion + Tdrop        &   \textbf{82.2 }&  \textbf{50.6} &  \textbf{42.0} &1\;791k \\
Late Fusion + Tdrop        &   81.6 &  50.5 &  41.6 & 2\;390k \\\bottomrule
%Late + Aux \& Tdrop &    &    &  & 2\;790k  \\ 
\end{tabular}
\end{table}

\paragraph{\bf Analysis} We report the results of this experiment on \tabref{tab:xp:pano}. Overall, the early and late fusion schemes increase the panoptic quality by $1.6$pt and $1.2$pt, respectively, compared to the optical baseline. This improvement is mostly driven by an increase in recognition quality, while the segmentation quality remains almost unchanged. This suggests that the radar modality helps correctly detect additional agricultural parcels rather than refining the delineation of their boundaries. Although modest, this improvement is valuable for this notoriously complex task.

\begin{figure}[h]
\centering
\begin{tabular}{ccccc}
    \includegraphics[width=.18\textwidth, trim=1cm 0.5cm 0cm 0cm, clip]{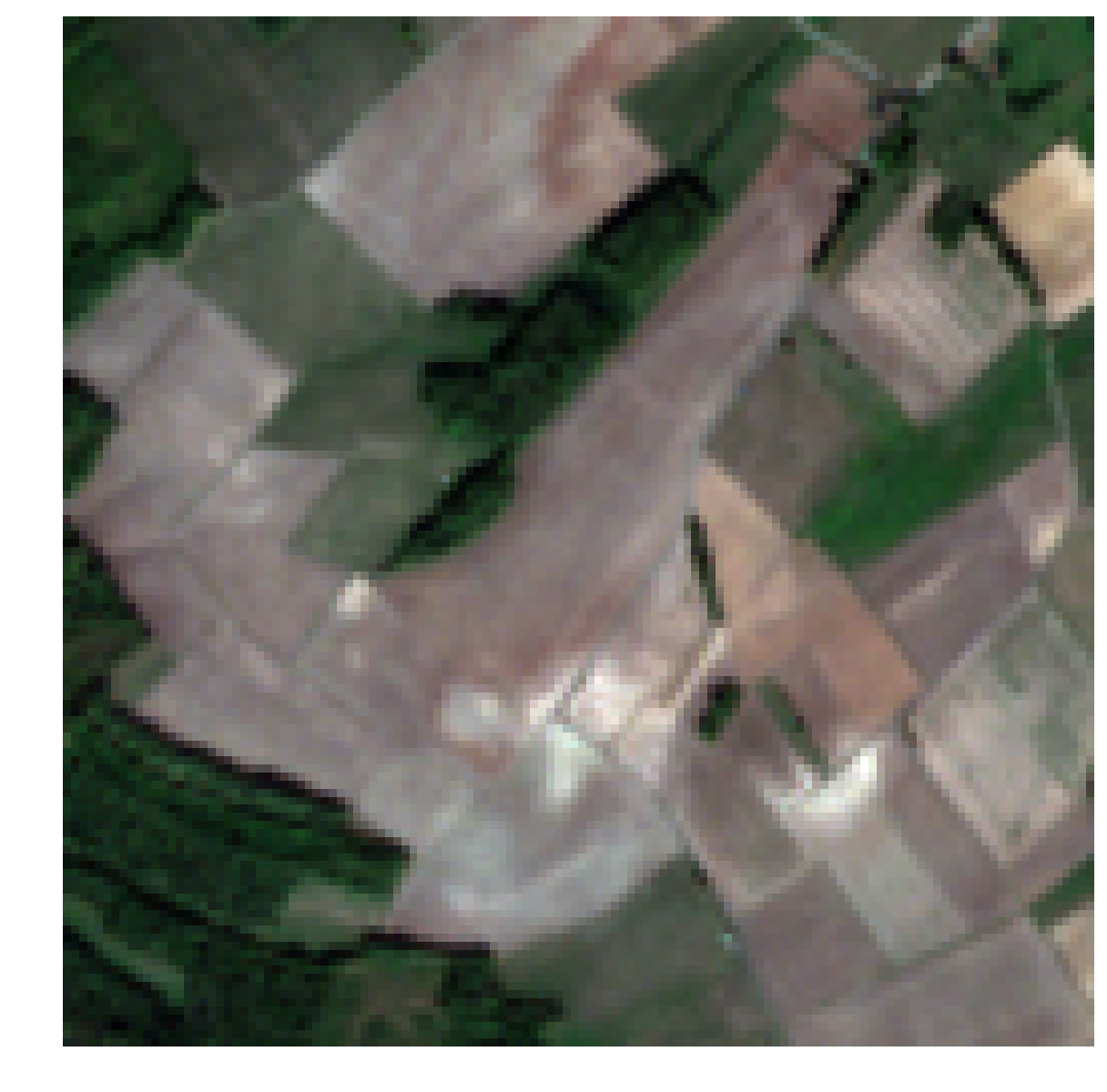}
     & 
    \includegraphics[width=.18\textwidth, trim=1cm 0.5cm 0cm 0cm, clip]{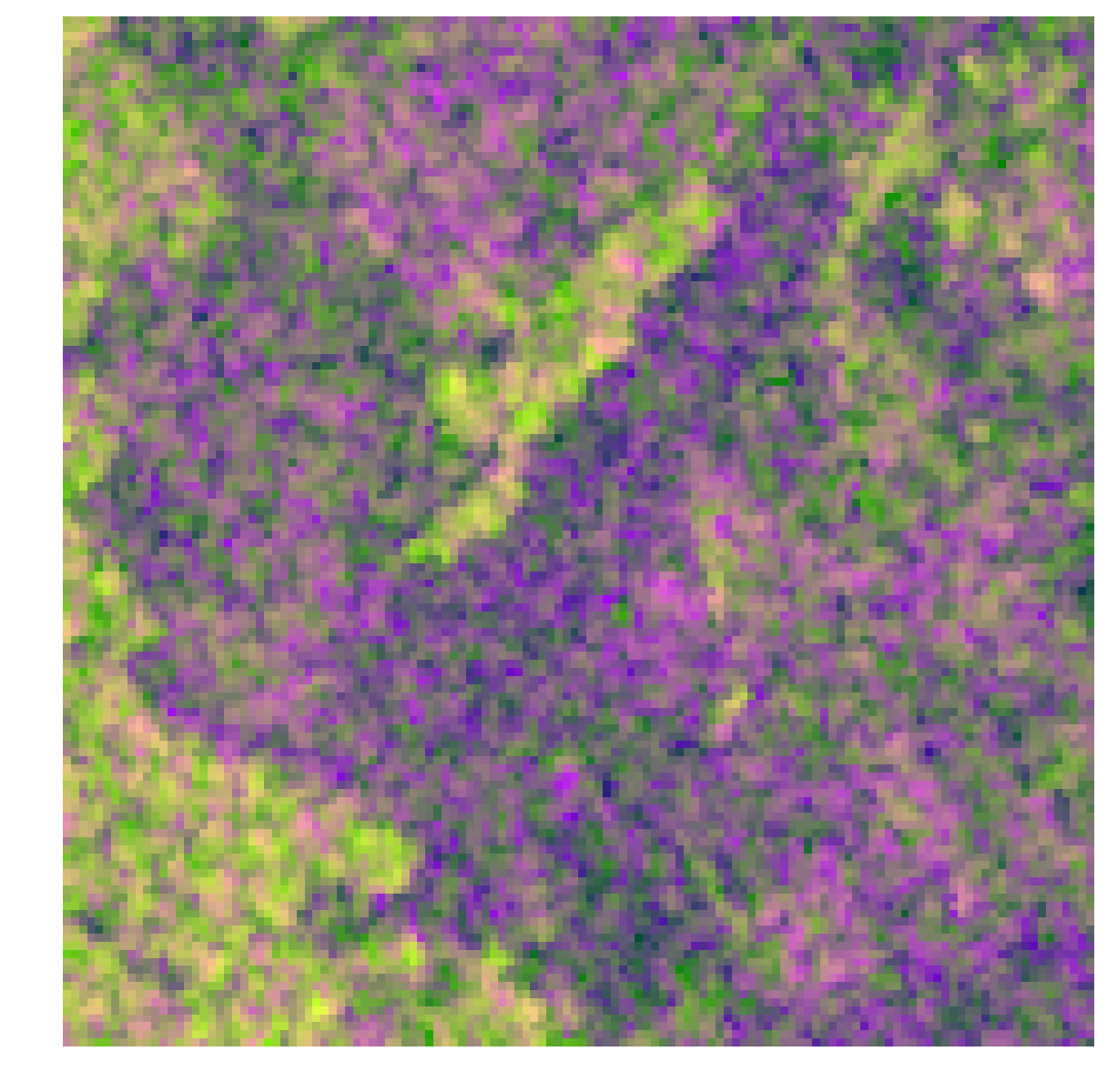}
     &
    \includegraphics[width=.18\textwidth, trim=1cm 0.5cm 0cm 0cm, clip]{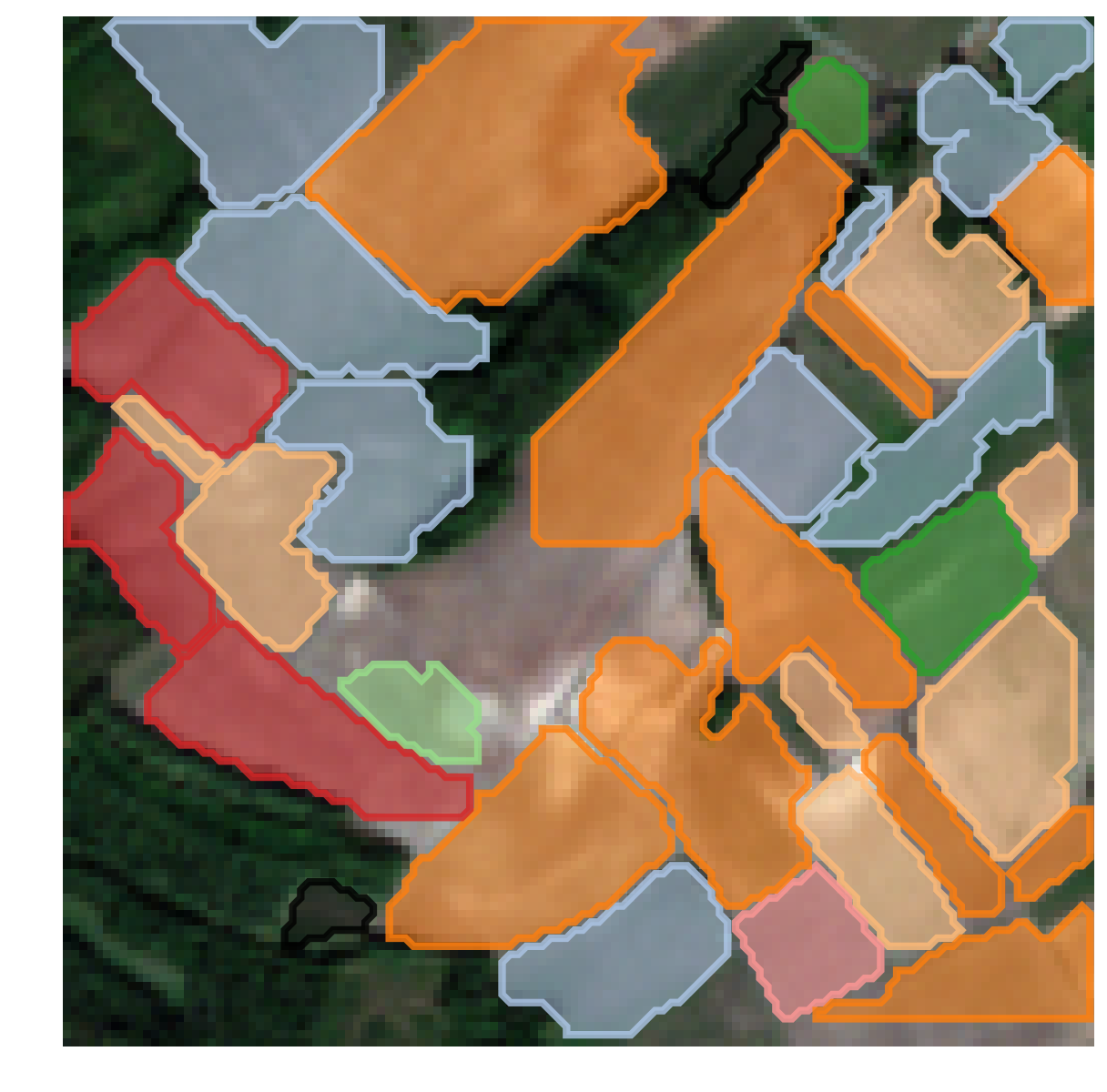}
     &
   \begin{tikzpicture}
        \node[anchor=south west,inner sep=0] (image) at (0,0) {       \includegraphics[width=.18\textwidth, trim=1cm 0.5cm 0cm 0cm, clip]{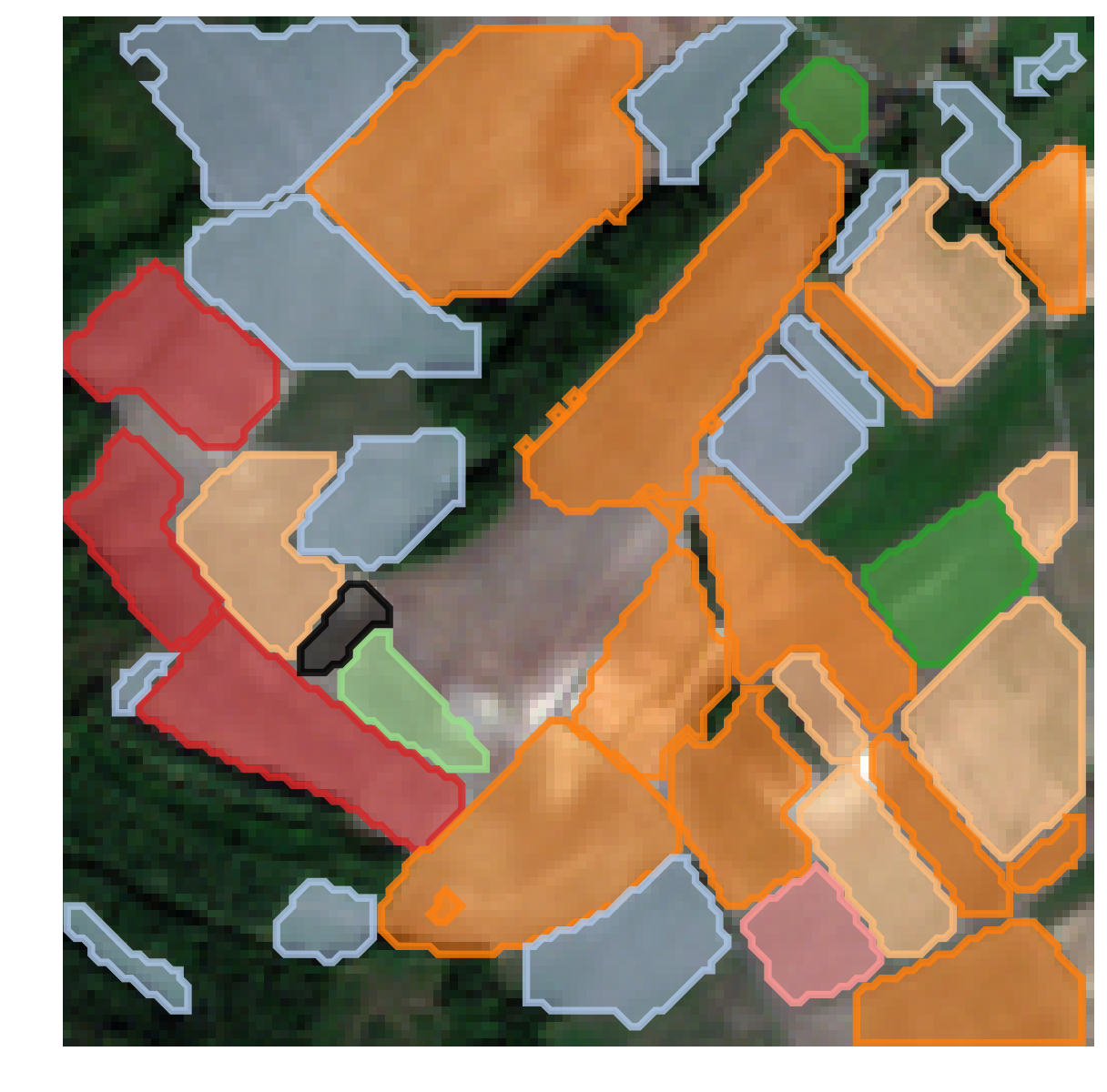}};
        \begin{scope}[x={(image.south east)},y={(image.north west)}]
            \draw[magenta,ultra thick] (0.1,0.35) circle (0.08);
            \draw[magenta,ultra thick] (0.26,0.14) circle (0.08);
        \end{scope}
    \end{tikzpicture}
     & 
    \includegraphics[width=.18\textwidth, trim=1cm 0.5cm 0cm 0cm, clip]{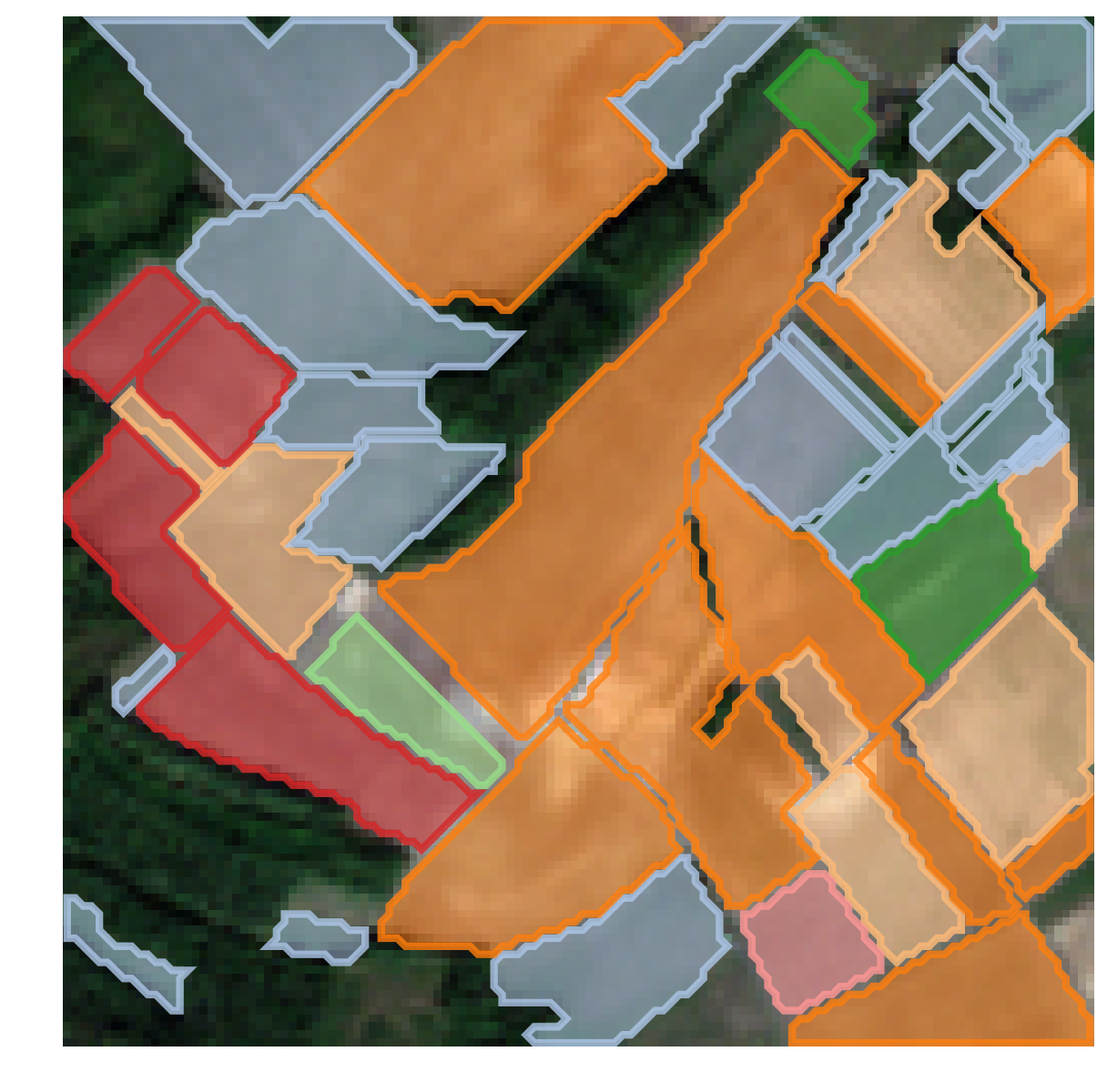}
     \\
     \includegraphics[width=.18\textwidth, trim=1cm 0.5cm 0cm 0cm, clip]{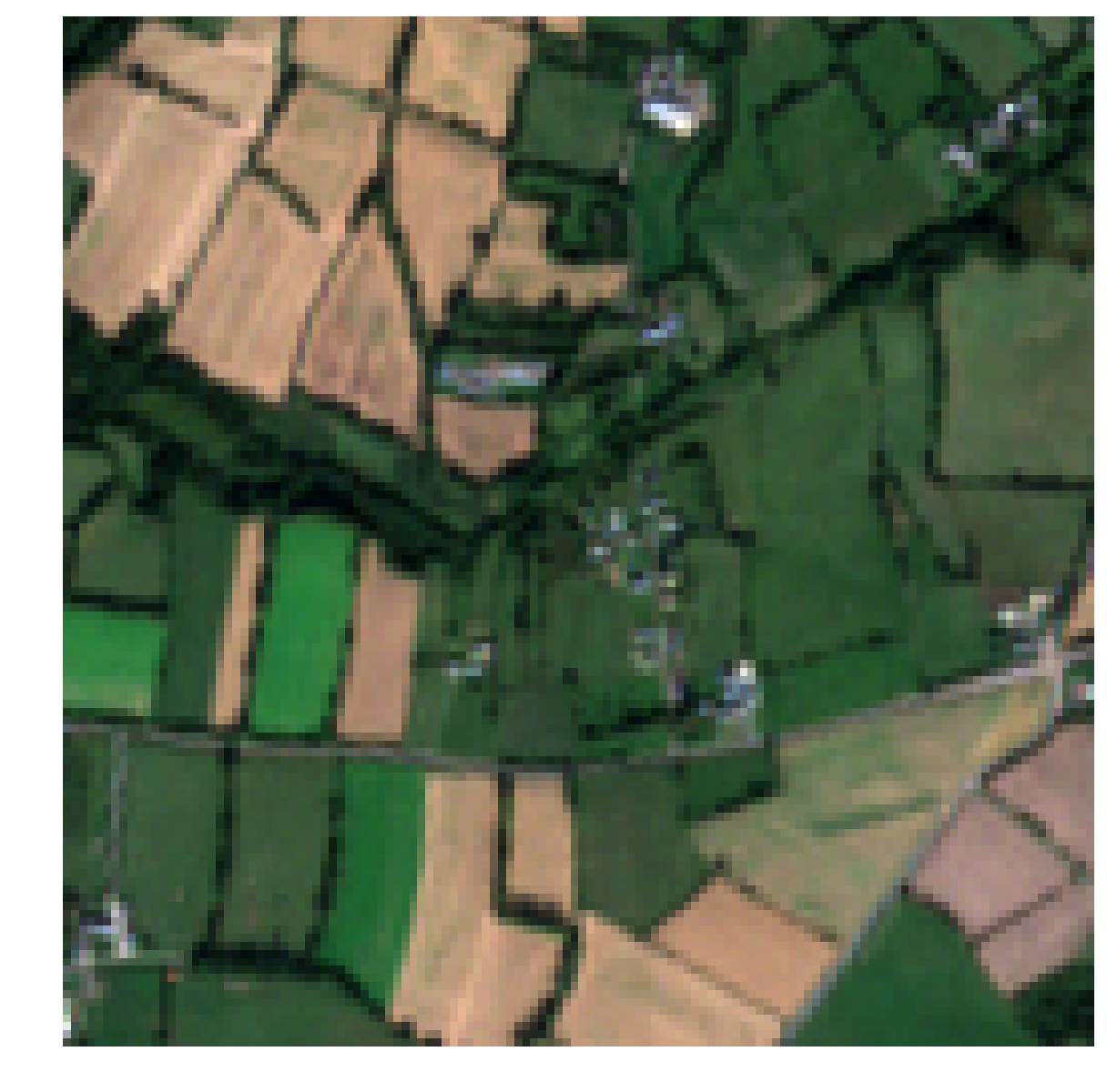}
     & 
    \includegraphics[width=.18\textwidth, trim=1cm 0.5cm 0cm 0cm, clip]{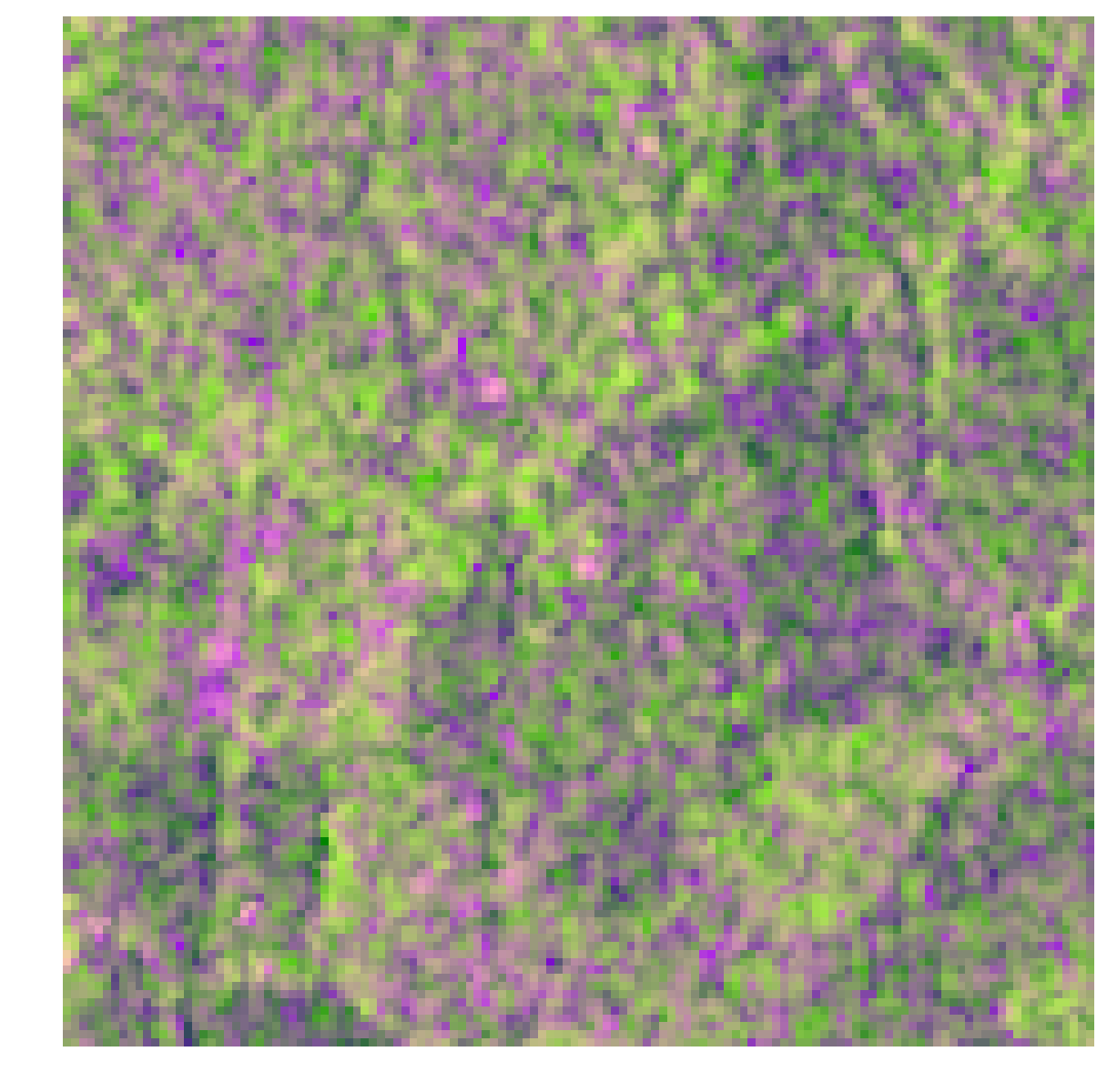}
     &
    \includegraphics[width=.18\textwidth, trim=1cm 0.5cm 0cm 0cm, clip]{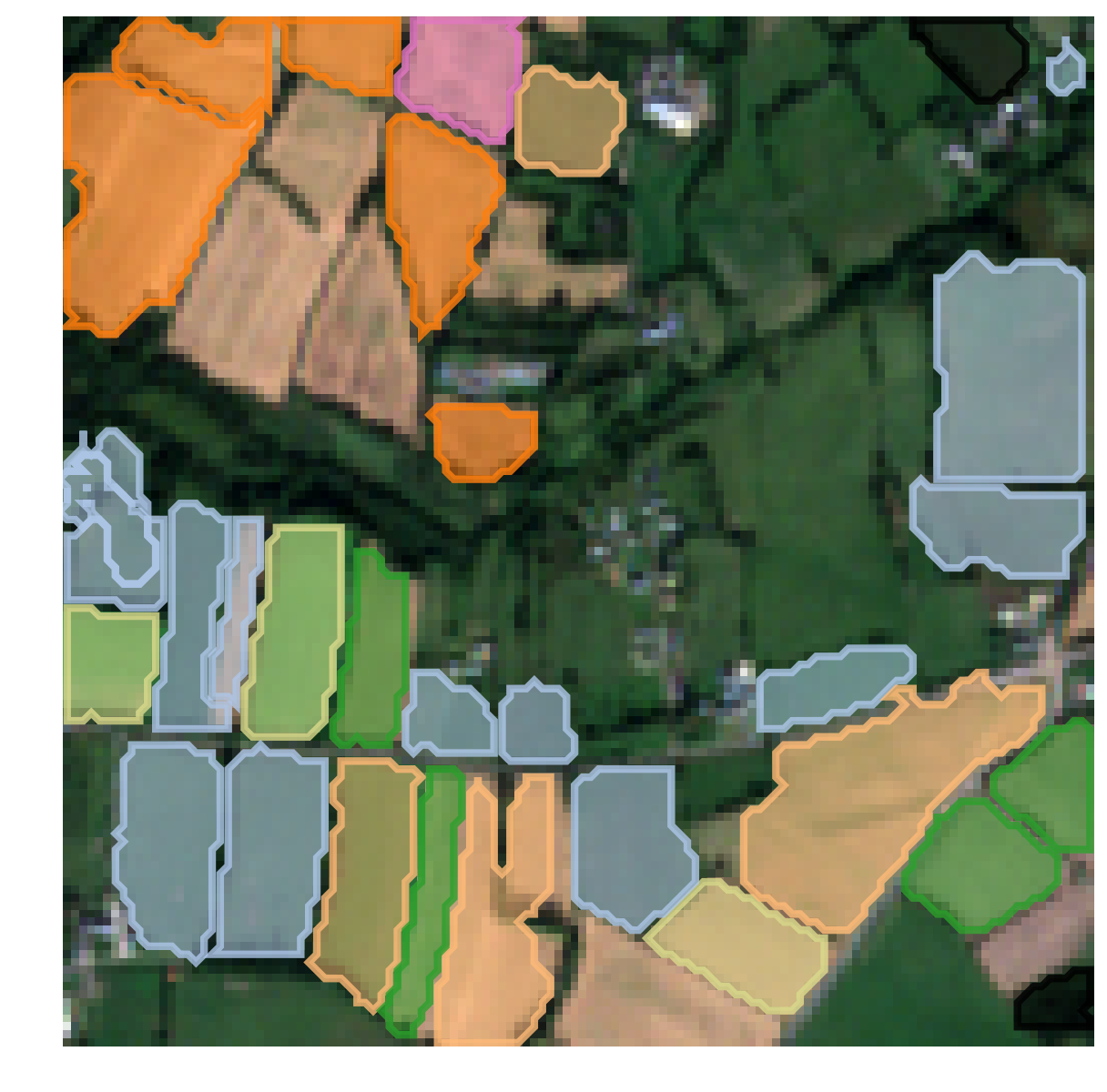}
     &
    \begin{tikzpicture}
    \node[anchor=south west,inner sep=0] (image) at (0,0) {       \includegraphics[width=.18\textwidth, trim=1cm 0.5cm 0cm 0cm, clip]{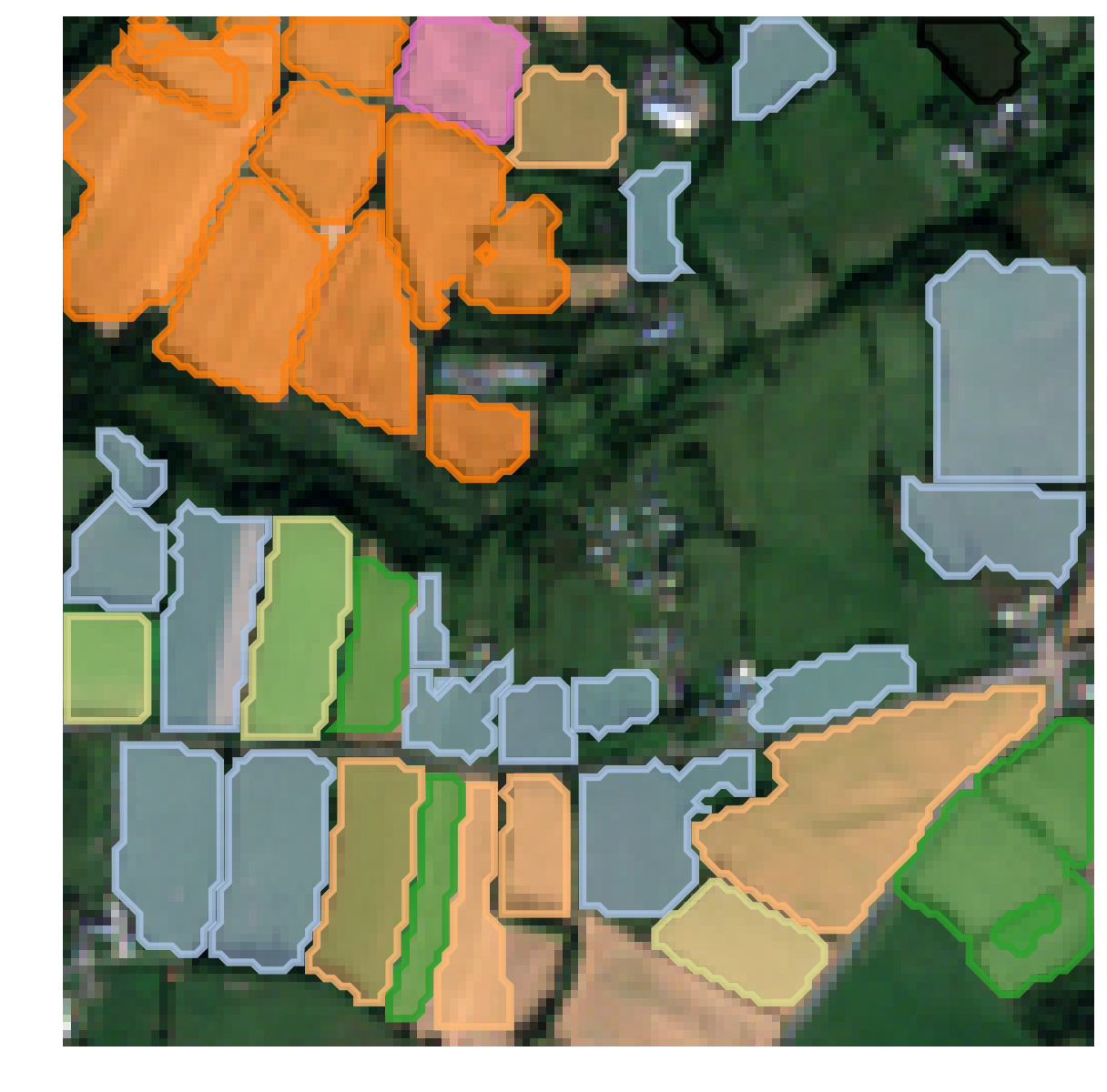}};
    \begin{scope}[x={(image.south east)},y={(image.north west)}]
        \draw[cyan,ultra thick] (0.25,0.75) circle (0.2);
    \end{scope}
    \end{tikzpicture}
     & 
    \includegraphics[width=.18\textwidth, trim=1cm 0.5cm 0cm 0cm, clip]{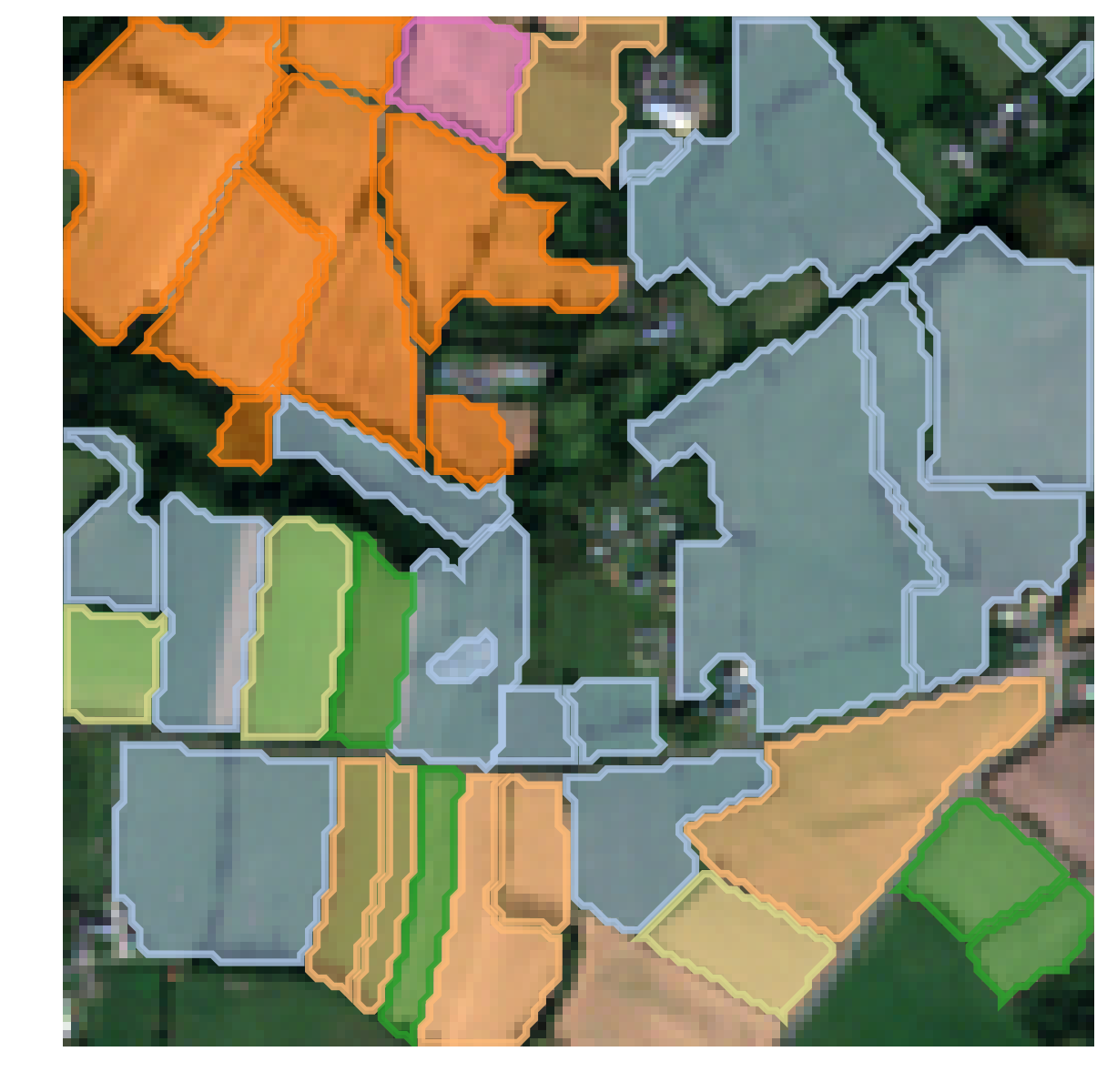}
     \\
     \includegraphics[width=.18\textwidth, trim=1cm 0.5cm 0cm 0cm, clip]{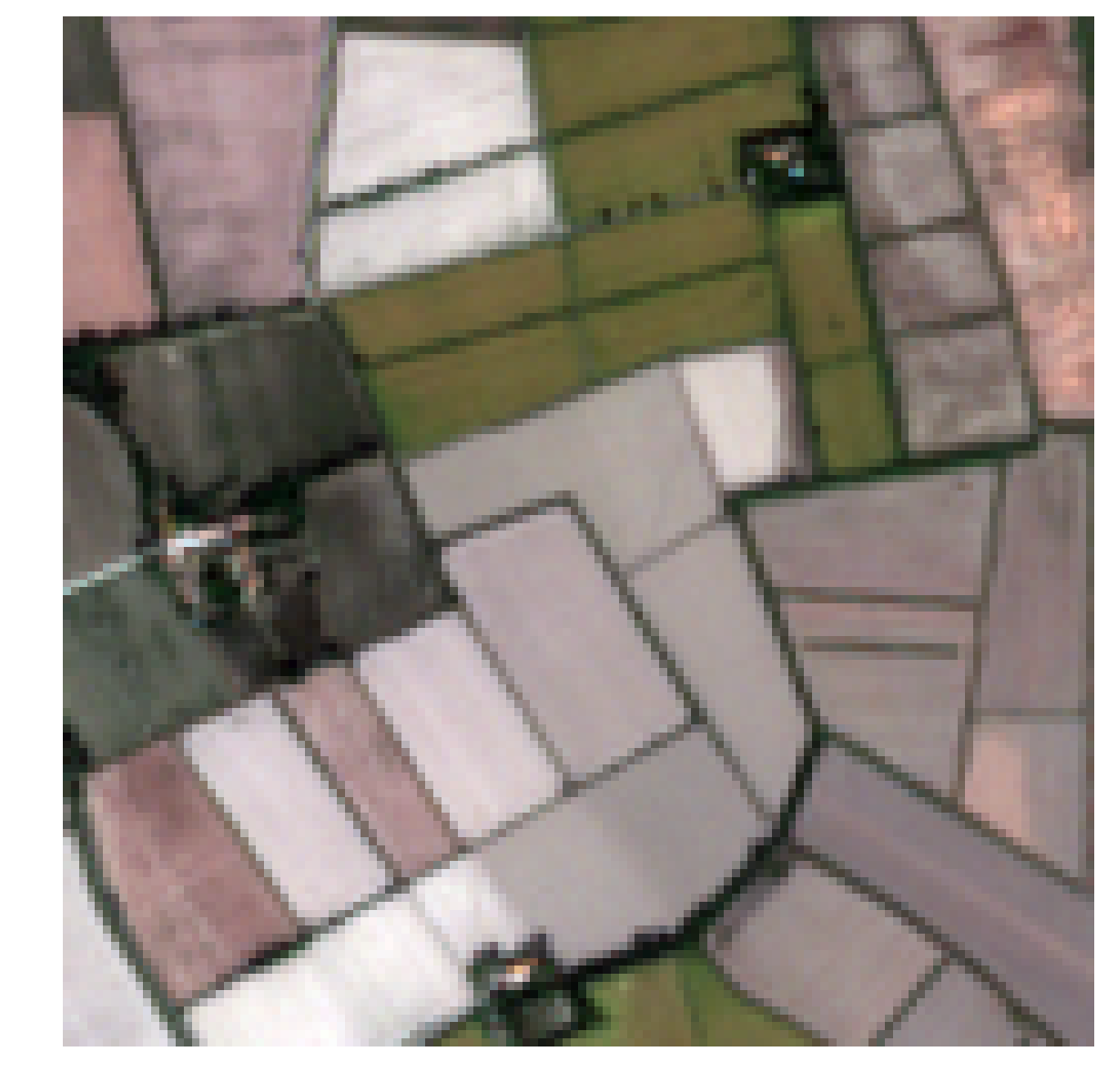}
     & 
    \includegraphics[width=.18\textwidth, trim=1cm 0.5cm 0cm 0cm, clip]{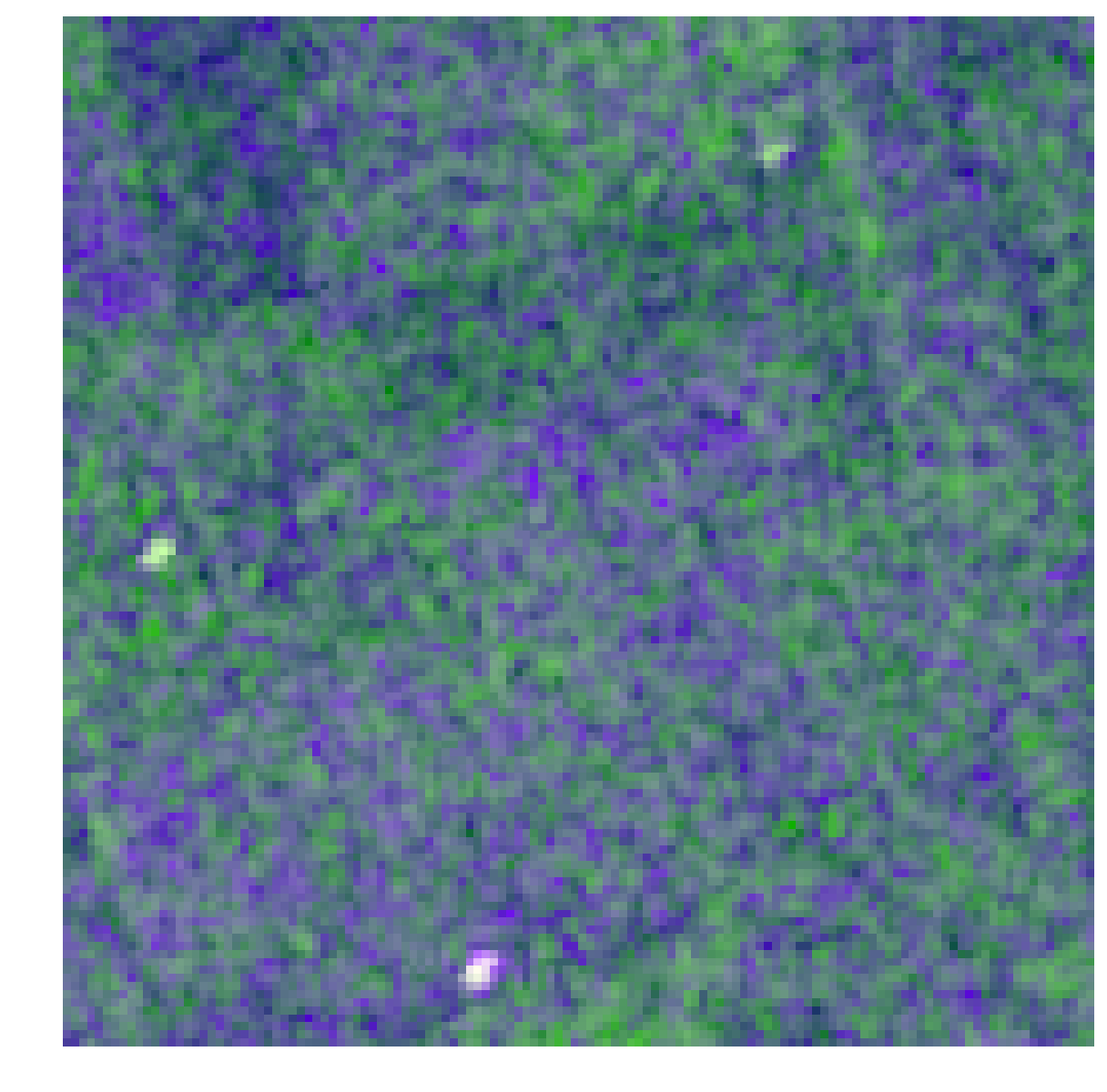}
     &
    \includegraphics[width=.18\textwidth, trim=1cm 0.5cm 0cm 0cm, clip]{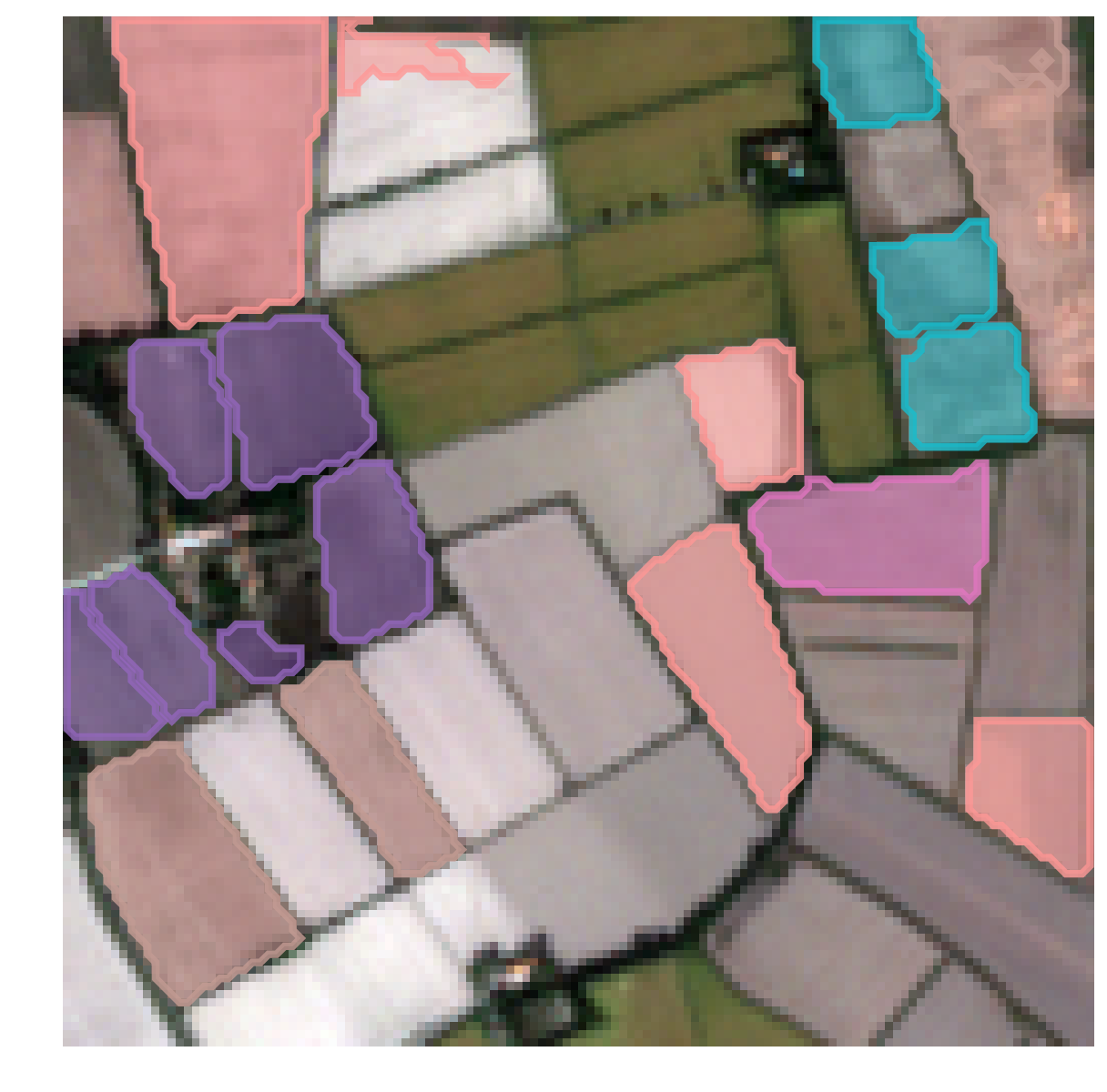}
     &
    \begin{tikzpicture}
    \node[anchor=south west,inner sep=0] (image) at (0,0) {       \includegraphics[width=.18\textwidth, trim=1cm 0.5cm 0cm 0cm, clip]{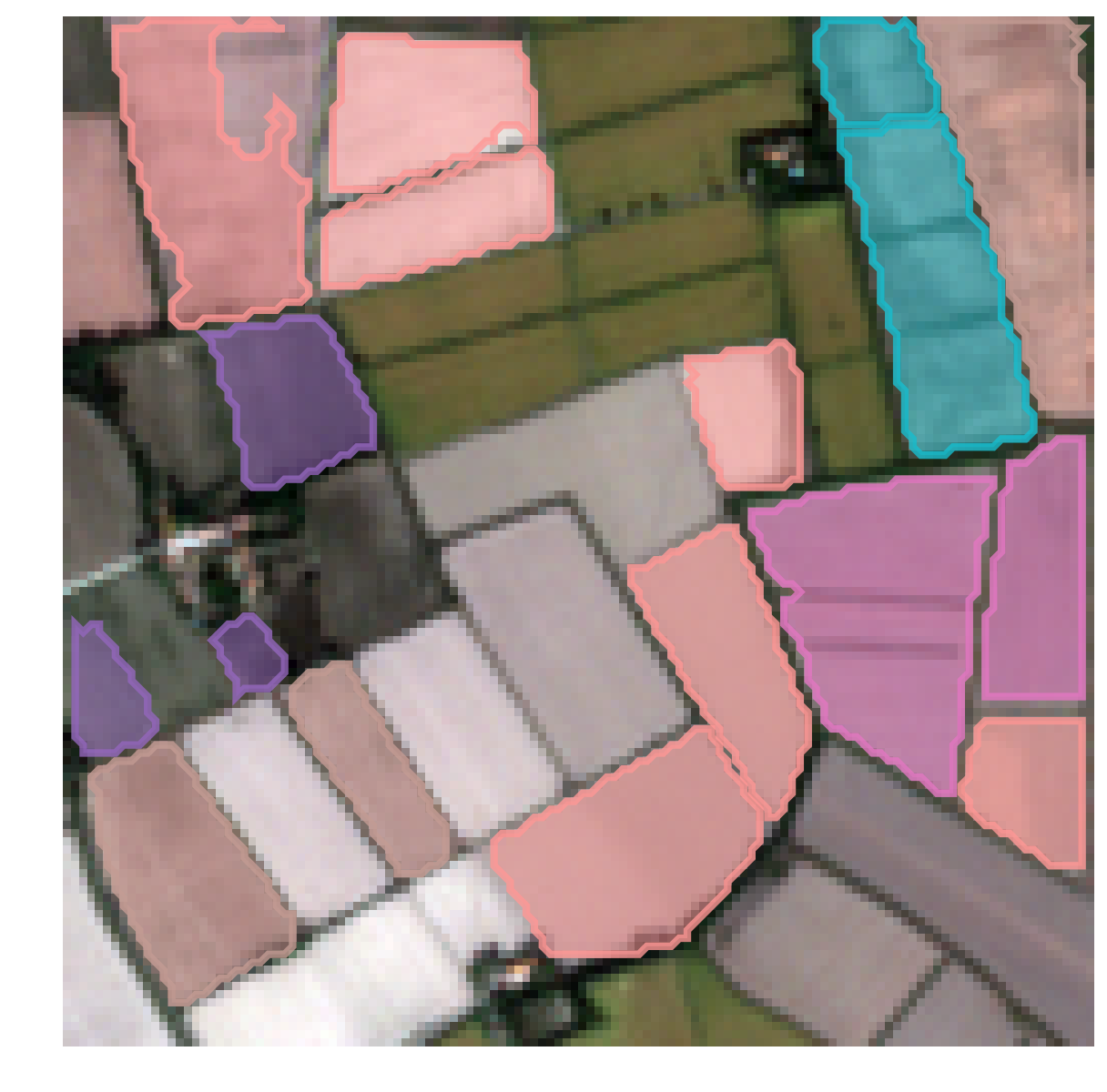}};
    \begin{scope}[x={(image.south east)},y={(image.north west)}]
        \draw[green,ultra thick] (0.8,0.75) circle (0.18);
        \draw[green,ultra thick] (0.8,0.4) circle (0.15);
    \end{scope}
    \end{tikzpicture}
     & 
    \includegraphics[width=.18\textwidth, trim=1cm 0.5cm 0cm 0cm, clip]{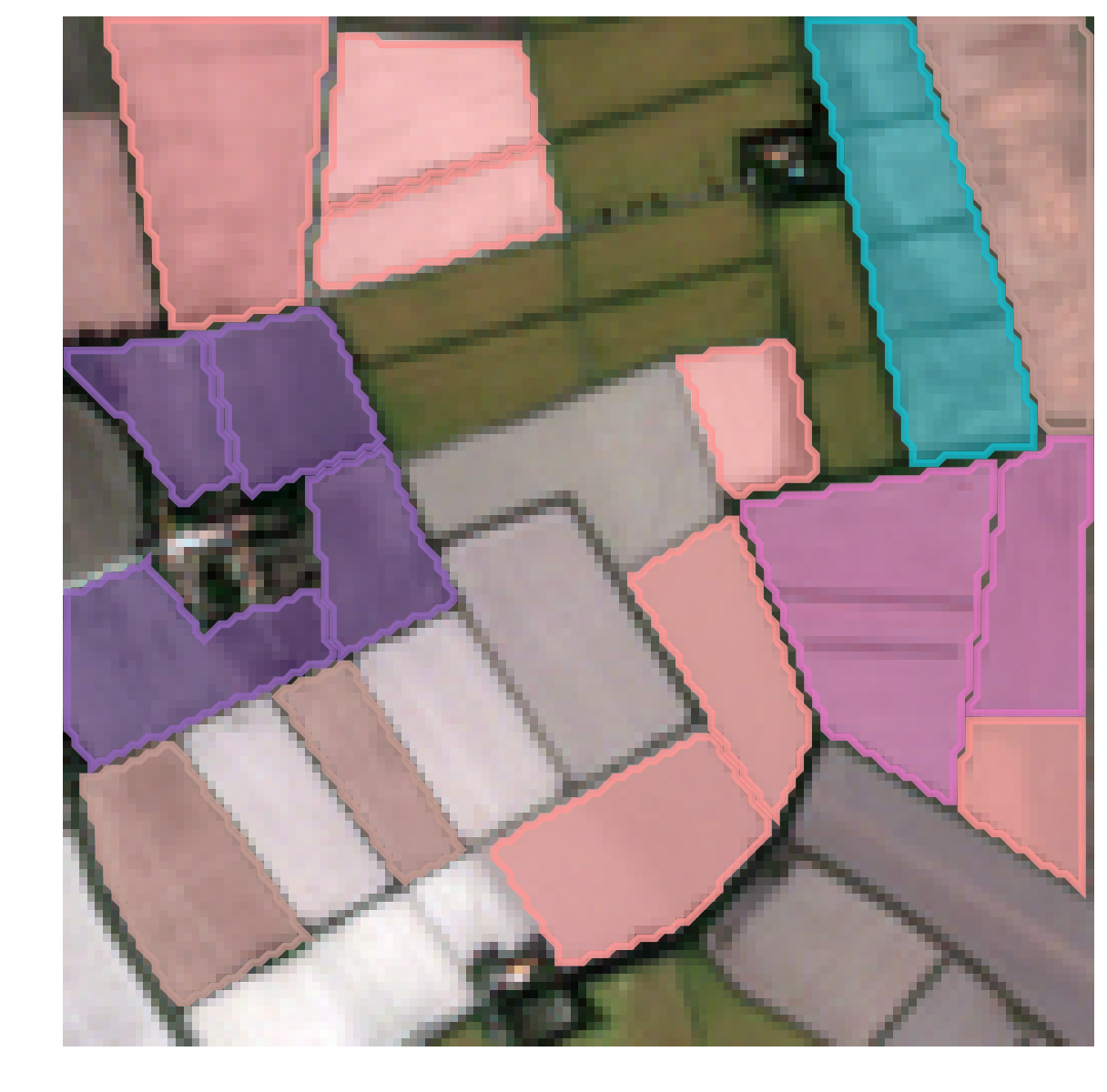}
     \\
     \begin{subfigure}{.19\textwidth}
    \caption{Optical}
    \label{fig:qualipano:s2}
    \end{subfigure}
    &
    \begin{subfigure}{.19\textwidth}
    \caption{Radar}
    \label{fig:qualipano:s1}
    \end{subfigure}

    &
    \begin{subfigure}{.19\textwidth}
    \caption{S2 Prediction}
    \label{fig:qualipano:mono}
    \end{subfigure}
    &
    \begin{subfigure}{.19\textwidth}
    \caption{Fusion Prediction}
    \label{fig:qualipano:late}
    \end{subfigure}
    &
     \begin{subfigure}{.19\textwidth}
    \caption{Ground Truth}
    \label{fig:qualipano:gt}
    \end{subfigure}
    \end{tabular}

\caption{{\bf Qualitative Results for Panoptic Segmentation.}  We show one observation from the optical time series in \Subfigref{fig:qualipano:s2} and {one} from the radar time series in \Subfigref{fig:qualipano:s1}. The prediction for the unimodal optical model is represented in \Subfigref{fig:qualipano:mono} and the multimodal model in \Subfigref{fig:qualipano:late}, and finally the ground truth in \Subfigref{fig:qualipano:gt}, with the same colormap as in \figref{fig:qualisem}. The fusion model {retrieves} more parcels (cyan circle 
\protect\tikz \protect\node[circle, thick, draw = cyan, fill = none, scale = 0.7] {};), and even {small parcels that were missed by the purely optical model} (magenta circle 
\protect\tikz \protect\node[circle, thick, draw = magenta, fill = none, scale = 0.7] {};). We also note that the fusion model seems to handle parcels with internal subdivisions (green circle 
\protect\tikz \protect\node[circle, thick, draw = green, fill = none, scale = 0.7] {};) better than the optical model.}
\label{fig:qualipano}
\end{figure}

\begin{figure}
    \centering
    \includegraphics[width=.9\linewidth]{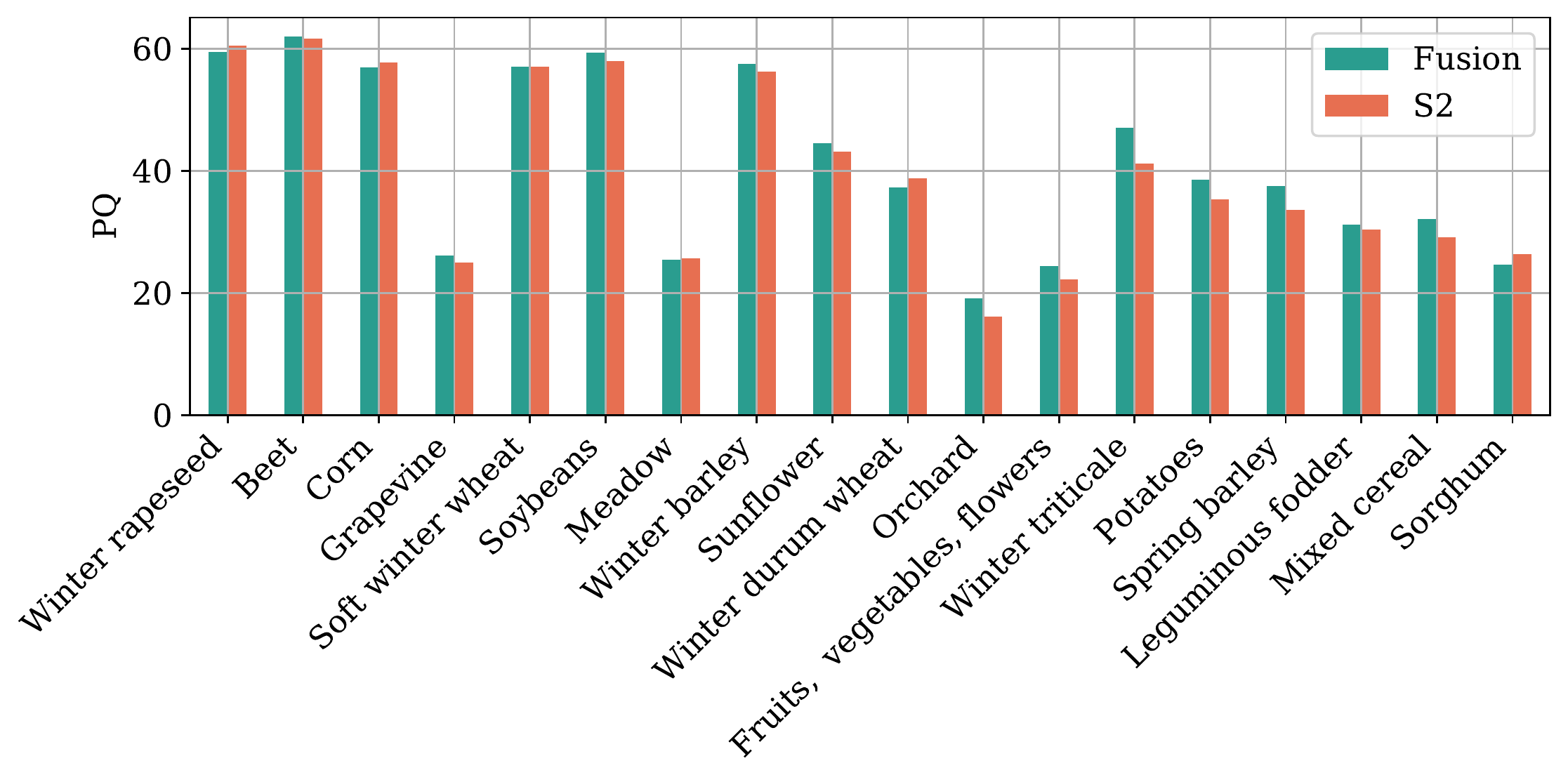}
    \caption{{\bf Classwise Performance for Parcel Classification.} We report the Panoptic Quality of the late fusion model with temporal dropout and the model trained purely on the optical modality. The classes are ordered as in \figref{fig:perclass_parcel}. In the panoptic setting, the radar modality is also specifically beneficial for hard classes such as \emph{Winter triticale}.}
    \label{fig:perclass_pano}
\end{figure}

\begin{figure}[h]
\centering
\begin{tabular}{ccccc}
    \includegraphics[width=.18\textwidth, trim=1cm 0.5cm 0cm 0cm, clip]{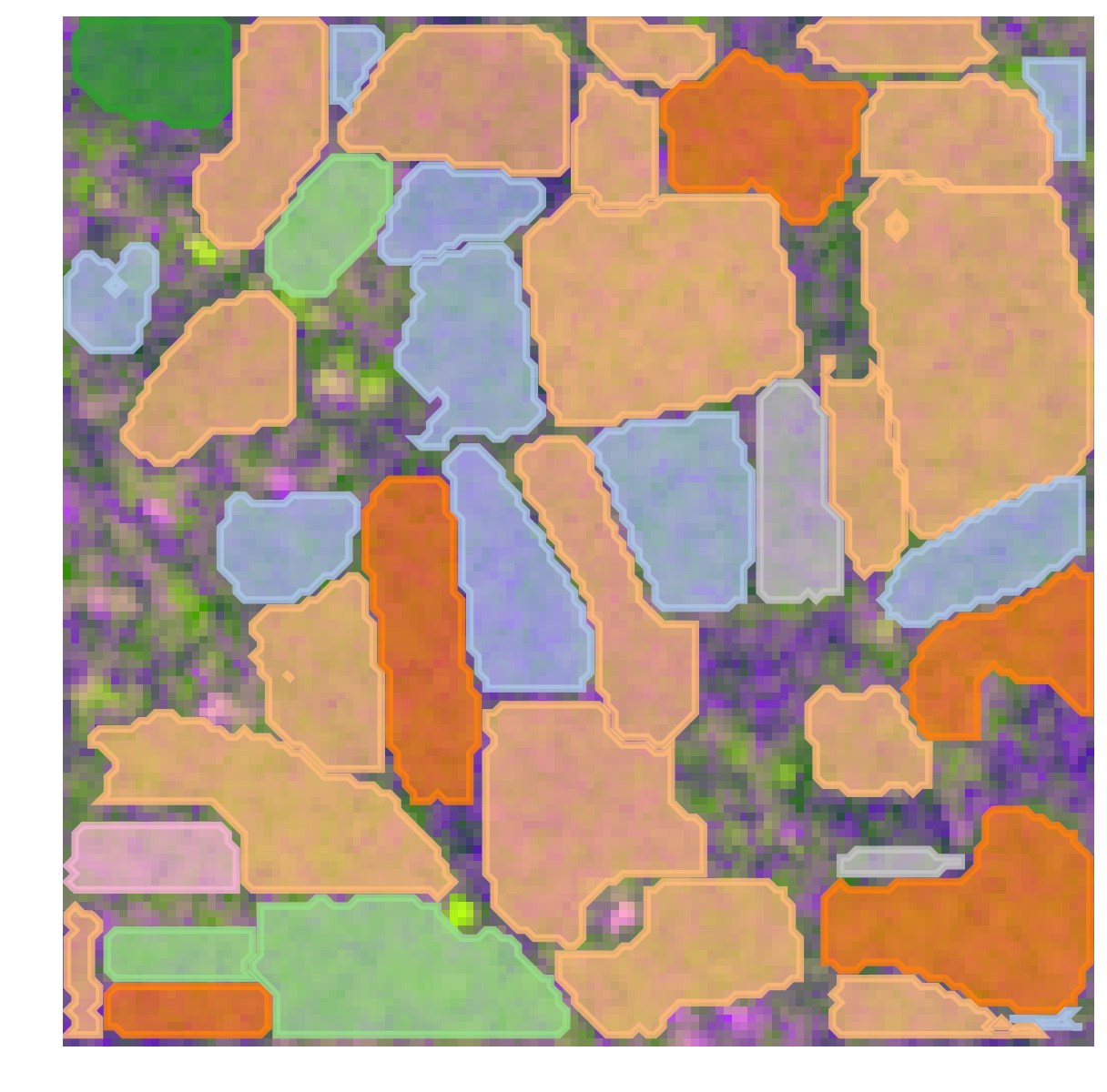}
     & 
    \includegraphics[width=.18\textwidth, trim=1cm 0.5cm 0cm 0cm, clip]{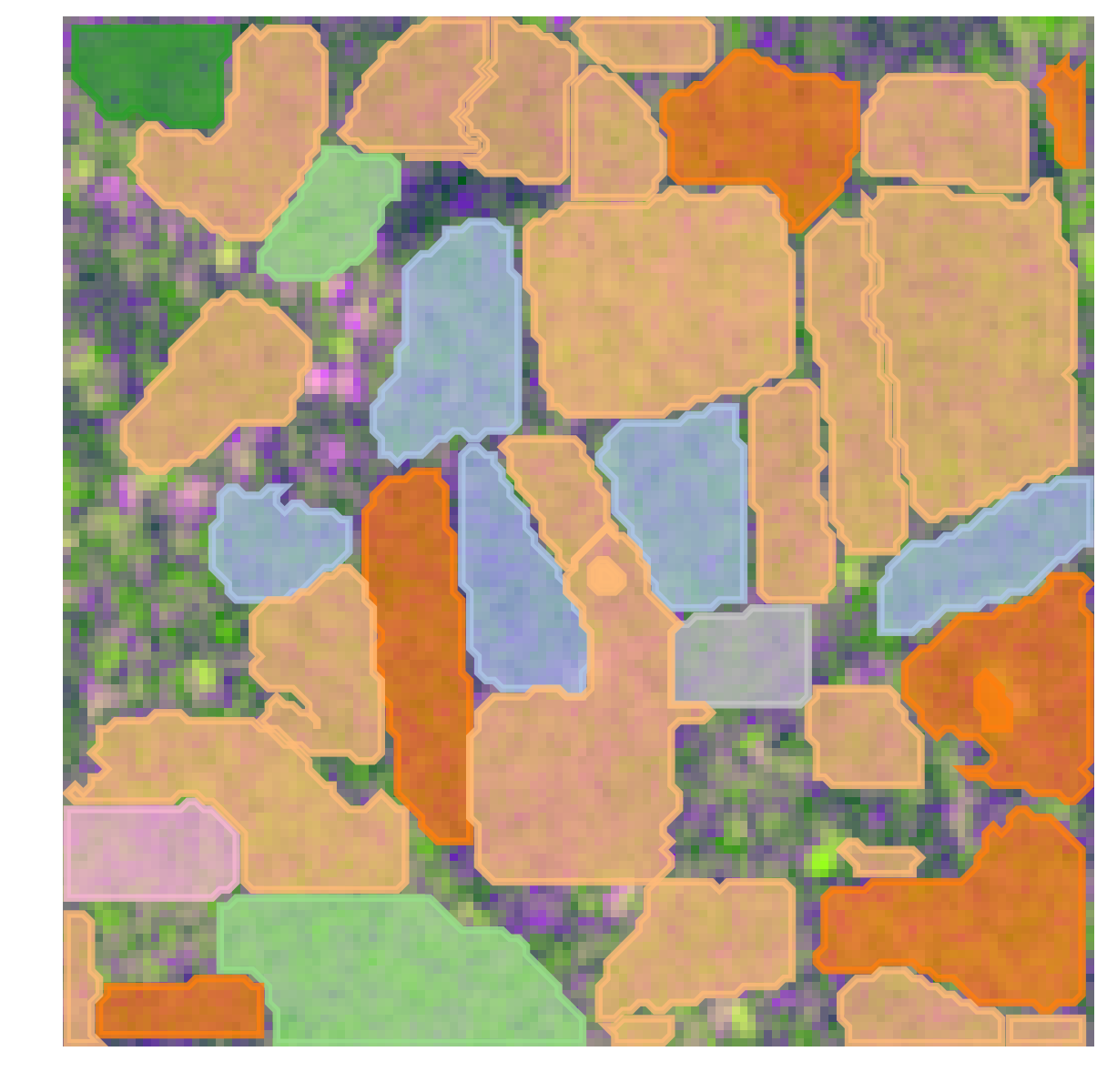}
     & 
    \includegraphics[width=.18\textwidth, trim=1cm 0.5cm 0cm 0cm, clip]{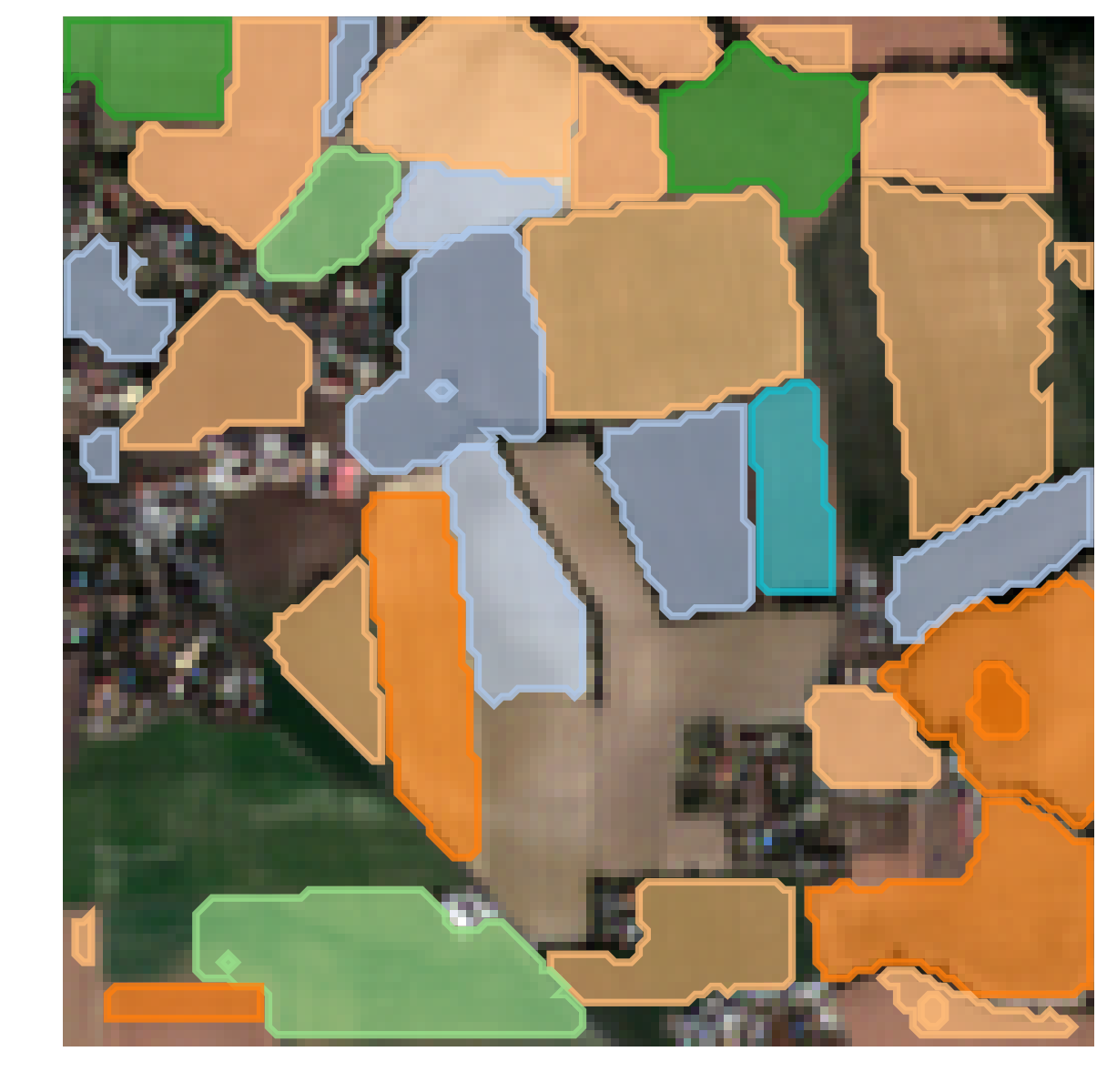}
     & 
     \begin{tikzpicture}
    \node[anchor=south west,inner sep=0] (image) at (0,0) {           \includegraphics[width=.18\textwidth, trim=1cm 0.5cm 0cm 0cm, clip]{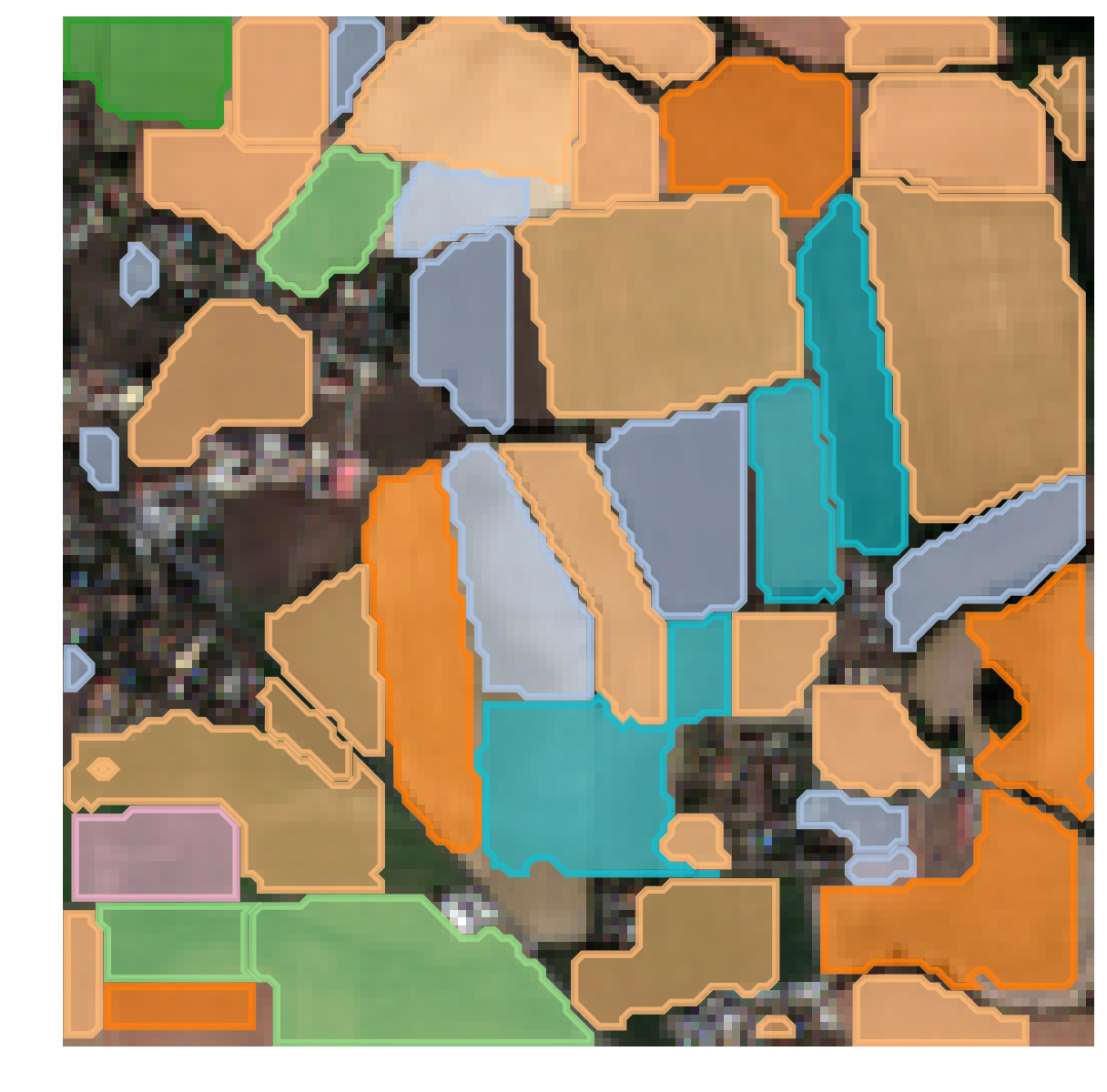}};
    \begin{scope}[x={(image.south east)},y={(image.north west)}]
        \draw[green,ultra thick] (0.15,0.23) circle (0.12);
        \draw[red,ultra thick] (0.8,0.62) circle (0.12);

    \end{scope}
    \end{tikzpicture}
     & 
     \includegraphics[width=.18\textwidth, trim=1cm 0.5cm 0cm 0cm, clip]{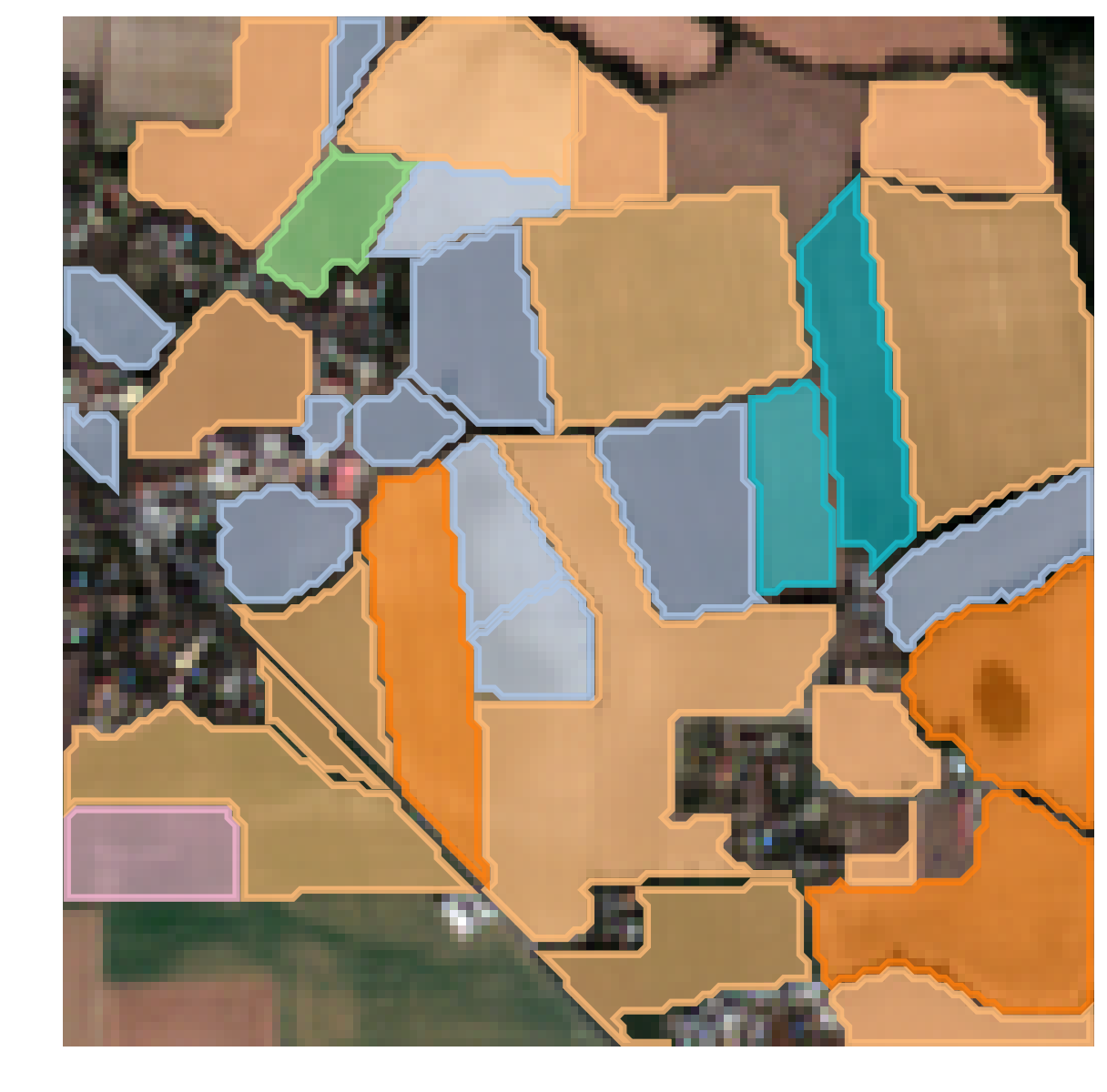}
 \\
     \includegraphics[width=.18\textwidth, trim=1cm 0.5cm 0cm 0cm, clip]{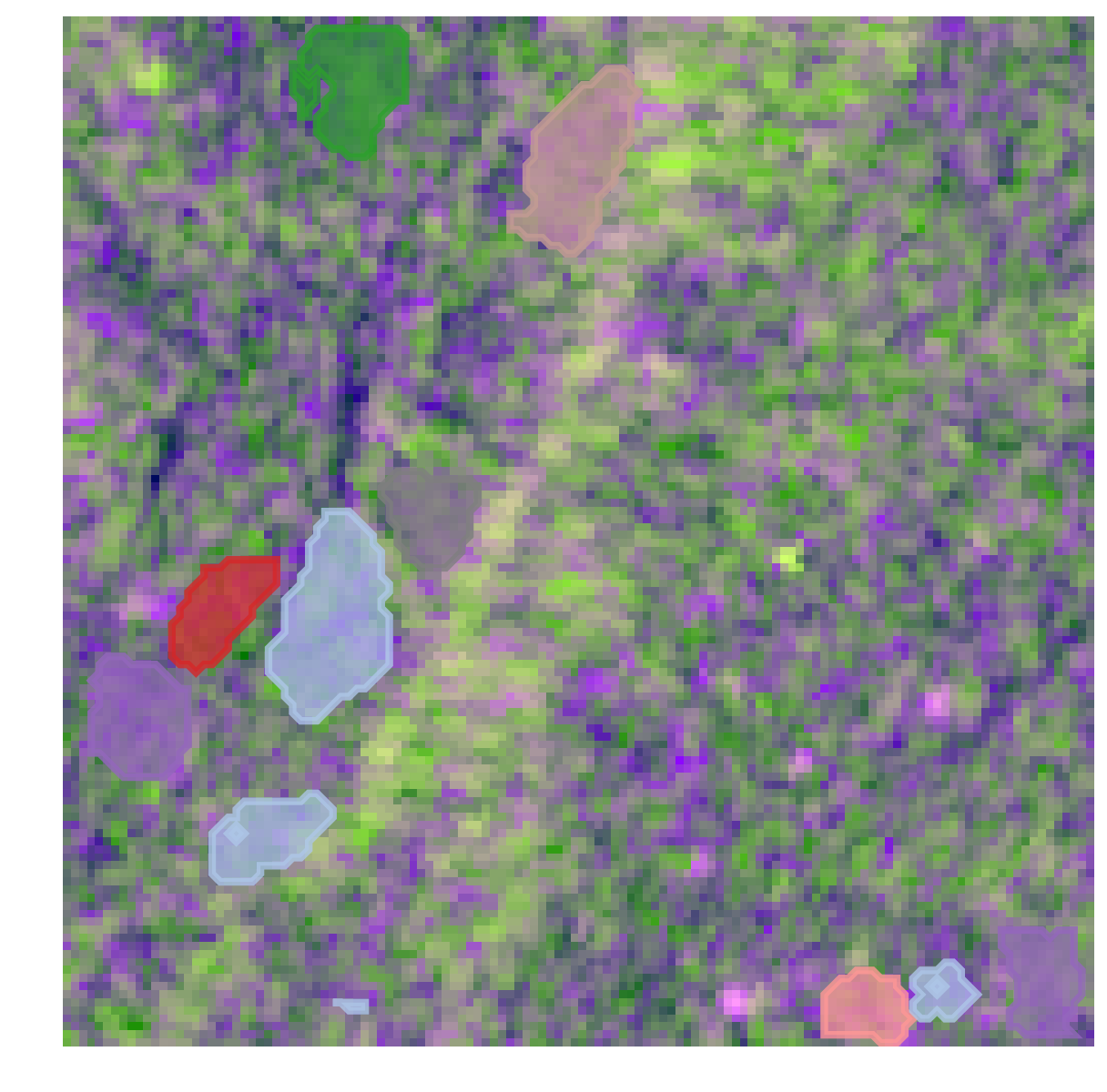}
     & 
    \includegraphics[width=.18\textwidth, trim=1cm 0.5cm 0cm 0cm, clip]{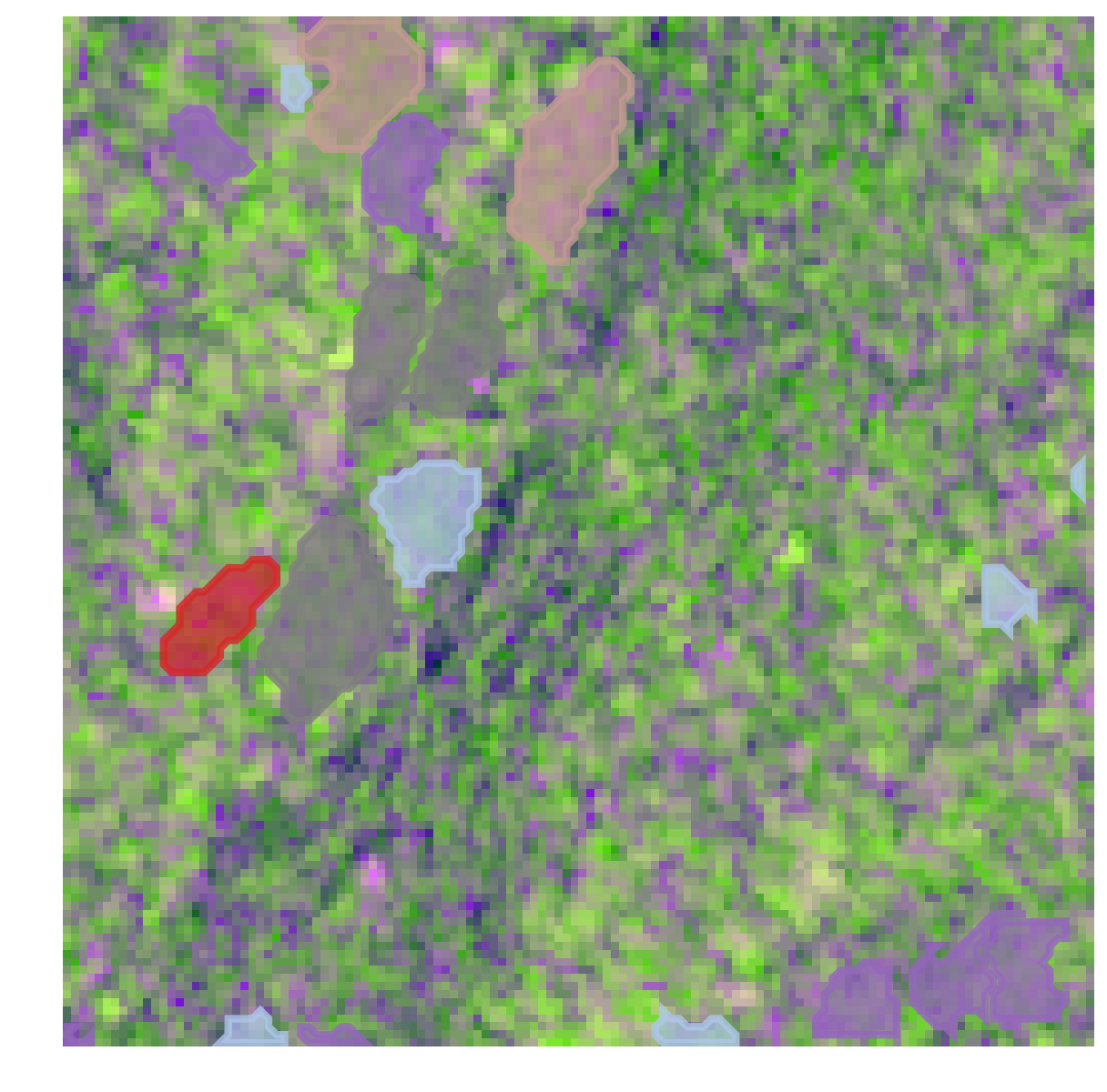}
     & 
    \includegraphics[width=.18\textwidth, trim=1cm 0.5cm 0cm 0cm, clip]{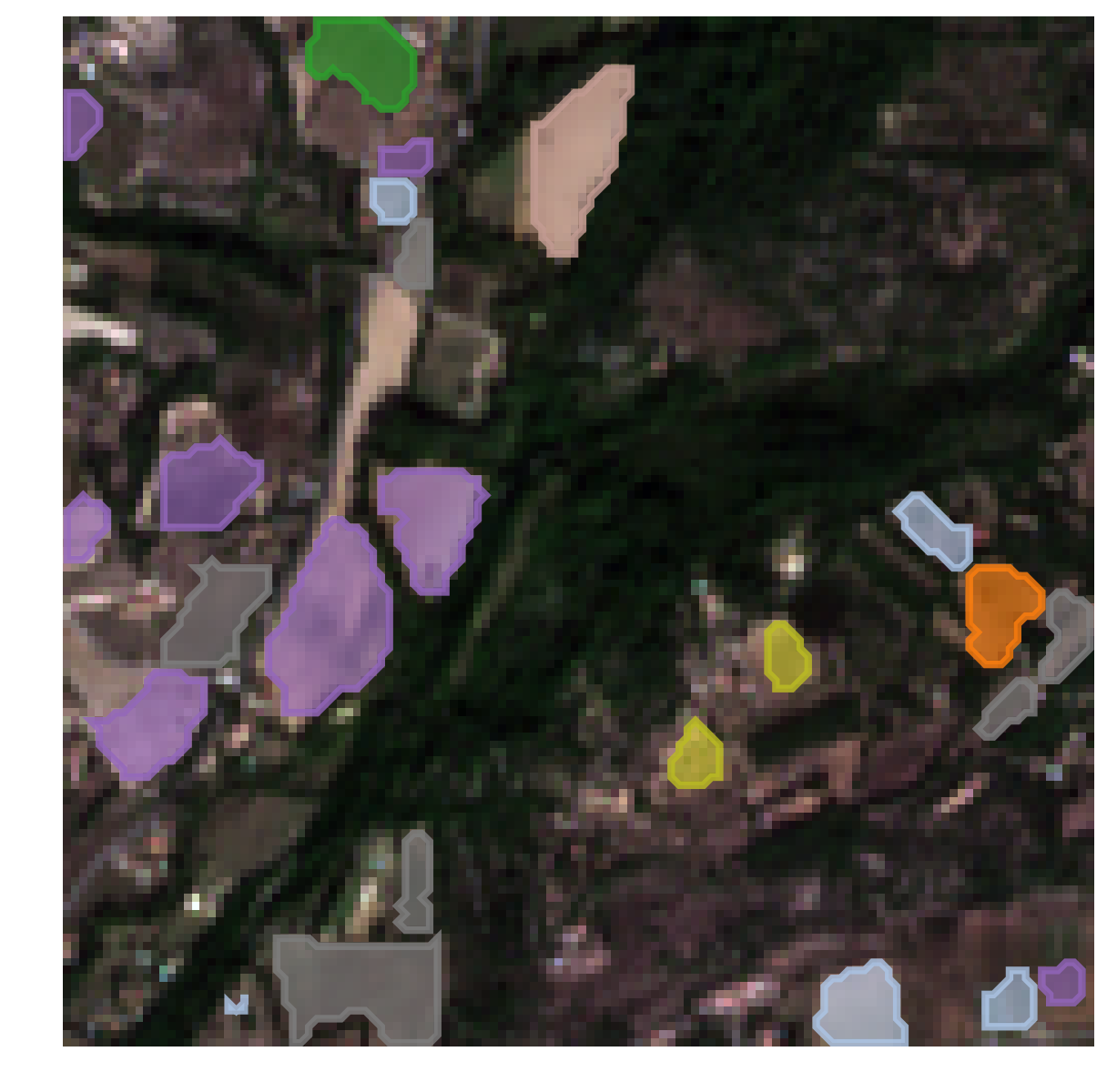}
     & 
         \begin{tikzpicture}
    \node[anchor=south west,inner sep=0] (image) at (0,0) {               \includegraphics[width=.18\textwidth, trim=1cm 0.5cm 0cm 0cm, clip]{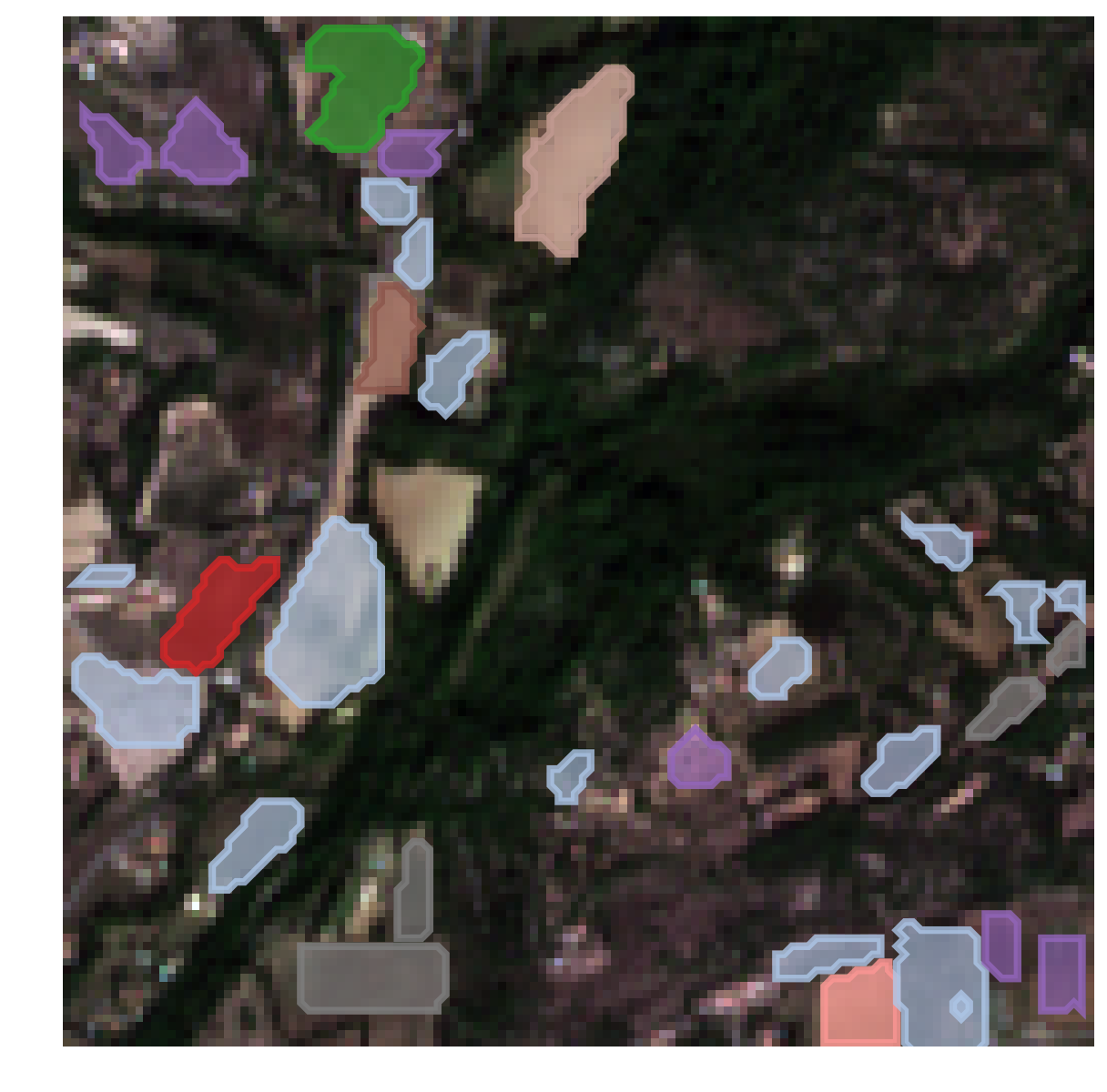}};
    \begin{scope}[x={(image.south east)},y={(image.north west)}]
        \draw[cyan,ultra thick] (0.15,0.43) circle (0.09);
    \end{scope}
    \end{tikzpicture}
    
     & 
     \includegraphics[width=.18\textwidth, trim=1cm 0.5cm 0cm 0cm, clip]{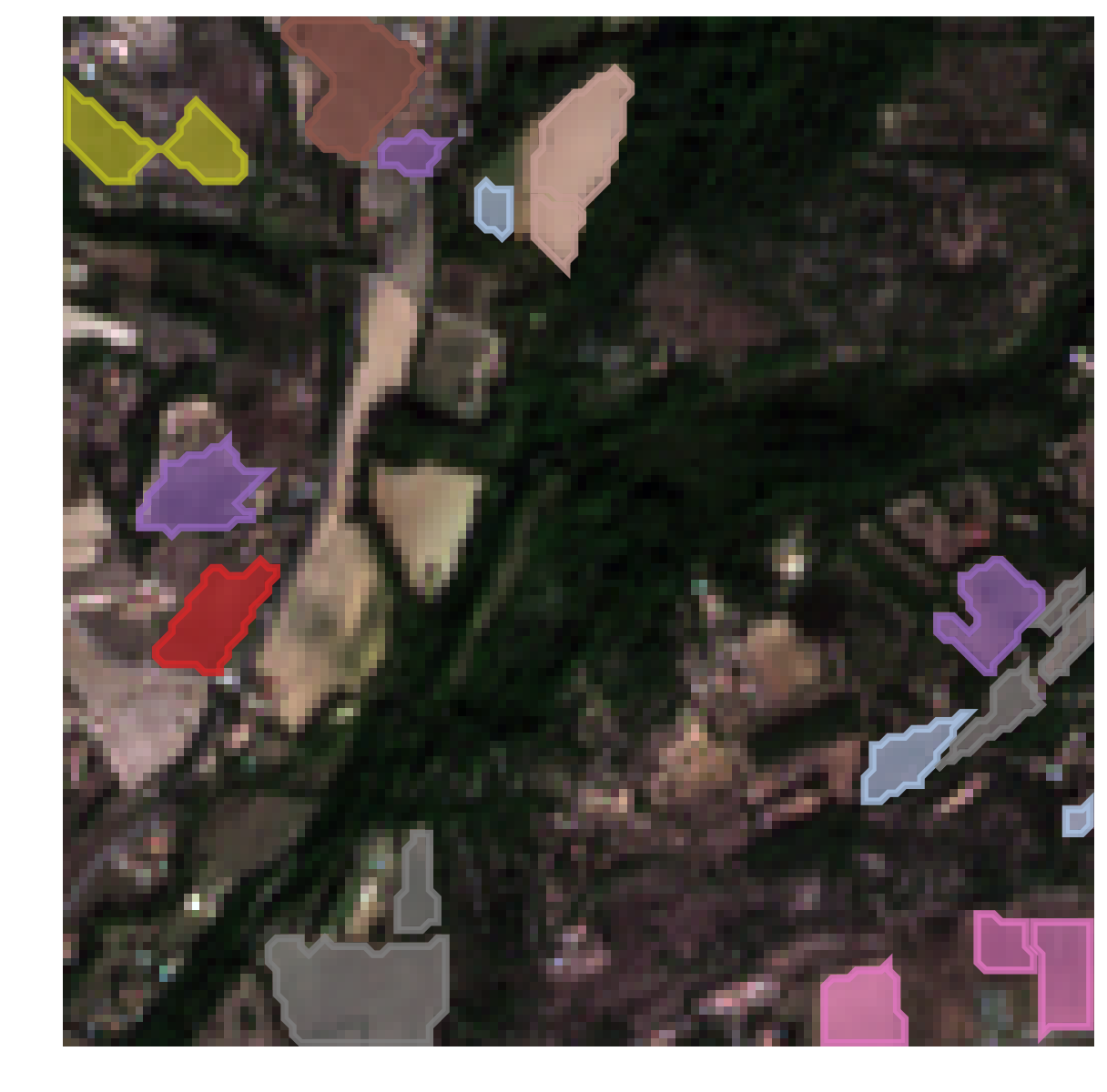}
 \\
     \includegraphics[width=.18\textwidth, trim=1cm 0.5cm 0cm 0cm, clip]{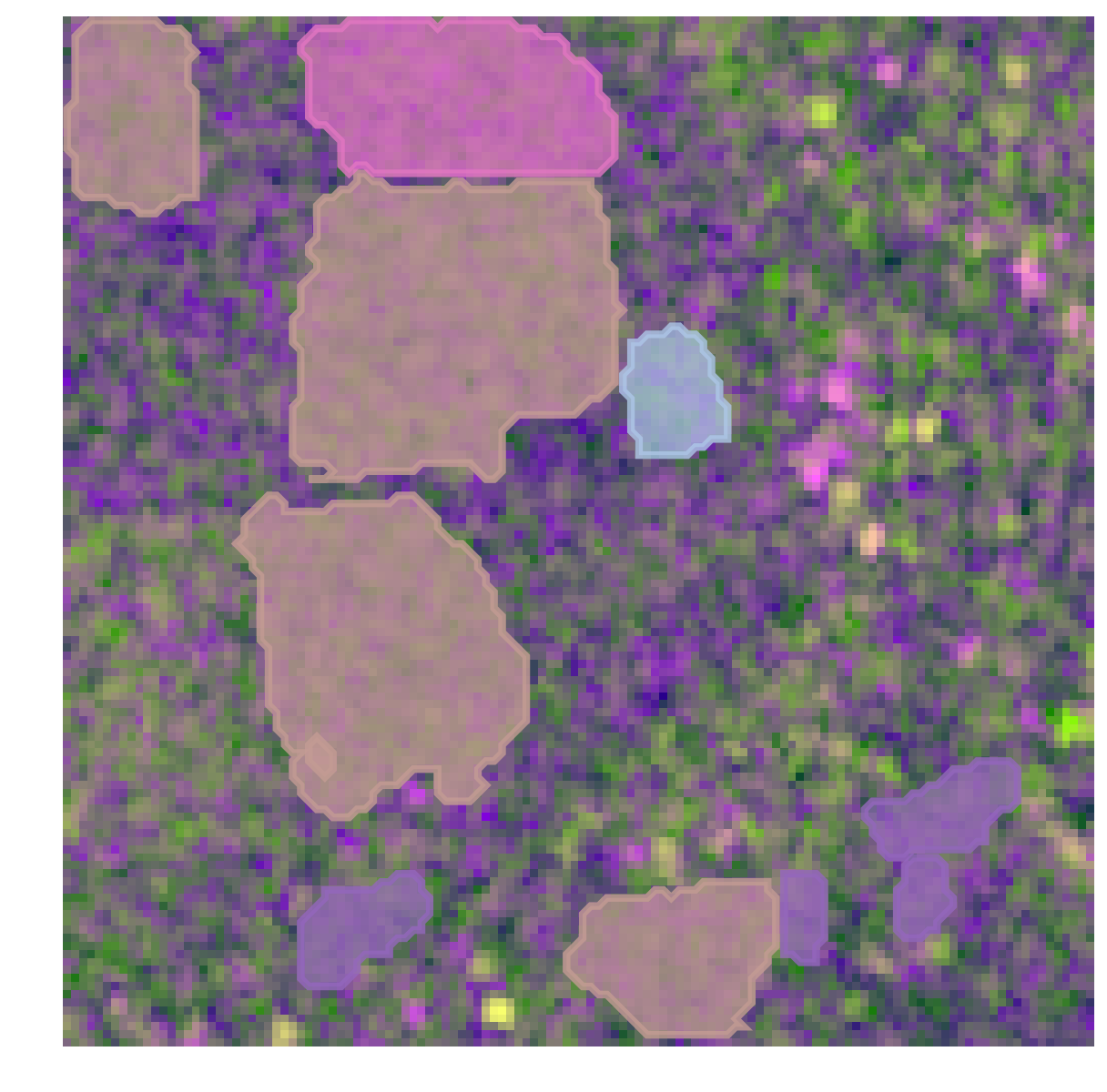}
     & 
    \includegraphics[width=.18\textwidth, trim=1cm 0.5cm 0cm 0cm, clip]{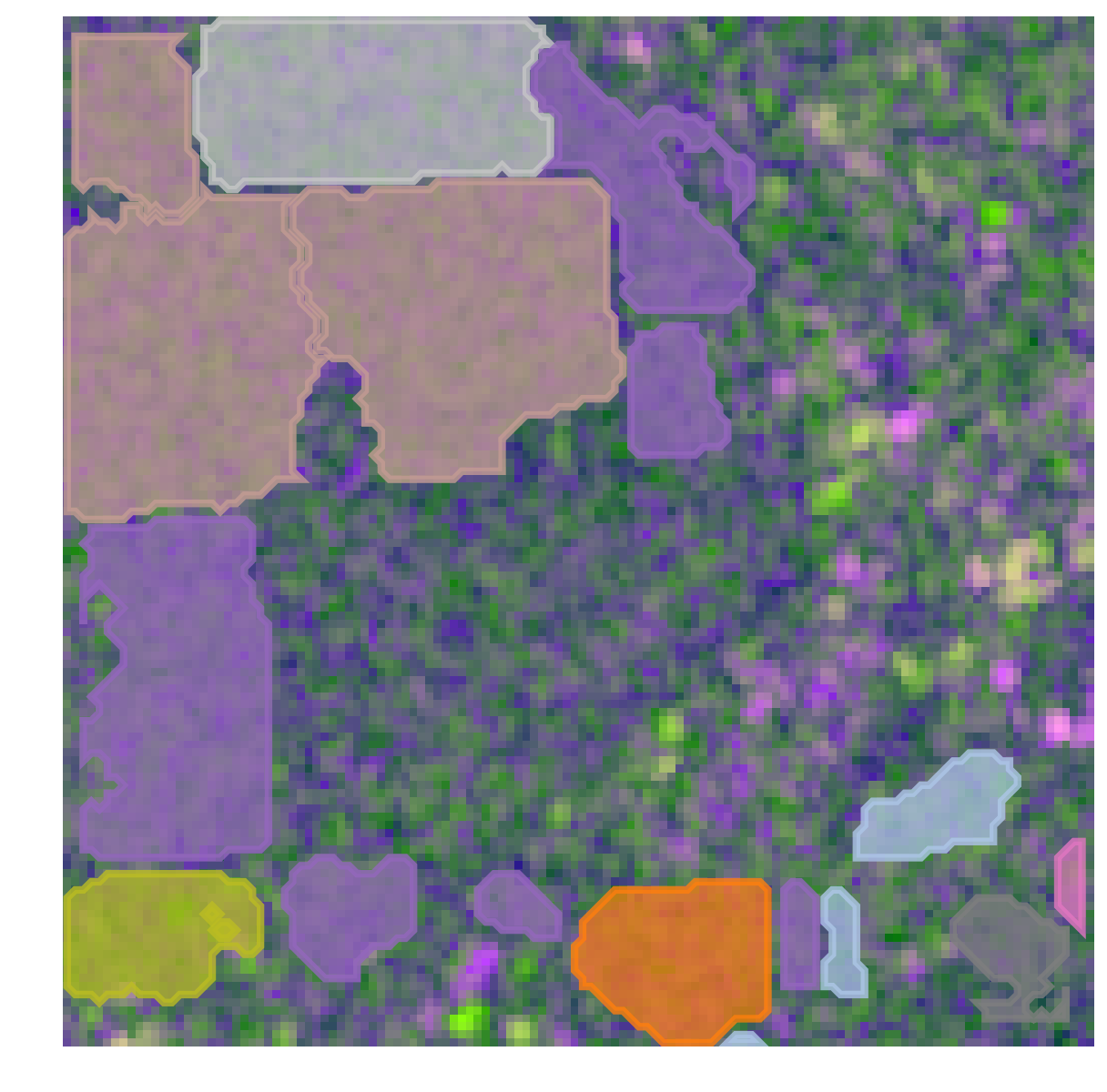}
     & 
    \includegraphics[width=.18\textwidth, trim=1cm 0.5cm 0cm 0cm, clip]{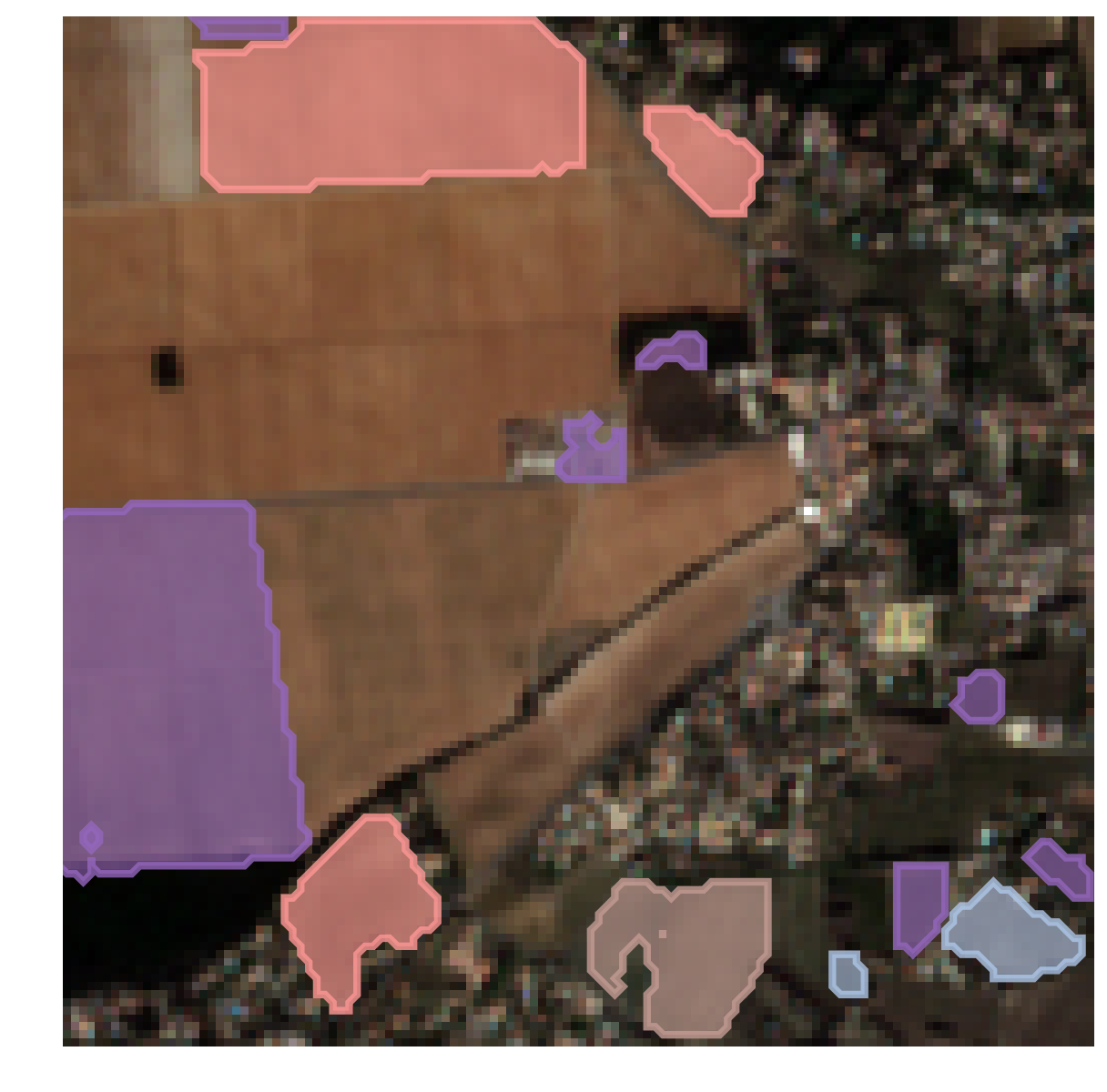}
     & 
    \begin{tikzpicture}
    \node[anchor=south west,inner sep=0] (image) at (0,0) {               \includegraphics[width=.18\textwidth, trim=1cm 0.5cm 0cm 0cm, clip]{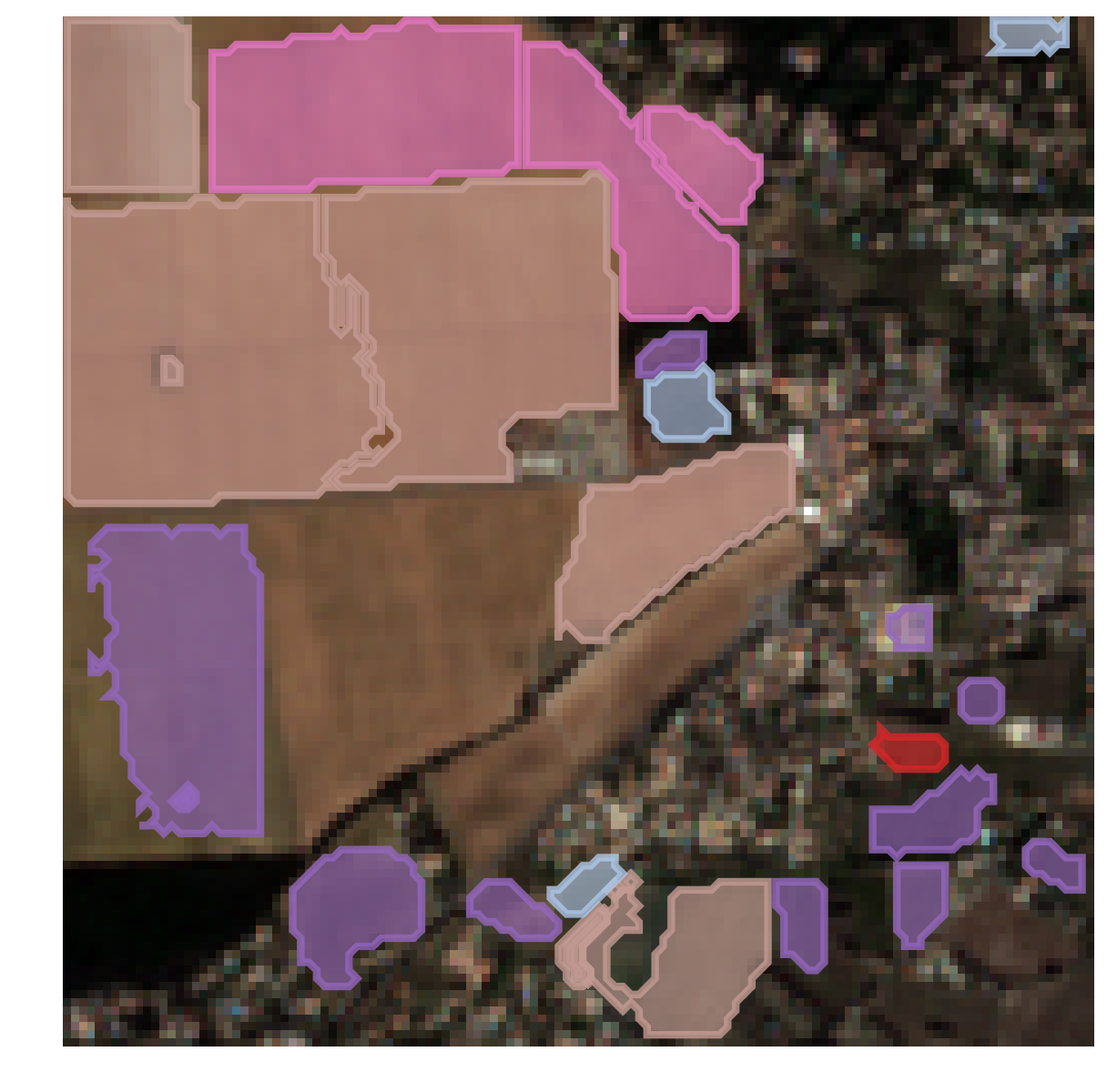}};
    \begin{scope}[x={(image.south east)},y={(image.north west)}]
        \draw[green,ultra thick] (0.3,0.7) circle (0.2);
    \end{scope}
    \end{tikzpicture}
     & 
     \includegraphics[width=.18\textwidth, trim=1cm 0.5cm 0cm 0cm, clip]{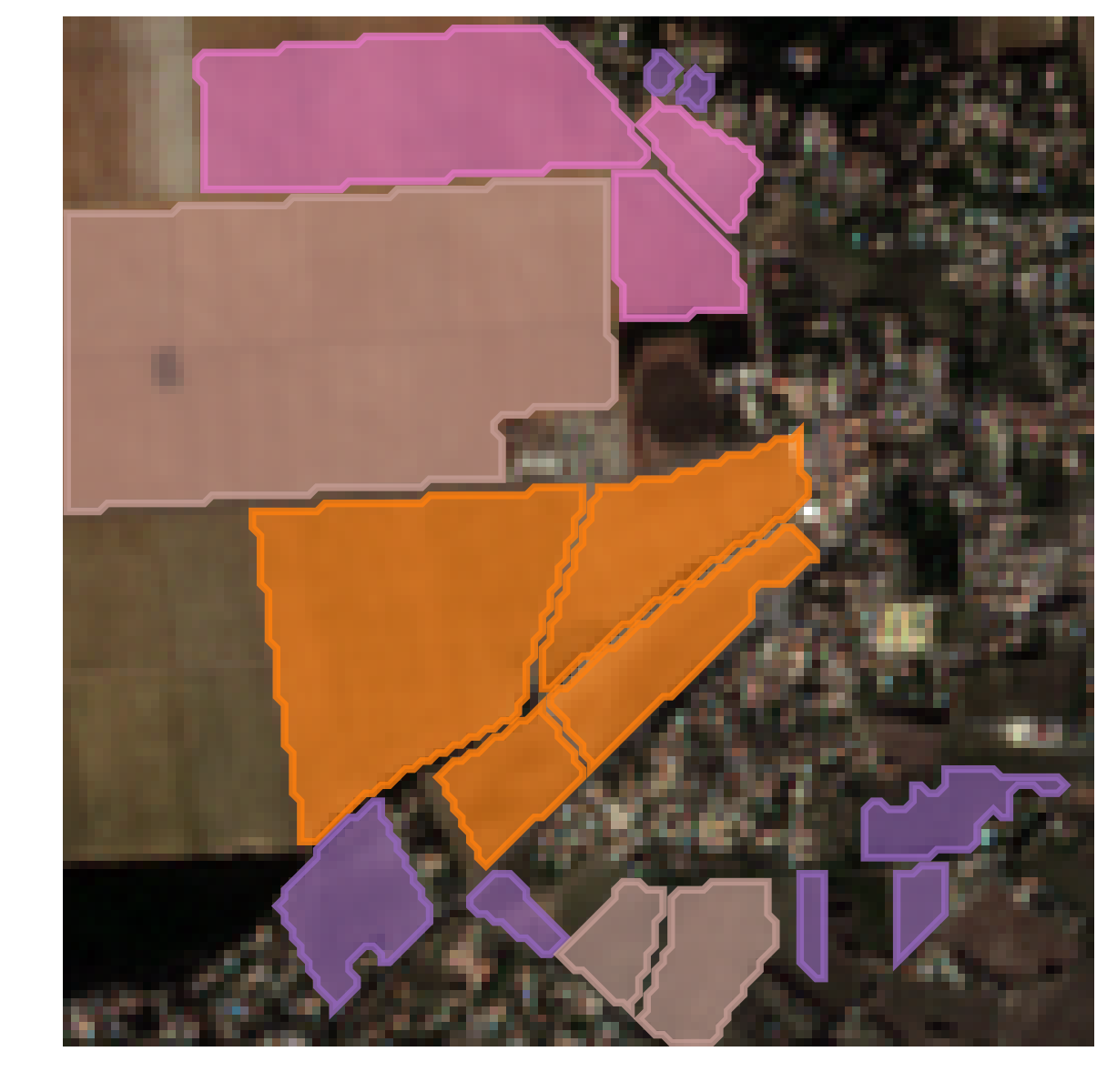}
    \\
     \begin{subfigure}{.19\textwidth}
    \caption{S1A Prediction}
    \label{fig:qualipano:s2}
    \end{subfigure}
    &
    \begin{subfigure}{.19\textwidth}
    \caption{S1D Prediction}
    \label{fig:qualipano:s1}
    \end{subfigure}

    &
    \begin{subfigure}{.19\textwidth}
    \caption{S2 Prediction}
    \label{fig:qualipano:mono}
    \end{subfigure}
    &
    \begin{subfigure}{.19\textwidth}
    \caption{Fusion Prediction}
    \label{fig:qualipano:late}
    \end{subfigure}
    &
     \begin{subfigure}{.19\textwidth}
    \caption{Ground Truth}
    \label{fig:qualipano:gt}
    \end{subfigure}
\end{tabular}

\caption{{\bf Qualitative Results for Panoptic Segmentation.} We compare the predictions made by unimodal models operating on S1A (a), S1D (b), S2 (c), and the predictions made by the late fusion model (d). We also show the ground truth annotations (e). We observe cases where the optical model does not detect parcels but successfully predicted by the radar-only models and by the fusion model as well (green circle 
\protect\tikz \protect\node[circle, thick, draw = green, fill = none, scale = 0.7] {};). We also note that the optical model detects some parcels, but the crop type is corrected by the addition of the radar modality
(red circle 
\protect\tikz \protect\node[circle, thick, draw = red, fill = none, scale = 0.7] {};).
Conversely, some parcels are detected by the radar-only model with an incorrect crop type and not detected by the optical model. Combining both modalities in the fusion model leads to a correct prediction. (cyan circle 
\protect\tikz \protect\node[circle, thick, draw = cyan, fill = none, scale = 0.7] {};)}

\label{fig:qualipanomod}
\end{figure}

We show on \figref{fig:qualipano} the qualitative evaluation of the panoptic fusion model compared to the optical baseline. In practice, the fusion model seems to retrieve more agricultural parcels successfully and manages to retrieve small parcels missed by the optical model. We also display the predictions made by the unimodal models and the predictions of the fusion model in \figref{fig:qualipanomod}. These qualitative results show how the radar modality helps detect more parcels than the optical baseline or improve the fusion model's semantic predictions. Additionally, given the relative noisiness of radar observations, the radar-only models retrieve surprisingly well the parcel boundaries. {As mentioned previously, this could be attributed to the distinct volumetric radar response on parcel boundaries.}  We report the per-class performances on \figref{fig:perclass_pano}. 

Regarding robustness to clouds, when performing inference on only $30$\% of the optical observations, the S2 baseline model drops to $33.0$ PQ. In contrast, our late fusion model maintains a score of $37.6$ PQ. Consistently with the previous experiments, the addition of the radar modality helps improve the panoptic predictions with reduced availability of optical observations.

%\begin{figure}
%    \centering
%    \includegraphics[width=\linewidth]{gfx/Fusion_per_class.pdf}
%    \caption{Caption}
%    \label{fig:per_class}
%\end{figure}

\section{Discussion}
In this section, we discuss the relevance of the different modality fusion strategies, with a focus on Sentinel-1 \& 2 data for crop mapping. Our experiments showed that combining optical and radar imagery allowed for an increase in performance for all tasks considered (\tabref{tab:xp:parcelbased}, \tabref{tab:semantic}, \tabref{tab:xp:pano}) as well as robustness to cloud cover (\figref{fig:london}).

\begin{table}[h]
    \caption{\textbf{Inference times.} We report the inference times in seconds of Early and Late fusion for one fold of PASTIS  ($500$ patches, $820$km$^2$). We measure the combined data loading and prediction time to account for the interpolation step in early fusion }
    \label{tab:inftime:fusion}
    \centering
    \begin{tabular}{lccc}
      & Parcel & Semantic & Panoptic \\
      & classification & segmentation & segmentation \\ \midrule
    Early      & 192 & 280 &\textbf{414}  \\
    Late      & \textbf{149}* & \textbf{259}* & 819 \\ \bottomrule
    \multicolumn{2}{l}{* with auxiliary loss.}
    \end{tabular}
\end{table}

\subsection{Recommendations.}
Our experiments showed that each fusion scheme has advantages and limitations influencing when its use is most relevant:
\begin{itemize}
    \item {\bf Early Fusion.} It is the most compact of the fusion models and shows competitive performance on all three tasks. The main drawback of this approach is the necessity of an expensive interpolation.  As reported in \tabref{tab:inftime:fusion}, this preprocessing makes the early fusion scheme slower than late fusion despite relying on a smaller network for parcel classification and semantic segmentation. Early fusion is the least robust fusion scheme to cloud cover. 
    %It is the most compact of the fusion models, but it underperforms all other fusion schemes. 
    %Additionally, early fusion is incompatible with the two enhancement schemes. We discourage the use of this approach.
    \item {\bf Mid Fusion.} Of all methods without preprocessing, this strategy leads to the fastest run time and the lowest memory requirement. It yields the second-best performance for parcel-based classification but suffers more than late and decision fusion when the cloud cover is extensive. Its dependence on separate spatial and temporal encoders prevents its straightforward adaptation to pixel-based tasks. We recommend using this scheme for parcel classification in areas without extensive cloud cover and when inference speed is critical.  
    \item {\bf Late Fusion.} This fusion method, when combined with enhancement schemes, leads to the best performance and the highest adaptability, as well as excellent resilience to even extreme cloud cover. This method is our default recommendation when using temporal attention methods with multimodal time series.
    \item {\bf Decision Fusion.} Despite having the highest parameter count, this method lags in terms of performance and is prohibitively costly for panoptic segmentation. However, it is the most resilient to cloud cover. We recommend using decision fusion when it is expected that only a few optical observations may be available for inference.
\end{itemize}
We also have evaluated the influence of two enhancement schemes:
\begin{itemize}
    \item {\bf Auxiliary Supervision.} This method consists in adding alongside the main prediction auxiliary predictions based on one modality alone. The rationale is to help each specialized module to learn meaningful features regardless of the interplay with other modalities. We observe a strong effect in precision for late and decision fusion, which have dedicated encoding modules for each modality.   
    \item {\bf Temporal Dropout.}  This simple method consists in randomly dropping acquisitions of the time series considered. Its effect was beneficial to all fusion schemes and the optical baseline across our experiments. Another benefit of this scheme is that it reduces the memory footprint of networks during training.
\end{itemize}

\subsection{Limitations.} 

Our study hinged on the PASTIS dataset, which contains annotated agricultural parcels from four different regions of the French metropolitan territory. In this regard, our results are most relevant for crop mapping applications with the same meteorological context, terrain conditions, and crop types as this region. Certain crop types not observed in PASTIS could benefit even more from the radar modality than our results show. For instance, rice fields are often filled with water and thus have a distinctive SAR response but are not represented in PASTIS.

Furthermore, our evaluation of cloud robustness focused on assessing the effect of a reduced number of optical observations \emph{at inference time}.
This corresponds to artificially increasing the cloud cover in the test set without affecting crop growth. A more rigorous approach would  constitute a dataset comprising truly observed cloud coverage by varying the regions and years of acquisition. This is complicated by the lack of harmonization between LPIS across different countries in nomenclature and open-access policy. 
%Our exploration is thus especially relevant for applications of fusion models trained and applied in two different meteorological contexts. Our study could be complemented by an exploration of robustness to clouds in a setting with similar cloud obstruction both at training and inference time. This would require the preparation of a dataset covering multiple zones with different meteorological conditions, or \LOIC{retraining our models after having withheld  specific ratio of occlusion during training}. 
Lastly, we only used backscattering coefficients from the SAR data in our experiments, as is commonly done in the crop type mapping literature \citep{orynbaikyzy2019crop}. \citet{mestre2020time} found that the addition of interferometric radar features is beneficial to crop classification when using only radar inputs. Further work is needed to  assess the benefit of interferometric radar features in a fusion setting with optical imagery.
%\VIVIEN{Lastly, the SAR data used in our experiments was processed in backscatering coefficients as is most commonly done in the crop type mapping literature \cite{orynbaikyzy2019crop}. \citet{mestre2020time} found that combining backscattering coefficients with interferometric radar features was beneficial to crop type mapping performance when using only radar inputs. Further work could  assess the benefit of interferometric radar features in a fusion setting with optical imagery.}
Moreover, we chose to prepare the SAR inputs with limited preprocessing. We do not apply speckle filtering or  radiometric terrain correction to compensate for the effect of the local incident angle. Interestingly, our experiments showed that this does not prevent the radar modality from benefiting crop mapping models. 
However, further studies could evaluate the benefit of adding speckle filtering, elevation information, or meteorological context to networks using radar images for crop mapping.
%On the matter of speckle, one can hypothesise that the spatial and temporal encoders operating on the radar inputs are sufficient to offset the effect of speckle. However, further studies could aim at evaluating the benefit of adding speckle filtering for radar inputs to self-attention crop mapping models. }

\FloatBarrier
\section{Conclusion}
This article formulated and explored different schemes to design fusion architectures using temporal attention for predicting agricultural crop type maps from radar and optical satellite time series. 

Across the three tasks of parcel-based classification, semantic segmentation, and panoptic segmentation, we experimentally confirmed that the multispectral information of Sentinel-2 proves more discriminative than the SAR signal of Sentinel-1. Yet, for all three tasks, leveraging both modalities led to improvements in the overall performance and the robustness to cloud obstruction. The late fusion scheme, where the learned representations of each modality are concatenated before decoding, outperformed the other approaches on parcel-based classification. Our subsequent exploration of this approach on semantic and panoptic segmentation confirmed its validity to leverage optical and radar time series for crop type mapping. Our experiments also showed that models with less interplay in the encoding of the modalities are most robust to changes in cloud obstruction. In this regard, decision fusion may be favored in contexts with highly unpredictable cloud conditions. Yet, both late and decision fusion approaches proved computationally costly as they incur distinct spatio-temporal encoders for each modality. We introduced a mid-fusion scheme that circumvents this problem by using separate spatial encoders and a shared temporal encoder. This approach performed marginally worse than late fusion on parcel-based classification while having close to half the trainable parameters. Mid-fusion can thus be a valid choice for applications with limited computational resources. Furthermore, the extension of this approach to semantic and panoptic segmentation should be explored in future works. We release PASTIS-R, the augmented version of PASTIS with radar time series, to encourage further endeavors in multi-temporal fusion for Earth Observation.

\paragraph{\bf Aknowledgements} We thank Hugo Lecomte and Nicolas David for their help in the preparation of the PASTIS-R dataset. This work was partly supported by \href{https://www.asp-public.fr}{ASP}, the French Payment Agency.




\bibliographystyle{elsarticle-num-names}
\bibliography{fusion}

\begin{thebibliography}{62}
\expandafter\ifx\csname natexlab\endcsname\relax\def\natexlab#1{#1}\fi
\providecommand{\url}[1]{\texttt{#1}}
\providecommand{\href}[2]{#2}
\providecommand{\path}[1]{#1}
\providecommand{\DOIprefix}{doi:}
\providecommand{\ArXivprefix}{arXiv:}
\providecommand{\URLprefix}{URL: }
\providecommand{\Pubmedprefix}{pmid:}
\providecommand{\doi}[1]{\href{http://dx.doi.org/#1}{\path{#1}}}
\providecommand{\Pubmed}[1]{\href{pmid:#1}{\path{#1}}}
\providecommand{\bibinfo}[2]{#2}
\ifx\xfnm\relax \def\xfnm[#1]{\unskip,\space#1}\fi
%Type = Misc
\bibitem[{Garnot and Landrieu(2021)}]{pastis}
\bibinfo{author}{V.~S.~F. Garnot}, \bibinfo{author}{L.~Landrieu},
  \bibinfo{title}{Pastis - panoptic segmentation of satellite image time
  series}, \bibinfo{year}{2021}. \URLprefix
  \url{https://zenodo.org/record/5012942}.
  \DOIprefix\doi{10.5281/ZENODO.5012942}.
%Type = Article
\bibitem[{Van~Tricht et~al.(2018)Van~Tricht, Gobin, Gilliams, and
  Piccard}]{van2018synergistic}
\bibinfo{author}{K.~Van~Tricht}, \bibinfo{author}{A.~Gobin},
  \bibinfo{author}{S.~Gilliams}, \bibinfo{author}{I.~Piccard},
\newblock \bibinfo{title}{Synergistic use of radar {S}entinel-1 and optical
  {S}entinel-2 imagery for crop mapping: a case study for {B}elgium},
\newblock \bibinfo{journal}{Remote Sensing}  (\bibinfo{year}{2018}).
%Type = Article
\bibitem[{Steinhausen et~al.(2018)Steinhausen, Wagner, Narasimhan, and
  Waske}]{steinhausen2018combining}
\bibinfo{author}{M.~J. Steinhausen}, \bibinfo{author}{P.~D. Wagner},
  \bibinfo{author}{B.~Narasimhan}, \bibinfo{author}{B.~Waske},
\newblock \bibinfo{title}{Combining {S}entinel-1 and {S}entinel-2 data for
  improved land use and land cover mapping of monsoon regions},
\newblock \bibinfo{journal}{International journal of applied earth observation
  and geoinformation}  (\bibinfo{year}{2018}).
%Type = Article
\bibitem[{Campos-Taberner et~al.(2019)Campos-Taberner, Garc{\'\i}a-Haro,
  Mart{\'\i}nez, S{\'a}nchez-Ru{\'\i}z, and Gilabert}]{campos2019copernicus}
\bibinfo{author}{M.~Campos-Taberner}, \bibinfo{author}{F.~J. Garc{\'\i}a-Haro},
  \bibinfo{author}{B.~Mart{\'\i}nez},
  \bibinfo{author}{S.~S{\'a}nchez-Ru{\'\i}z}, \bibinfo{author}{M.~A. Gilabert},
\newblock \bibinfo{title}{A {C}opernicus {S}entinel-1 and {S}entinel-2
  classification framework for the 2020+ {E}uropean common agricultural policy:
  A case study in {V}al{\`e}ncia ({S}pain)},
\newblock \bibinfo{journal}{Agronomy}  (\bibinfo{year}{2019}).
%Type = Article
\bibitem[{Vrieling et~al.(2018)Vrieling, Meroni, Darvishzadeh, Skidmore, Wang,
  Zurita-Milla, Oosterbeek, O'Connor, and Paganini}]{vrieling2018vegetation}
\bibinfo{author}{A.~Vrieling}, \bibinfo{author}{M.~Meroni},
  \bibinfo{author}{R.~Darvishzadeh}, \bibinfo{author}{A.~K. Skidmore},
  \bibinfo{author}{T.~Wang}, \bibinfo{author}{R.~Zurita-Milla},
  \bibinfo{author}{K.~Oosterbeek}, \bibinfo{author}{B.~O'Connor},
  \bibinfo{author}{M.~Paganini},
\newblock \bibinfo{title}{Vegetation phenology from sentinel-2 and field
  cameras for a dutch barrier island},
\newblock \bibinfo{journal}{Remote sensing of environment}
  (\bibinfo{year}{2018}).
%Type = Article
\bibitem[{Segarra et~al.(2020)Segarra, Buchaillot, Araus, and
  Kefauver}]{segarra2020remote}
\bibinfo{author}{J.~Segarra}, \bibinfo{author}{M.~L. Buchaillot},
  \bibinfo{author}{J.~L. Araus}, \bibinfo{author}{S.~C. Kefauver},
\newblock \bibinfo{title}{Remote sensing for precision agriculture: Sentinel-2
  improved features and applications},
\newblock \bibinfo{journal}{Agronomy}  (\bibinfo{year}{2020}).
%Type = Article
\bibitem[{Tucker(1979)}]{TUCKERNDVI1979127}
\bibinfo{author}{C.~J. Tucker},
\newblock \bibinfo{title}{Red and photographic infrared linear combinations for
  monitoring vegetation},
\newblock \bibinfo{journal}{Remote Sensing of Environment}
  (\bibinfo{year}{1979}).
%Type = Article
\bibitem[{Sudmanns et~al.(2020)Sudmanns, Tiede, Augustin, and
  Lang}]{sudmanns2020assessing}
\bibinfo{author}{M.~Sudmanns}, \bibinfo{author}{D.~Tiede},
  \bibinfo{author}{H.~Augustin}, \bibinfo{author}{S.~Lang},
\newblock \bibinfo{title}{Assessing global sentinel-2 coverage dynamics and
  data availability for operational earth observation (eo) applications using
  the eo-compass},
\newblock \bibinfo{journal}{International Journal of Digital Earth}
  (\bibinfo{year}{2020}).
%Type = Article
\bibitem[{McNairn et~al.(2014)McNairn, Kross, Lapen, Caves, and
  Shang}]{mcnairn2014early}
\bibinfo{author}{H.~McNairn}, \bibinfo{author}{A.~Kross},
  \bibinfo{author}{D.~Lapen}, \bibinfo{author}{R.~Caves},
  \bibinfo{author}{J.~Shang},
\newblock \bibinfo{title}{Early season monitoring of corn and soybeans with
  terrasar-x and radarsat-2},
\newblock \bibinfo{journal}{International Journal of Applied Earth Observation
  and Geoinformation} \bibinfo{volume}{28} (\bibinfo{year}{2014}).
%Type = Misc
\bibitem[{Commission(2016)}]{CAP}
\bibinfo{author}{E.~Commission}, \bibinfo{title}{The common agricultural policy
  at a glance},
  \bibinfo{howpublished}{\url{ec.europa.eu/info/food-farming-fisheries/key-policies/common-agricultural-policy/cap-glance_en}},
  \bibinfo{year}{2016}. \bibinfo{note}{Accessed: 2021-09-24}.
%Type = Inproceedings
\bibitem[{Koetz et~al.(2019)Koetz, Defourny, Bontemps, Bajec, Cara,
  de~Vendictis, Kucera, Malcorps, Milcinski, Nicola et~al.}]{koetz2019sen4cap}
\bibinfo{author}{B.~Koetz}, \bibinfo{author}{P.~Defourny},
  \bibinfo{author}{S.~Bontemps}, \bibinfo{author}{K.~Bajec},
  \bibinfo{author}{C.~Cara}, \bibinfo{author}{L.~de~Vendictis},
  \bibinfo{author}{L.~Kucera}, \bibinfo{author}{P.~Malcorps},
  \bibinfo{author}{G.~Milcinski}, \bibinfo{author}{L.~Nicola}, et~al.,
\newblock \bibinfo{title}{Sen4cap sentinels for cap monitoring approach},
\newblock in: \bibinfo{booktitle}{JRC IACS Workshop}, \bibinfo{year}{2019}.
%Type = Article
\bibitem[{Drusch et~al.(2012)Drusch, Del~Bello, Carlier, Colin, Fernandez,
  Gascon, Hoersch, Isola, Laberinti, Martimort et~al.}]{drusch2012sentinel}
\bibinfo{author}{M.~Drusch}, \bibinfo{author}{U.~Del~Bello},
  \bibinfo{author}{S.~Carlier}, \bibinfo{author}{O.~Colin},
  \bibinfo{author}{V.~Fernandez}, \bibinfo{author}{F.~Gascon},
  \bibinfo{author}{B.~Hoersch}, \bibinfo{author}{C.~Isola},
  \bibinfo{author}{P.~Laberinti}, \bibinfo{author}{P.~Martimort}, et~al.,
\newblock \bibinfo{title}{Sentinel-2: Esa's optical high-resolution mission for
  gmes operational services},
\newblock \bibinfo{journal}{Remote sensing of Environment}
  (\bibinfo{year}{2012}).
%Type = Misc
\bibitem[{EtaLab(2017)}]{RPG}
\bibinfo{author}{EtaLab}, \bibinfo{title}{Registre parcellaire graphique (rpg)
  : contours des parcelles et îlots culturaux et leur groupe de cultures
  majoritaire},
  \bibinfo{howpublished}{\url{data.gouv.fr/en/datasets/registre-parcellaire-graphique-rpg-contours-des-parcelles-et-ilots-culturaux-et-leur-groupe-de-cultures-majoritaire/}},
  \bibinfo{year}{2017}.
%Type = Article
\bibitem[{He and Yokoya(2018)}]{he2018multi}
\bibinfo{author}{W.~He}, \bibinfo{author}{N.~Yokoya},
\newblock \bibinfo{title}{Multi-temporal {S}entinel-1 and-2 data fusion for
  optical image simulation},
\newblock \bibinfo{journal}{ISPRS Journal}  (\bibinfo{year}{2018}).
%Type = Article
\bibitem[{Orynbaikyzy et~al.(2020)Orynbaikyzy, Gessner, Mack, and
  Conrad}]{orynbaikyzy2020crop}
\bibinfo{author}{A.~Orynbaikyzy}, \bibinfo{author}{U.~Gessner},
  \bibinfo{author}{B.~Mack}, \bibinfo{author}{C.~Conrad},
\newblock \bibinfo{title}{Crop type classification using fusion of {S}entinel-1
  and {S}entinel-2 data: Assessing the impact of feature selection, optical
  data availability, and parcel sizes on the accuracies},
\newblock \bibinfo{journal}{Remote Sensing}  (\bibinfo{year}{2020}).
%Type = Article
\bibitem[{Giordano et~al.(2020)Giordano, Bailly, Landrieu, and
  Chehata}]{giordano2020improved}
\bibinfo{author}{S.~Giordano}, \bibinfo{author}{S.~Bailly},
  \bibinfo{author}{L.~Landrieu}, \bibinfo{author}{N.~Chehata},
\newblock \bibinfo{title}{Improved crop classification with rotation knowledge
  using sentinel-1 and-2 time series},
\newblock \bibinfo{journal}{Photogrammetric Engineering \& Remote Sensing}
  (\bibinfo{year}{2020}).
%Type = Article
\bibitem[{Ienco et~al.(2019)Ienco, Interdonato, Gaetano, and
  Minh}]{ienco2019combining}
\bibinfo{author}{D.~Ienco}, \bibinfo{author}{R.~Interdonato},
  \bibinfo{author}{R.~Gaetano}, \bibinfo{author}{D.~H.~T. Minh},
\newblock \bibinfo{title}{Combining {S}entinel-1 and {S}entinel-2 satellite
  image time series for land cover mapping via a multi-source deep learning
  architecture},
\newblock \bibinfo{journal}{ISPRS Journal}  (\bibinfo{year}{2019}).
%Type = Article
\bibitem[{Ru{\ss}wurm and K{\"o}rner(2020)}]{russwurm2020self}
\bibinfo{author}{M.~Ru{\ss}wurm}, \bibinfo{author}{M.~K{\"o}rner},
\newblock \bibinfo{title}{Self-attention for raw optical satellite time series
  classification},
\newblock \bibinfo{journal}{ISPRS Journal}  (\bibinfo{year}{2020}).
%Type = Inproceedings
\bibitem[{Garnot et~al.(2020)Garnot, Landrieu, Giordano, and
  Chehata}]{garnot2020satellite}
\bibinfo{author}{V.~S.~F. Garnot}, \bibinfo{author}{L.~Landrieu},
  \bibinfo{author}{S.~Giordano}, \bibinfo{author}{N.~Chehata},
\newblock \bibinfo{title}{Satellite image time series classification with
  pixel-set encoders and temporal self-attention},
\newblock in: \bibinfo{booktitle}{CPVR}, \bibinfo{year}{2020}.
%Type = Article
\bibitem[{Kondmann et~al.(2021)Kondmann, Toker, Ru{\ss}wurm, Unzueta,
  Peressuti, Milcinski, Mathieu, Long{\'e}p{\'e}, Davis, Marchisio
  et~al.}]{kondmann2021denethor}
\bibinfo{author}{L.~Kondmann}, \bibinfo{author}{A.~Toker},
  \bibinfo{author}{M.~Ru{\ss}wurm}, \bibinfo{author}{A.~C. Unzueta},
  \bibinfo{author}{D.~Peressuti}, \bibinfo{author}{G.~Milcinski},
  \bibinfo{author}{P.-P. Mathieu}, \bibinfo{author}{N.~Long{\'e}p{\'e}},
  \bibinfo{author}{T.~Davis}, \bibinfo{author}{G.~Marchisio}, et~al.,
\newblock \bibinfo{title}{Denethor: The dynamicearthnet dataset for harmonized,
  inter-operable, analysis-ready, daily crop monitoring from space}
  (\bibinfo{year}{2021}). \URLprefix \url{openreview.net/forum?id=uUa4jNMLjrL}.
%Type = Inproceedings
\bibitem[{Garnot and Landrieu(2020)}]{garnot2020lightweight}
\bibinfo{author}{V.~S.~F. Garnot}, \bibinfo{author}{L.~Landrieu},
\newblock \bibinfo{title}{Lightweight temporal self-attention for classifying
  satellite images time series},
\newblock in: \bibinfo{booktitle}{AALTD}, \bibinfo{year}{2020}.
%Type = Article
\bibitem[{Ofori-Ampofo et~al.(2021)Ofori-Ampofo, Pelletier, and
  Lang}]{pelletier2021fusion}
\bibinfo{author}{S.~Ofori-Ampofo}, \bibinfo{author}{C.~Pelletier},
  \bibinfo{author}{S.~Lang},
\newblock \bibinfo{title}{Crop type mapping from optical and radar time series
  using attention-based deep learning},
\newblock \bibinfo{journal}{{R}emote {S}ensing}  (\bibinfo{year}{2021}).
%Type = Inproceedings
\bibitem[{Kirillov et~al.(2019)Kirillov, He, Girshick, Rother, and
  Doll{\'a}r}]{kirillov2019panoptic}
\bibinfo{author}{A.~Kirillov}, \bibinfo{author}{K.~He},
  \bibinfo{author}{R.~Girshick}, \bibinfo{author}{C.~Rother},
  \bibinfo{author}{P.~Doll{\'a}r},
\newblock \bibinfo{title}{Panoptic segmentation},
\newblock in: \bibinfo{booktitle}{CVPR}, \bibinfo{year}{2019}.
%Type = Inproceedings
\bibitem[{Garnot and Landrieu(2021)}]{garnot2021utae}
\bibinfo{author}{V.~S.~F. Garnot}, \bibinfo{author}{L.~Landrieu},
\newblock \bibinfo{title}{Panoptic segmentation of satellite image time series
  with convolutional temporal attention networks},
\newblock in: \bibinfo{booktitle}{ICCV}, \bibinfo{year}{2021}.
%Type = Article
\bibitem[{Joshi et~al.(2016)Joshi, Baumann, Ehammer, Fensholt, Grogan, Hostert,
  Jepsen, Kuemmerle, Meyfroidt, Mitchard et~al.}]{joshi2016review}
\bibinfo{author}{N.~Joshi}, \bibinfo{author}{M.~Baumann},
  \bibinfo{author}{A.~Ehammer}, \bibinfo{author}{R.~Fensholt},
  \bibinfo{author}{K.~Grogan}, \bibinfo{author}{P.~Hostert},
  \bibinfo{author}{M.~R. Jepsen}, \bibinfo{author}{T.~Kuemmerle},
  \bibinfo{author}{P.~Meyfroidt}, \bibinfo{author}{E.~T. Mitchard}, et~al.,
\newblock \bibinfo{title}{A review of the application of optical and radar
  remote sensing data fusion to land use mapping and monitoring},
\newblock \bibinfo{journal}{Remote Sensing}  (\bibinfo{year}{2016}).
%Type = Article
\bibitem[{Mercier et~al.(2019)Mercier, Betbeder, Rumiano, Baudry, Gond, Blanc,
  Bourgoin, Cornu, Marchamalo, Poccard-Chapuis et~al.}]{mercier2019evaluation}
\bibinfo{author}{A.~Mercier}, \bibinfo{author}{J.~Betbeder},
  \bibinfo{author}{F.~Rumiano}, \bibinfo{author}{J.~Baudry},
  \bibinfo{author}{V.~Gond}, \bibinfo{author}{L.~Blanc},
  \bibinfo{author}{C.~Bourgoin}, \bibinfo{author}{G.~Cornu},
  \bibinfo{author}{M.~Marchamalo}, \bibinfo{author}{R.~Poccard-Chapuis},
  et~al.,
\newblock \bibinfo{title}{Evaluation of {S}entinel-1 and 2 time series for land
  cover classification of forest-agriculture mosaics in temperate and tropical
  landscapes},
\newblock \bibinfo{journal}{Remote Sensing}  (\bibinfo{year}{2019}).
%Type = Article
\bibitem[{Tarpanelli et~al.(2018)Tarpanelli, Santi, Tourian, Filippucci,
  Amarnath, and Brocca}]{tarpanelli2018daily}
\bibinfo{author}{A.~Tarpanelli}, \bibinfo{author}{E.~Santi},
  \bibinfo{author}{M.~J. Tourian}, \bibinfo{author}{P.~Filippucci},
  \bibinfo{author}{G.~Amarnath}, \bibinfo{author}{L.~Brocca},
\newblock \bibinfo{title}{Daily river discharge estimates by merging satellite
  optical sensors and radar altimetry through artificial neural network},
\newblock \bibinfo{journal}{IEEE Transactions on Geoscience and Remote Sensing}
   (\bibinfo{year}{2018}).
%Type = Article
\bibitem[{Kussul et~al.(2017)Kussul, Lavreniuk, Skakun, and
  Shelestov}]{kussul2017deep}
\bibinfo{author}{N.~Kussul}, \bibinfo{author}{M.~Lavreniuk},
  \bibinfo{author}{S.~Skakun}, \bibinfo{author}{A.~Shelestov},
\newblock \bibinfo{title}{Deep learning classification of land cover and crop
  types using remote sensing data},
\newblock \bibinfo{journal}{Geoscience and Remote Sensing Letters}
  (\bibinfo{year}{2017}).
%Type = Article
\bibitem[{Benedetti et~al.(2018)Benedetti, Ienco, Gaetano, Ose, Pensa, and
  Dupuy}]{benedetti2018m}
\bibinfo{author}{P.~Benedetti}, \bibinfo{author}{D.~Ienco},
  \bibinfo{author}{R.~Gaetano}, \bibinfo{author}{K.~Ose},
  \bibinfo{author}{R.~G. Pensa}, \bibinfo{author}{S.~Dupuy},
\newblock \bibinfo{title}{{M3Fusion}: A deep learning architecture for
  multiscale multimodal multitemporal satellite data fusion},
\newblock \bibinfo{journal}{JSTARS}  (\bibinfo{year}{2018}).
%Type = Article
\bibitem[{Tom et~al.(2021)Tom, Jiang, Baltsavias, and
  Schindler}]{tom2021fusionVIIRSS1}
\bibinfo{author}{M.~Tom}, \bibinfo{author}{Y.~Jiang},
  \bibinfo{author}{E.~Baltsavias}, \bibinfo{author}{K.~Schindler},
\newblock \bibinfo{title}{Learning a sensor-invariant embedding of satellite
  data: A case study for lake ice monitoring},
\newblock \bibinfo{journal}{arXiv preprint arXiv:2107.09092}
  (\bibinfo{year}{2021}).
%Type = Article
\bibitem[{Liu et~al.(2016)Liu, Gong, Qin, and Zhang}]{liu2016deep}
\bibinfo{author}{J.~Liu}, \bibinfo{author}{M.~Gong}, \bibinfo{author}{K.~Qin},
  \bibinfo{author}{P.~Zhang},
\newblock \bibinfo{title}{A deep convolutional coupling network for change
  detection based on heterogeneous optical and radar images},
\newblock \bibinfo{journal}{Transactions on neural networks and learning
  systems}  (\bibinfo{year}{2016}).
%Type = Article
\bibitem[{Garioud et~al.(2020)Garioud, Valero, Giordano, and
  Mallet}]{garioud2020joint}
\bibinfo{author}{A.~Garioud}, \bibinfo{author}{S.~Valero},
  \bibinfo{author}{S.~Giordano}, \bibinfo{author}{C.~Mallet},
\newblock \bibinfo{title}{On the joint exploitation of optical and {SAR}
  satellite imagery for grassland monitoring},
\newblock \bibinfo{journal}{International Archives of the Photogrammetry,
  Remote Sensing and Spatial Information Sciences}  (\bibinfo{year}{2020}).
%Type = Article
\bibitem[{Meraner et~al.(2020)Meraner, Ebel, Zhu, and
  Schmitt}]{meraner2020cloudcgan}
\bibinfo{author}{A.~Meraner}, \bibinfo{author}{P.~Ebel}, \bibinfo{author}{X.~X.
  Zhu}, \bibinfo{author}{M.~Schmitt},
\newblock \bibinfo{title}{Cloud removal in {S}entinel-2 imagery using a deep
  residual neural network and {SAR}-optical data fusion},
\newblock \bibinfo{journal}{ISPRS Journal}  (\bibinfo{year}{2020}).
%Type = Book
\bibitem[{Richards et~al.(2009)}]{richards2009remote}
\bibinfo{author}{J.~A. Richards}, et~al., \bibinfo{title}{Remote sensing with
  imaging radar}, volume~\bibinfo{volume}{1}, \bibinfo{publisher}{Springer},
  \bibinfo{year}{2009}.
%Type = Article
\bibitem[{Orynbaikyzy et~al.(2019)Orynbaikyzy, Gessner, and
  Conrad}]{orynbaikyzy2019crop}
\bibinfo{author}{A.~Orynbaikyzy}, \bibinfo{author}{U.~Gessner},
  \bibinfo{author}{C.~Conrad},
\newblock \bibinfo{title}{Crop type classification using a combination of
  optical and radar remote sensing data: a review},
\newblock \bibinfo{journal}{{I}nternational {J}ournal of {R}emote {S}ensing}
  (\bibinfo{year}{2019}).
%Type = Article
\bibitem[{Simons and Rosen(2007)}]{simons2007interferometric}
\bibinfo{author}{M.~Simons}, \bibinfo{author}{P.~Rosen},
\newblock \bibinfo{title}{Interferometric synthetic aperture radar geodesy},
\newblock \bibinfo{journal}{Treatise on Geophysics - Geodesy}
  (\bibinfo{year}{2007}).
%Type = Article
\bibitem[{Monserrat et~al.(2014)Monserrat, Crosetto, and
  Luzi}]{monserrat2014review}
\bibinfo{author}{O.~Monserrat}, \bibinfo{author}{M.~Crosetto},
  \bibinfo{author}{G.~Luzi},
\newblock \bibinfo{title}{A review of ground-based sar interferometry for
  deformation measurement},
\newblock \bibinfo{journal}{ISPRS Journal of Photogrammetry and Remote Sensing}
   (\bibinfo{year}{2014}).
%Type = Article
\bibitem[{Tarchi et~al.(2003)Tarchi, Casagli, Fanti, Leva, Luzi, Pasuto,
  Pieraccini, and Silvano}]{tarchi2003landslide}
\bibinfo{author}{D.~Tarchi}, \bibinfo{author}{N.~Casagli},
  \bibinfo{author}{R.~Fanti}, \bibinfo{author}{D.~D. Leva},
  \bibinfo{author}{G.~Luzi}, \bibinfo{author}{A.~Pasuto},
  \bibinfo{author}{M.~Pieraccini}, \bibinfo{author}{S.~Silvano},
\newblock \bibinfo{title}{Landslide monitoring by using ground-based sar
  interferometry: an example of application to the tessina landslide in italy},
\newblock \bibinfo{journal}{Engineering geology} \bibinfo{volume}{68}
  (\bibinfo{year}{2003}) \bibinfo{pages}{15--30}.
%Type = Article
\bibitem[{Tom{\'a}s et~al.(2012)Tom{\'a}s, Garc{\'\i}a-Barba, Cano, Sanabria,
  Ivorra, Duro, and Herrera}]{tomas2012subsidence}
\bibinfo{author}{R.~Tom{\'a}s}, \bibinfo{author}{J.~Garc{\'\i}a-Barba},
  \bibinfo{author}{M.~Cano}, \bibinfo{author}{M.~P. Sanabria},
  \bibinfo{author}{S.~Ivorra}, \bibinfo{author}{J.~Duro},
  \bibinfo{author}{G.~Herrera},
\newblock \bibinfo{title}{Subsidence damage assessment of a gothic church using
  differential interferometry and field data},
\newblock \bibinfo{journal}{Structural Health Monitoring}
  (\bibinfo{year}{2012}).
%Type = Article
\bibitem[{Tarchi et~al.(1997)Tarchi, Ohlmer, and Sieber}]{tarchi1997monitoring}
\bibinfo{author}{D.~Tarchi}, \bibinfo{author}{E.~Ohlmer},
  \bibinfo{author}{A.~Sieber},
\newblock \bibinfo{title}{Monitoring of structural changes by radar
  interferometry},
\newblock \bibinfo{journal}{Journal of Research in Nondestructive Evaluation}
  (\bibinfo{year}{1997}).
%Type = Article
\bibitem[{Tison et~al.(2007)Tison, Tupin, and
  Ma{\^\i}tre}]{tisontupin2007fusion}
\bibinfo{author}{C.~Tison}, \bibinfo{author}{F.~Tupin},
  \bibinfo{author}{H.~Ma{\^\i}tre},
\newblock \bibinfo{title}{A fusion scheme for joint retrieval of urban height
  map and classification from high-resolution interferometric sar images},
\newblock \bibinfo{journal}{IEEE Transactions on Geoscience and remote Sensing}
  \bibinfo{volume}{45} (\bibinfo{year}{2007}) \bibinfo{pages}{496--505}.
%Type = Article
\bibitem[{Tamm et~al.(2016)Tamm, Zalite, Voormansik, and
  Talgre}]{tamm2016relating}
\bibinfo{author}{T.~Tamm}, \bibinfo{author}{K.~Zalite},
  \bibinfo{author}{K.~Voormansik}, \bibinfo{author}{L.~Talgre},
\newblock \bibinfo{title}{Relating sentinel-1 interferometric coherence to
  mowing events on grasslands},
\newblock \bibinfo{journal}{{R}emote {S}ensing}  (\bibinfo{year}{2016}).
%Type = Article
\bibitem[{Mestre-Quereda et~al.(2020)Mestre-Quereda, Lopez-Sanchez,
  Vicente-Guijalba, Jacob, and Engdahl}]{mestre2020time}
\bibinfo{author}{A.~Mestre-Quereda}, \bibinfo{author}{J.~M. Lopez-Sanchez},
  \bibinfo{author}{F.~Vicente-Guijalba}, \bibinfo{author}{A.~W. Jacob},
  \bibinfo{author}{M.~E. Engdahl},
\newblock \bibinfo{title}{Time-series of sentinel-1 interferometric coherence
  and backscatter for crop-type mapping},
\newblock \bibinfo{journal}{{J}ournal of {S}elected {T}opics in {A}pplied
  {E}arth {O}bservations and {R}emote {S}ensing}  (\bibinfo{year}{2020}).
%Type = Article
\bibitem[{Shang et~al.(2020)Shang, Liu, Poncos, Geng, Qian, Chen, Dong,
  Macdonald, Martin, Kovacs et~al.}]{shang2020detection}
\bibinfo{author}{J.~Shang}, \bibinfo{author}{J.~Liu},
  \bibinfo{author}{V.~Poncos}, \bibinfo{author}{X.~Geng},
  \bibinfo{author}{B.~Qian}, \bibinfo{author}{Q.~Chen},
  \bibinfo{author}{T.~Dong}, \bibinfo{author}{D.~Macdonald},
  \bibinfo{author}{T.~Martin}, \bibinfo{author}{J.~Kovacs}, et~al.,
\newblock \bibinfo{title}{Detection of crop seeding and harvest through
  analysis of time-series sentinel-1 interferometric sar data},
\newblock \bibinfo{journal}{{R}emote {S}ensing}  (\bibinfo{year}{2020}).
%Type = Article
\bibitem[{Srivastava et~al.(2006)Srivastava, Patel, and
  Navalgund}]{srivastava2006application}
\bibinfo{author}{H.~S. Srivastava}, \bibinfo{author}{P.~Patel},
  \bibinfo{author}{R.~R. Navalgund},
\newblock \bibinfo{title}{Application potentials of synthetic aperture radar
  interferometry for land-cover mapping and crop-height estimation},
\newblock \bibinfo{journal}{{C}urrent {S}cience}  (\bibinfo{year}{2006}).
%Type = Article
\bibitem[{Cloude and Pottier(1996)}]{cloude1996review}
\bibinfo{author}{S.~R. Cloude}, \bibinfo{author}{E.~Pottier},
\newblock \bibinfo{title}{A review of target decomposition theorems in radar
  polarimetry},
\newblock \bibinfo{journal}{Transactions on geoscience and remote sensing}
  (\bibinfo{year}{1996}).
%Type = Article
\bibitem[{Yamaguchi et~al.(2005)Yamaguchi, Moriyama, Ishido, and
  Yamada}]{yamaguchi2005four}
\bibinfo{author}{Y.~Yamaguchi}, \bibinfo{author}{T.~Moriyama},
  \bibinfo{author}{M.~Ishido}, \bibinfo{author}{H.~Yamada},
\newblock \bibinfo{title}{Four-component scattering model for polarimetric sar
  image decomposition},
\newblock \bibinfo{journal}{Transactions on Geoscience and Remote Sensing}
  (\bibinfo{year}{2005}).
%Type = Article
\bibitem[{Srikanth et~al.(2016)Srikanth, Ramana, Deepika, Chakravarthi, and
  Sai}]{srikanth2016comparison}
\bibinfo{author}{P.~Srikanth}, \bibinfo{author}{K.~Ramana},
  \bibinfo{author}{U.~Deepika}, \bibinfo{author}{P.~K. Chakravarthi},
  \bibinfo{author}{M.~S. Sai},
\newblock \bibinfo{title}{Comparison of various polarimetric decomposition
  techniques for crop classification},
\newblock \bibinfo{journal}{{J}ournal of the {I}ndian {S}ociety of {R}emote
  {S}ensing}  (\bibinfo{year}{2016}).
%Type = Article
\bibitem[{Schuler et~al.(1996)Schuler, Lee, and
  De~Grandi}]{schuler1996measurement}
\bibinfo{author}{D.~L. Schuler}, \bibinfo{author}{J.-S. Lee},
  \bibinfo{author}{G.~De~Grandi},
\newblock \bibinfo{title}{Measurement of topography using polarimetric sar
  images},
\newblock \bibinfo{journal}{IEEE Transactions on Geoscience and Remote Sensing}
   (\bibinfo{year}{1996}).
%Type = Article
\bibitem[{Tupin et~al.(1998)Tupin, Maitre, Mangin, Nicolas, and
  Pechersky}]{tupin1998detection}
\bibinfo{author}{F.~Tupin}, \bibinfo{author}{H.~Maitre}, \bibinfo{author}{J.-F.
  Mangin}, \bibinfo{author}{J.-M. Nicolas}, \bibinfo{author}{E.~Pechersky},
\newblock \bibinfo{title}{Detection of linear features in sar images:
  Application to road network extraction},
\newblock \bibinfo{journal}{Transactions on geoscience and remote sensing}
  (\bibinfo{year}{1998}).
%Type = Book
\bibitem[{Kourgli et~al.(2010)Kourgli, Ouarzeddine, Oukil, and
  Belhadj-Aissa}]{kourgli2010land}
\bibinfo{author}{A.~Kourgli}, \bibinfo{author}{M.~Ouarzeddine},
  \bibinfo{author}{Y.~Oukil}, \bibinfo{author}{A.~Belhadj-Aissa},
  \bibinfo{title}{Land cover identification using polarimetric SAR images},
  \bibinfo{publisher}{na}, \bibinfo{year}{2010}.
%Type = Article
\bibitem[{Garnot and Landrieu(2021)}]{garnot2021panoptic}
\bibinfo{author}{V.~S.~F. Garnot}, \bibinfo{author}{L.~Landrieu},
\newblock \bibinfo{title}{Panoptic segmentation of satellite image time series
  with convolutional temporal attention networks},
\newblock \bibinfo{journal}{ICCV}  (\bibinfo{year}{2021}).
%Type = Article
\bibitem[{Kaplan et~al.(2021)Kaplan, Fine, Lukyanov, Manivasagam, Tanny, and
  Rozenstein}]{kaplan2021S1localinc}
\bibinfo{author}{G.~Kaplan}, \bibinfo{author}{L.~Fine},
  \bibinfo{author}{V.~Lukyanov}, \bibinfo{author}{V.~Manivasagam},
  \bibinfo{author}{J.~Tanny}, \bibinfo{author}{O.~Rozenstein},
\newblock \bibinfo{title}{Normalizing the local incidence angle in {S}entinel-1
  imagery to improve leaf area index, vegetation height, and crop coefficient
  estimations},
\newblock \bibinfo{journal}{Land}  (\bibinfo{year}{2021}).
%Type = Article
\bibitem[{Garkusha et~al.(2017)Garkusha, Hnatushenko, and
  V}]{garkusha2017research}
\bibinfo{author}{I.~N. Garkusha}, \bibinfo{author}{V.~Hnatushenko},
  \bibinfo{author}{V.~V},
\newblock \bibinfo{title}{Research of influence of atmosphere and humidity on
  the data of radar imaging by sentinel-1}  (\bibinfo{year}{2017}).
%Type = Article
\bibitem[{Abramov et~al.(2017)Abramov, Rubel, Lukin, Kozhemiakin, Kussul,
  Shelestov, and Lavreniuk}]{abramov2017speckle}
\bibinfo{author}{S.~Abramov}, \bibinfo{author}{O.~Rubel},
  \bibinfo{author}{V.~Lukin}, \bibinfo{author}{R.~Kozhemiakin},
  \bibinfo{author}{N.~Kussul}, \bibinfo{author}{A.~Shelestov},
  \bibinfo{author}{M.~Lavreniuk},
\newblock \bibinfo{title}{Speckle reducing for sentinel-1 sar data},
\newblock \bibinfo{journal}{IGARSS}  (\bibinfo{year}{2017}).
%Type = Article
\bibitem[{Srivastava et~al.(2014)Srivastava, Hinton, Krizhevsky, Sutskever, and
  Salakhutdinov}]{srivastava2014dropout}
\bibinfo{author}{N.~Srivastava}, \bibinfo{author}{G.~Hinton},
  \bibinfo{author}{A.~Krizhevsky}, \bibinfo{author}{I.~Sutskever},
  \bibinfo{author}{R.~Salakhutdinov},
\newblock \bibinfo{title}{Dropout: a simple way to prevent neural networks from
  overfitting},
\newblock \bibinfo{journal}{The journal of machine learning research}
  (\bibinfo{year}{2014}).
%Type = Inproceedings
\bibitem[{Schneider and K{\"o}rner(2020)}]{schneider2020re}
\bibinfo{author}{M.~Schneider}, \bibinfo{author}{M.~K{\"o}rner},
\newblock \bibinfo{title}{[re] satellite image time series classification with
  pixel-set encoders and temporal self-attention},
\newblock in: \bibinfo{booktitle}{ML Reproducibility Challenge 2020},
  \bibinfo{year}{2020}.
%Type = Article
\bibitem[{Christophe et~al.(2008)Christophe, Inglada, and
  Giros}]{christophe2008orfeo}
\bibinfo{author}{E.~Christophe}, \bibinfo{author}{J.~Inglada},
  \bibinfo{author}{A.~Giros},
\newblock \bibinfo{title}{Orfeo toolbox: a complete solution for mapping from
  high resolution satellite images},
\newblock \bibinfo{journal}{International Archives of the Photogrammetry,
  Remote Sensing and Spatial Information Sciences} \bibinfo{volume}{37}
  (\bibinfo{year}{2008}) \bibinfo{pages}{1263--1268}.
%Type = Article
\bibitem[{Singhroy and Saint-Jean(1999)}]{singhroy1999effects}
\bibinfo{author}{V.~Singhroy}, \bibinfo{author}{R.~Saint-Jean},
\newblock \bibinfo{title}{Effects of relief on the selection of radarsat-1
  incidence angle for geological applications},
\newblock \bibinfo{journal}{Canadian Journal of Remote Sensing}
  (\bibinfo{year}{1999}).
%Type = Misc
\bibitem[{Garnot and Landrieu(2021)}]{pastis-r}
\bibinfo{author}{V.~S.~F. Garnot}, \bibinfo{author}{L.~Landrieu},
  \bibinfo{title}{Pastis-r - panoptic segmentation of radar and optical
  satellite image time series}, \bibinfo{year}{2021}. \URLprefix
  \url{https://zenodo.org/record/5735646}.
  \DOIprefix\doi{10.5281/ZENODO.5735646}.
%Type = Article
\bibitem[{Kingma and Ba(2015)}]{kingma2014adam}
\bibinfo{author}{D.~P. Kingma}, \bibinfo{author}{J.~Ba},
\newblock \bibinfo{title}{Adam: A method for stochastic optimization},
\newblock \bibinfo{journal}{ICLR}  (\bibinfo{year}{2015}).
%Type = Article
\bibitem[{Wang et~al.(2020)Wang, Zhang, and Grosse}]{wang2020picking}
\bibinfo{author}{C.~Wang}, \bibinfo{author}{G.~Zhang},
  \bibinfo{author}{R.~Grosse},
\newblock \bibinfo{title}{Picking winning tickets before training by preserving
  gradient flow},
\newblock \bibinfo{journal}{ICLR}  (\bibinfo{year}{2020}).

\end{thebibliography}

\end{document}